\documentclass[review]{elsarticle}

\usepackage{lineno,hyperref}
\modulolinenumbers[5]

\usepackage{amsmath}
\usepackage{dsfont}
\usepackage{amssymb}
\usepackage{extarrows}
\usepackage{mathrsfs}
\usepackage{booktabs}
\usepackage{multirow}
\usepackage{threeparttable}
\usepackage[justification=raggedright]{caption}
\usepackage{threeparttable}
\usepackage[misc]{ifsym}
\usepackage{array}
\usepackage{graphicx}
\usepackage{float}
\usepackage{lineno}
\usepackage{subfigure}

\makeatletter
\@addtoreset{equation}{section}
\makeatother

\graphicspath{ {./figures/} }

\begin{document}
	\begin{frontmatter}
		
		\title{\bf\Large Parameter Identification for Partial Differential Equations with Spatiotemporal Varying Coefficients}

		\address[addr1]{SandGold AI Research, Guangzhou 510642, China}
		\address[addr2]{Department of Mathematics, Faculty of Science and  Technology, \\
			University of Macau, Macau 519000, China}
		\address[addr3]{School of Reliability and Systems Engineering, \\
			Beihang University, Beijing 10091, China}
		\address[addr4]{College of Mathematics and Informatics \\
			South China Agricultural University, Guangzhou 510642, China}
		\address[addr5]{College of Information Science and Technology, \\
			Jinan University, Guangzhou 511486, China}
		\address[addr6]{School of Mathematics and Statistics, \\
			Huazhong University of Sciences and Technology, Wuhan 430074, China}
		
		\author{Guangtao Zhang\fnref{addr1,addr2}}
		
		\author{Yiting Duan\fnref{addr3}}
		
		\author{Guanyu Pan\fnref{addr1,addr4} }
		
		\author{Qijing Chen\fnref{addr1,addr5} }
		
		\author{Huiyu Yang\fnref{addr1,addr4} }
		
		\author{Zhikun Zhang\fnref{addr1,addr6}\corref{mycorrespondingauthor}}
		\cortext[mycorrespondingauthor]{Corresponding author}
		\ead{zkzhang@hust.edu.cn}

		\begin{abstract}
			To comprehend complex systems with multiple states, it is imperative to reveal the identity of these states by system outputs. Nevertheless, the mathematical models describing these systems often exhibit nonlinearity so that render the resolution of the parameter inverse problem from the observed spatiotemporal data a challenging endeavor. Starting from the observed data obtained from such systems, we propose a novel framework that facilitates the investigation of parameter identification for multi-state systems governed by spatiotemporal varying parametric partial differential equations. Our framework consists of two integral components: a constrained self-adaptive physics-informed neural network, encompassing a sub-network, as our methodology for parameter identification, and a finite mixture model approach to detect regions of probable parameter variations. Through our scheme, we can precisely ascertain the unknown varying parameters of the complex multi-state system, thereby accomplishing the inversion of the varying parameters. Furthermore, we have showcased the efficacy of our framework on two numerical cases: the 1D Burgers' equation with time-varying parameters and the 2D wave equation with a space-varying parameter.
		\end{abstract}
		
		
	\end{frontmatter}

	\section{Introduction}

	Parameter identification for partial differential equations (PDEs) is also known as the inverse problem, encompassing various mathematical branches such as numerical analysis, nonlinear analysis, and optimization algorithms. The target of the inverse problem is inferring unknown parameters of PDE from a set of spatiotemporal data with potential noise \cite{zhang2018robust} and this field has progressed rapidly over the past few decades with proposed methods such as the sparse Bayesian learning algorithm \cite{yuan2017sparse}, the least squares method \cite{qi2009multi}, the frequency and Bayesian methods \cite{frasso2016parameter}, and the physics-informed neural networks (PINNs) \cite{raissi2019physics}, etc. In the mechanics of material fields, accurate property parameter detection will benefit the damage detection and design for new multi-functional materials \cite{li2004multi}. In biomechanics, identifying important parameters in human tissue can be helpful for treatment and disease prevention \cite{fatemi1998ultrasound, cai2021artificial}. And parameter identification method is also widely used in other engineering fields such as oil exploration and fluid mechanism \cite{fienen2004application, brigham2007inverse}.     
	
	Nowadays, multi-state systems with time-varying or space-varying parameters have been widely used in fields such as physics, biology, chemical processes \cite{swain1984handbook}, and society \cite{helfmann2020extending}. One of the most powerful ways of understanding a multi-state complex system with time-varying parameters is discovering its state transition path. Various theories have been proposed to model and characterize the system dynamics such as the transition path theory \cite{vanden2006transition}, the transition path sampling \cite{dellago2002transition}, and the Markov state model \cite{chodera2014markov}. For the system governed by a varying parametric PDE, the evolutionary process of the varying parameters determines the state transition path of the multi-state system. For space-varying parameters in higher dimensions, the transition region could be inscribed instead of the transition path. As such, identifying the unknown varying parameters is becoming a necessary first step for discovering the pattern variation in complex systems.
	
	The PINNs have been demonstrated as an efficient way to infer the unknown parameters of PDEs from the observed data. The original idea of PINNs was introduced by Lagaris in 1998 \cite{lagris1998artificial}, and has been well established by Raissi et al. for solving two main problems: the forward problem for PDE resolution and parameter identification for PDE \cite{raissi2017physics, Raissi2017PhysicsID}. From Raissi, varied numerical techniques have been proposed to improve the performance of PINNs for that two problems \cite {yu2022gradient,krishnapriyan2021characterizing}
	and been successfully used in solving problems in materials \cite{chen2020physics}, biology \cite{daneker2022systems}, topological optimization \cite{lu2021physics}, and fluid\cite{kadeethum2020physics}.  For the varying parameter inferring task, Revanth et al. proposed the backward compatible PINNs(bc-PINNs)\cite{mattey2022novel} to learn time-varying parameters of time-varying parametric Burgers' equation from the observed data without any prior information. However, the inferring results of bc-PINNs only follow a trend similar to the true values. As a result, such inaccurate results are insufficient for us to explore the transition path. To solve the above, we need a more accurate parameter identification method. 
	
	After obtaining the inferring results of the varying parametric PDEs, the next part is detecting the change region of the varying parameters. Change-point detection is an important part of time series analysis and probability anomaly detection \cite{aminikhanghahi2017survey}. This work requires us to pinpoint the locations of changes in statistical characteristics and points in time at which the probability density functions change \cite{james2020novel}. Based on the parameter inferring results of a varying system, a fast and accurate change point detection method may contribute to detecting the change points of the system and locating their position, which may be signification for us to discover the state transition path. For time series data, There has been extensive work in detection change points \cite{ross2015parametric, truong2020selective, goswami2022change} and becomes a signification part of controlling the reliability and stability of the system. Unlike time series analysis, in this study, the change points of time-varying and space-varying parameters make up the region of variation about the intrinsic nature of the multi-state complex system. It would be interesting research to reveal this hidden parameter variation from the output of the system.
	
	Data-driven statistical modeling based on finite mixture distributions is a rapidly evolving field, with a wide range of applications expanding rapidly \cite{mclachlan2019finite}. Recently,  the finite mixture models is utilized in various fields, such as biometrics, physics, medicine, and marketing. It offers a straightforward method for describing a continuous system's variation through discrete state space. Despite being a simple linear extension of the classical statistical model, finite mixture models share features concerning inference, specifically a discrete latent structure that results in certain fundamental challenges in estimation, such as the need to determine the unknown number of groups, states, and clusters. The expectation maximization(EM) algorithm is an iterative technique based on maximum likelihood estimation for estimating the parameters of statistical models when the data comprises both observed and hidden variables in the context of finite mixture models \cite{mclachlan2007algorithm}. The key advantage of the EM algorithm is that it provides a means of estimating the parameters of models with latent variables without explicitly computing the posterior distribution of the latent variables. This statistical method is particularly useful when there are missing or incomplete data, or when the data is partially observed. This can be computationally efficient, especially when dealing with complex models.
	
	In this paper, we introduce a novel framework for discovering the state transition path of a multi-state parametric PDE system in two steps. Firstly, we use the modified constrained self-adaptive physics-informed neural networks (cSPINNs) to identify the unknown varying parameters and then detect the change region via a change point detection method based on a finite mixture model. Specifically, we modify the cSPINNs by adding a sub-network to learn the varying parameters and this can obtain more accurate results than the previous bc-PINNs. Next, we detect the change points concerning where the parameter change based on the inferring results by employing the finite mixture method. Finally, we take the 1D time-varying parametric Burgers' Equation and 2D space-varying wave equation as test examples to demonstrate the performance of our method.
	
	This paper is structured as follows. In section \ref{Second_sec}, we describe forms of parametric partial differential equations with time and space-varying parameters which are the test cases for our method. In section \ref{Third_sec}, the proposed framework containing two main methods for discovering the state transition path is presented in detail. In section \ref{forth_sec}, we test the performance of our framework based on the 1D time-varying parametric Burgers' equation and the 2D space-varying parametric wave equation and analysis their results. Section \ref{fifth_sec} is the comparison of cSPINNs and bc-PINNs via 1D Burgers' equation. Section \ref{six_sec} is the performance of our framework on 2D space-varying wave equation and section \ref{seven_sec} is the conclusion and discussion.
	
	\section{Parametric Partial Differential Equations with Time and Space Varying Parameter}\label{Second_sec}
	To elucidate the situation of the partial differential equations with Time-varying parameters, we use the following 1D time-varying parametric Burgers' equation as an example.
	The Burgers' equation is a nonlinear second-order partial differential equation that is used as a simplified model in fluid mechanics. The equation is given by the Dutch mathematician Johannes Burgers' \cite{burgers2013nonlinear} and in this study, we generally write as the following form
	\begin{equation}\label{2.1}
		u_t=\lambda_1 uu_x+\lambda_2 u_{xx},
	\end{equation}
	where $u$ is the fluid velocity at position $x$ and time $t$, the term $\lambda_1 u u_x$ is known as the convective term, the term $\lambda_2 u_{xx}$ is the diffusive term and $\lambda_2$ is the kinematic viscosity of the fluid. The Burgers' equation combines the effects of convection and diffusion in a non-linear way and is used to model a variety of phenomena in fluid mechanics, including shock waves, turbulence, and flow in porous media \cite{smoller2012shock}. Besides fluid mechanics, it has also been used in other areas of physics, such as in modeling traffic flow in transportation engineering \cite{velasco2007first}.
	
	Let $\lambda_1$ and $\lambda_2$ be time-varying parameters and take values in a finite discrete parameter space. We rewrite equation \eqref{2.1} as
	\begin{equation}\label{2.2}
		u_t=\lambda_1(t) uu_x+\lambda_2(t) u_{xx}. 
	\end{equation}
	Thus we get a continuous system with discrete states. In this time-varying parameter system, the parameter may exhibit local invariance. As such, in the global time domain, research attention is directed toward how the system state changes over time and at which points these changes occur. The subsequent objective of this study is to establish a comprehensive mathematical framework that builds upon existing solutions of the system. This framework serves to address the inverse problem for parameters in the equation and change point detection of time-varying parameter systems.
	
\begin{figure}[t]
		\centering
		\begin{minipage}{0.32\textwidth}
			\centering
			\includegraphics[width=1\textwidth]{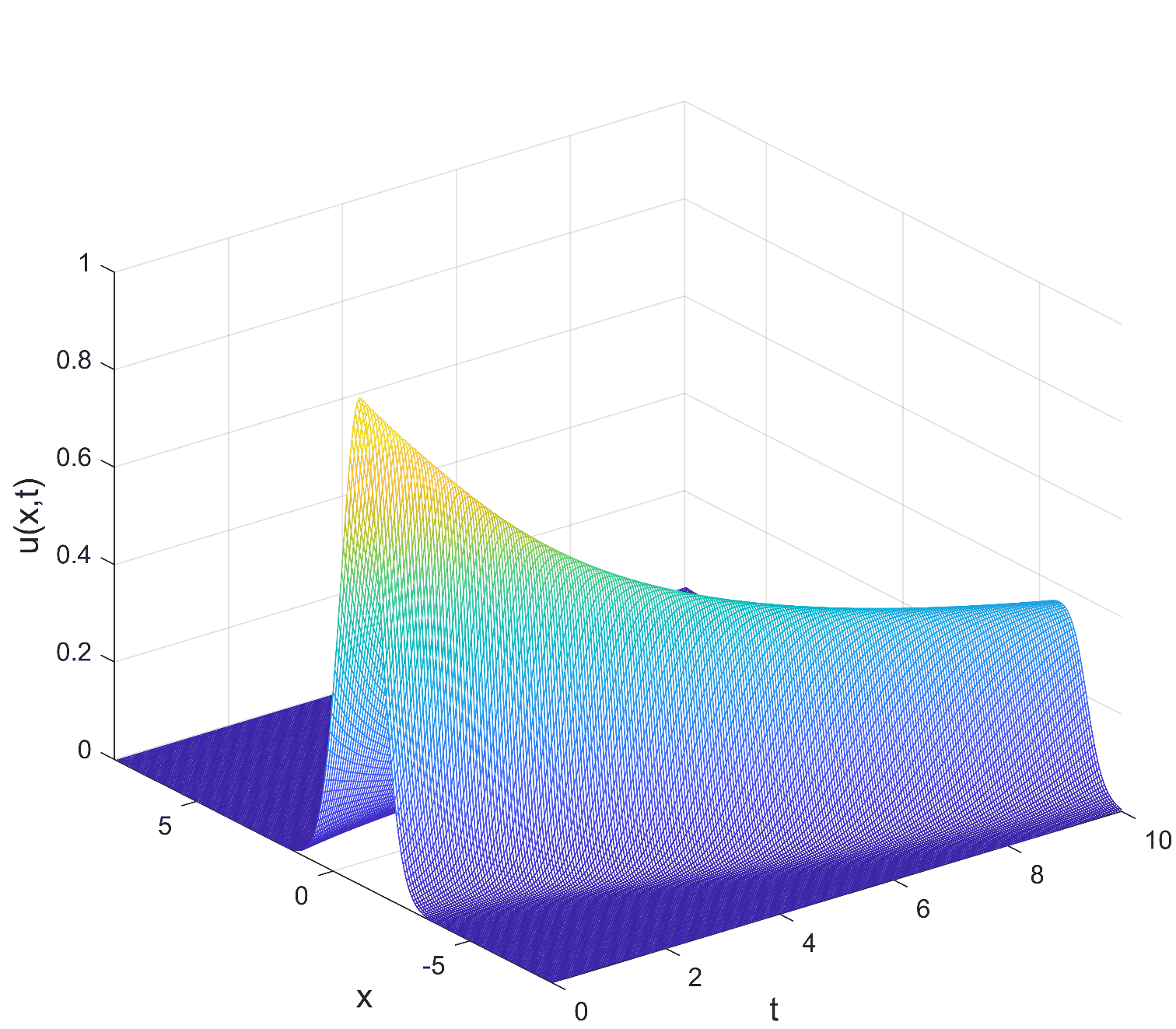}
		\end{minipage}
		\begin{minipage}{0.32\textwidth}
			\centering
			\includegraphics[width=1\textwidth]{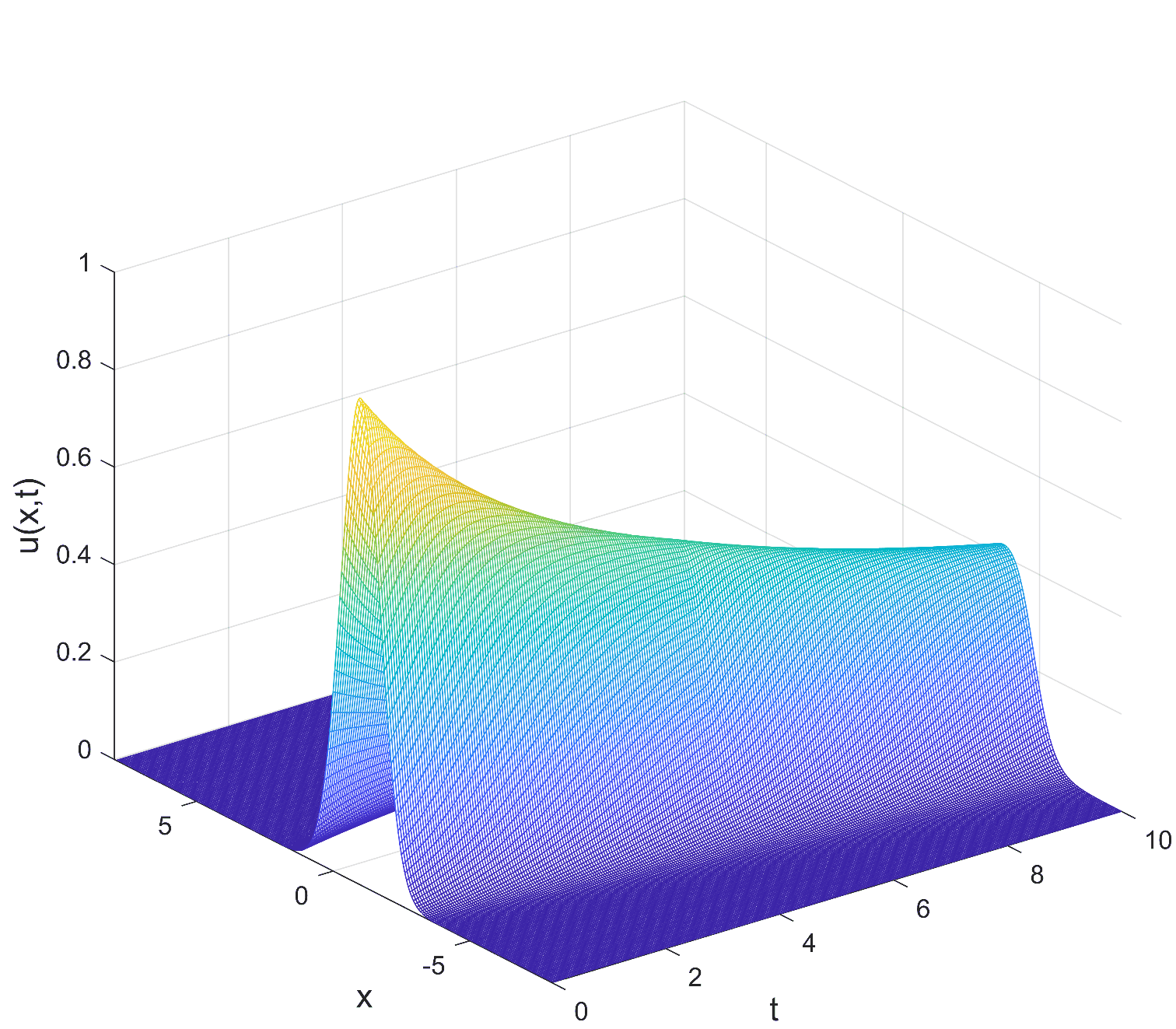}
		\end{minipage}
		\begin{minipage}{0.32\textwidth}
			\centering
			\includegraphics[width=1\textwidth]{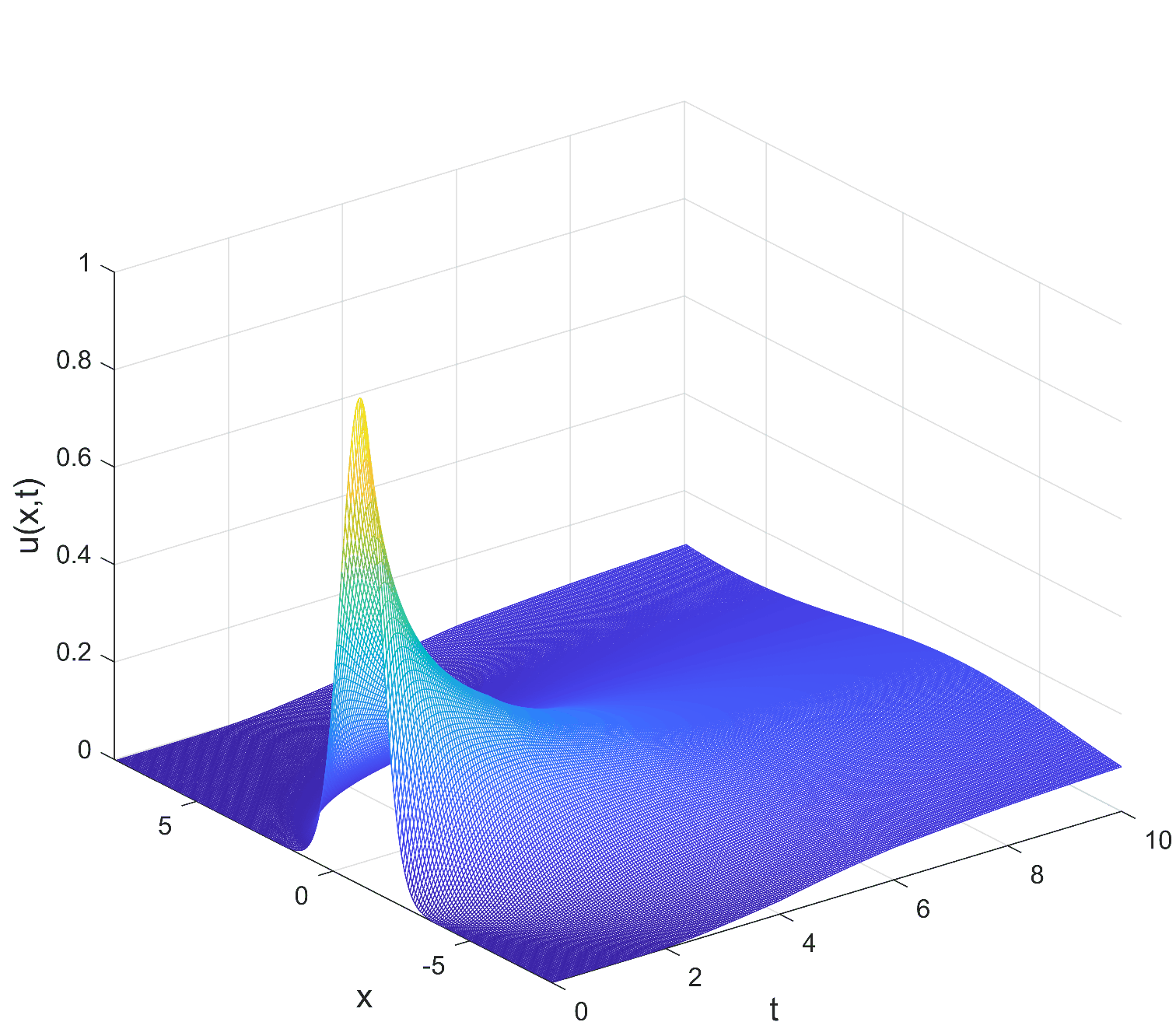}
		\end{minipage}
		\caption{Numerical solution of the 1D time-varying parametric Burgers' equation for three cases. From left to right, figures correspond to constant parameters without change point, time-varying $\lambda_{1}$ with one change point, and time-varying $\lambda_{1}$ and $\lambda_{2}$ with multiple change points.\label{Numerical Solution of three cases}}
	\end{figure}

	In this paper, the observed data of the 1D time-varying parametric Burgers' equation is computed via the numerical method fast Fourier transform where the initial value is as givens:
	\begin{equation}
		u(x,0)=\exp{\{-(x+1)^2\}},
	\end{equation}
	and the domain is $(x, t) \in[-8,8] \times(0,10]$. The observed data of three cases: constant parameters without change point, only $\lambda_{1}$ changes once, and $\lambda_{1}$ and $\lambda_{2}$ are all change with multiple change points are shown in the figure \ref{Numerical Solution of three cases}.
	
	From the above figures, it is clear that the transition path of states and the change points of the time-varying system can not be revealed directly from the observed data. Moreover, the difference between the first and the second is obscure, let alone the system information and state transfer paths. Therefore, we can apply the modified cSPINNs as a bridge to link the observed data and the unknown parameters. In this way, we can discover the hidden information together with the transition path.    
	
	For the situation of partial differential equations with space-varying parameters, as a contrast to that previous example, we will introduce the 2D wave equation, whose parameters are not a constant in the space plane. The wave equation is a mathematical model that describes wave phenomena. It is typically expressed as a partial differential equation and can describe wave processes in both space and time. It has widespread applications in physics, engineering, mathematics, and other fields. The general form of the wave equation can be written as:
	\begin{equation}\label{2.4}
		u_{tt}=\alpha^2  \nabla^2 u.
	\end{equation}
	Here, the $u$ represents the wave amplitude, the $\alpha$ is the wave speed, and the $\nabla^2$ is the Laplacian operator, which represents the second derivative in space. This equation describes how the wave amplitude changes and propagates during a wave process. The second time derivative represents the acceleration of the wave amplitude, while the Laplacian operator represents the second derivative in space. Let $\alpha$ be a space-varying parameter $\alpha(x,y)$ and then rewrite the \eqref{2.4} as
	\begin{equation}\label{2.5}
		u_{tt}=[\alpha(x,y)]^2  \nabla^2 u.
	\end{equation}
	In many cases of scientific computing research, there can be sudden and discontinuous changes in local regions, which can have a significant impact on the output of the system. Therefore, it is of great practical significance to obtain the regions of varying neutral states in such a space through scientific calculations. By doing so, we can better understand the underlying physical processes and develop more accurate models to describe them.

	\section{Data-driven Discovery of Parameter Identification Framework for Partial Differential Equations}\label{Third_sec}
	
	In this section, we use two parts to introduce our state transition path discovery framework for a varying parameter system. Firstly, we illustrate the modified cSPINNs method for identifying varying parameters from the observed data. Next, we describe the finite mixture model as our change point detection method.  
	
	\subsection{Modified Constrained Self-adaptive Physics Informed Neural Networks}
\begin{figure}[ht]
		\centering
		\includegraphics[width=1\textwidth]{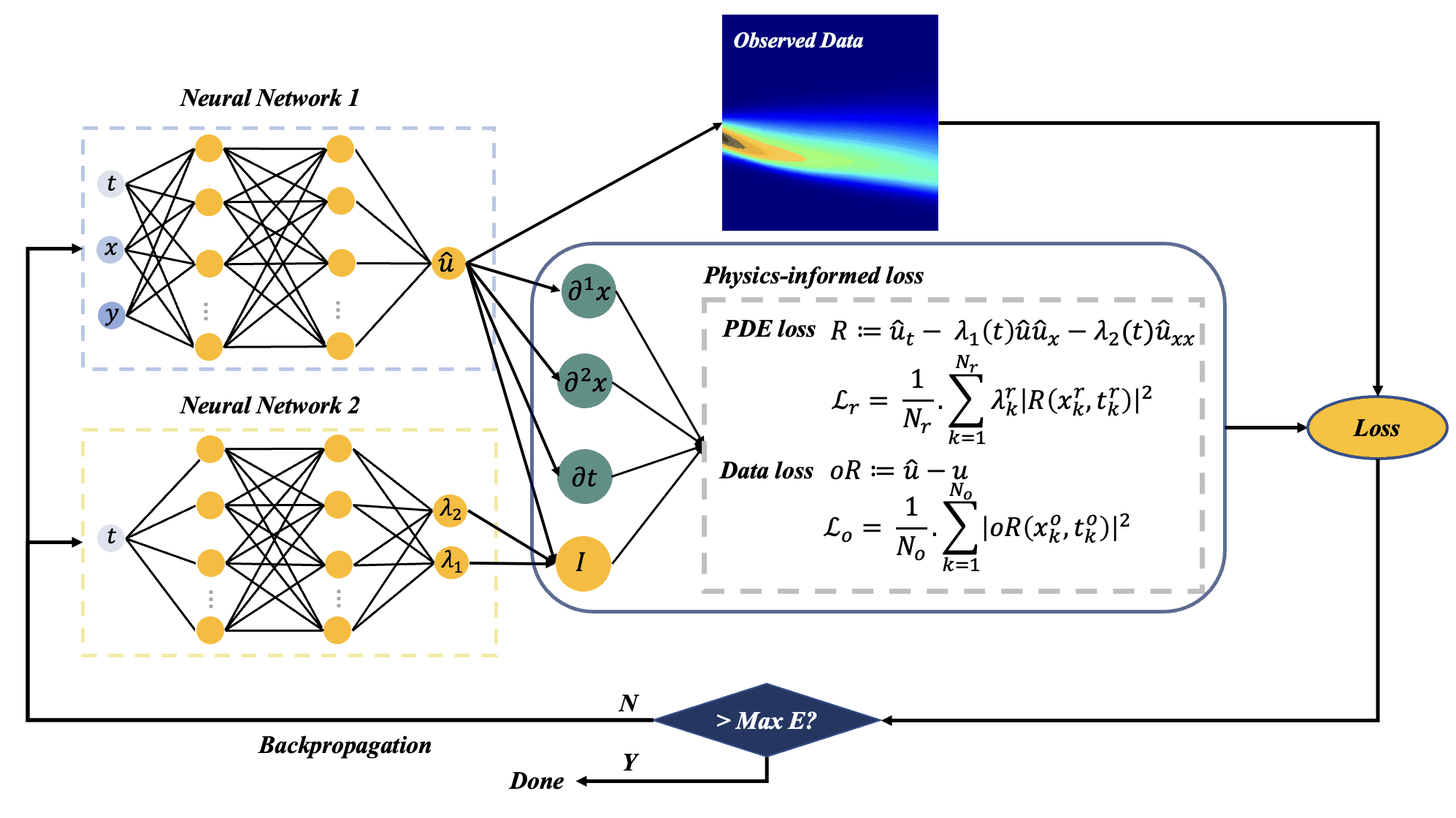}
		\caption{The schematics of the constrained self-adaptive PINNs' framework for the inverse problem. \label{fig::framework}}
	\end{figure}

	In this subsection, we introduce the modified cSPINNs to solve the inverse problem. We firstly consider the model problem given the spatial domain $\Omega$, and temporal domain $t \in[0, T]$, which with explicit parametric form of parameterized PDEs:
	\begin{align}
		&u_t[(x, t)]+\mathcal{N}[u(x, t);\lambda_p(t)]=0, \quad x \in \Omega, t \in(0, T], 
	\end{align}
	where $\mathcal{N}[\cdot]$ is an operator parameterized by physics parameter $\lambda_p(t)$, which includes any combination of linear and non-linear terms of spatial derivatives. To infer the unknown parameters of the PDE via PINNs \cite{raissi2019physics}, we need to construct a neural network $\hat{u}(x, t ; \boldsymbol{w})$ given the spatial $x \in \Omega$ and temporal $t \in[0, T]$ inputs with the trainable parameters $\boldsymbol{w}$ to fit the data $\{x_k^o, t_k^o, u_k^o\}_{k=1}^{N_o}$. Meanwhile, the neural network also needs to satisfy the physics laws, i.e. the parameterized governing PDE. Therefore, we can train a physics-informed model by minimizing the following loss function
	\begin{align}
		\mathcal{L}(\boldsymbol{w})= \lambda_r \mathcal{L}_r(\boldsymbol{w})+ \lambda_{o} \mathcal{L}_{o}(\boldsymbol{w}),
	\end{align}
	where
	\begin{subequations}
		\begin{align}
			& \mathcal{L}_r(\boldsymbol{w})=\frac{1}{N_r} \sum_{i=1}^{N_r}\left| \hat{u}_t[(x_{r}^i, t_{r}^i)]+\mathcal{N}[\hat{u}(x_{r}^i, t_{r}^i);\lambda_p(t_{r}^i)] \right|^2,\\
			&\mathcal{L}_{o}(\boldsymbol{w})=\frac{1}{N_{o}} \sum_{i=1}^{N_{o}}\left|\hat{u}\left({x}_{o}^i, {t}_{o}^i\right)-u_o\left({x}_{o}^i, {t}_{o}^i\right)\right|^2.
		\end{align}
	\end{subequations}
	Here, $\mathcal{L}_r$ and $\mathcal{L}_{o}$ are loss functions due to the residual in the PDE loss, data loss between observed data, and predicted value from the network. We use $\hat{u}$ to represent the output of the neural network or in other words, the PDE solution, which is parameterized by $\boldsymbol{w}$. The weights $\lambda_r$ and $\lambda_{o}$ could highly influence the convergence rate of different loss components and the final accuracy of PINNs \cite{mcclenny2020self}. Recently, many works \cite{mcclenny2020self,wight2020solving,Wang2022causal09,wang2021understanding,Revanth2022bcPINN} are proposed to explore the weighting strategy during PINNs training, which has become one of the mainstream directions of PINNs. To further enhance the learning ability in the physics domain with the complex solution and improve the accuracy of inferred parameters, we introduce a constrained self-adaptive weighting residual loss function. For the inverse problem, the training goal is determined by the residual loss and data loss, here we mainly consider the residual loss, which is closely related to the accuracy of inferred parameters. Then we first rewrite the residual loss function as
	\begin{subequations}
		\begin{align}
			& \mathcal{L}_r(\boldsymbol{w})=\frac{1}{N_r} \sum_{i=1}^{N_r} \hat{\lambda}_r^i \left| \hat{u}_t[(x_{r}^i, t_{r}^i)]+\mathcal{N}[\hat{u}(x_{r}^i, t_{r}^i);\lambda_p(t_{r}^i)] \right|^2,
		\end{align}
	\end{subequations}
	during training, we update the trainable weights $\{ \hat{\lambda}_r^i \}_{i=1}^{N_r}$ as 
	\begin{subequations}
		\begin{align}
			{\boldsymbol{\lambda}_r^{k+1}} &= {\boldsymbol{\hat{\lambda}}_r^k} +\eta_k \nabla_{ \hat{\boldsymbol{\lambda}}_r^k } \mathcal{L}\left(\boldsymbol{w}, {\lambda}_r, {\lambda}_{o}, \hat{\boldsymbol{\lambda}}_r^k \right),\\
			{\lambda}_{r_i}^{k+1} &= \frac{ |{\lambda}_{r_i}^{k+1}| }{\sum_{i=1}^{N_r} |{\lambda}_{r_i}^{k+1}| } \times C,\\
			\hat{{\lambda}}_{r_i}^{k+1} &= (1-\epsilon) \times \hat{{\lambda}}_{r_i}^{k} + \epsilon \times {{\lambda}}_{r_i}^{k+1},
		\end{align}
	\end{subequations}
	where we denotes $r_i$ as $i^{th}$ residual points in ${\{x_r^i, t_r^i\}}_{i=1}^{N_r}$, $k$ and $k+1$ the training iteration numbers. ${\boldsymbol{\lambda}_r^{k+1}}$ is a middle variable before normalization, in other words, we first normalize the ${\lambda}_{r_i}^{k+1} \in {\boldsymbol{\lambda}_r^{k+1}} $ and get the final $\hat{{\lambda}}_{r_i}^{k+1}$ by a weighted sum of the previous weight $\hat{{\lambda}}_{r_i}^{k}$ of iteration $k$ and the normalized  ${\lambda}_{r_i}^{k+1}$ in the current $k+1$ iteration. We set $C$ as the expectation of weights in PINNs here, i.e., we let $C= E(\sum_{i=1}^{N_r} \hat{{\lambda}_r}_i) = N_r$. We update the weights by gradient ascend here to raise PINNs' attention in the area that is difficult to learn. Figure \ref{fig::framework} illustrates the modified constrained self-adaptive PINNs framework for parameter identification problems. The Neural Network 1 is used to approximate the solution $\hat{u}$, and the Neural Network 2 which is the adding sub-network we mentioned above is applied to reconstruct the varying parameters $\lambda_1(t)$ and $\lambda_2(t)$. Training loss is composed of the modified PDE loss and the data loss, which correspond to the physics laws and the real observed data, respectively. Here, we obtained the observed data by numerically solving the time-dependent Burgers' equation as in \cite{Rudy2018Identification}, which depends on a spectral method and uses the specfem2D package to simulate the wave equation. It is worth noting that we consider the physical parameters of the system to evolve, which could be modeled using a neural network with time as input and predicted parameters as output. Readers could see the neural network structure in Figure \ref{fig::framework} for more details.

	\subsection{Change Point Detection by Finite Mixture Method}
	For the data of the system varying parameters $\lambda_1(t)$ and $\lambda_2(t)$ obtained from modified cSPINNs, we need to find a suitable way to perform change-point detection work for system \eqref{2.2}. In general, due to the observation noise and the biased estimation of training the network, we have
	\begin{equation}\label{3.6}
		\Pr(y_t|\mathbf{X})=\mathcal{N}(\mathrm{E}(y_t),\sigma^2),
	\end{equation}
	where $y_t \in \mathrm{Y}=\{y_1,y_2,\cdots,y_N\}$ is a biased estimate of the parameter $\lambda(t)$ at discrete-time points $\{t_1,t_2,\cdots,t_N\}^{T}_{0}$, $\mathbf{X}$ is the system output spatiotemporal data $u(x,t)$ and $\mathcal{N}$ is 1D Gaussian distribution with density function
	\begin{equation}
		f_{\mathcal{N}}(y;\, \mu, \sigma^2)=\frac{1}{\sigma \sqrt{2 \pi}}\mathrm{exp}{\left\{-\frac{(y-\mu)^2}{2 \sigma^2}\right\}}.
	\end{equation}
	Due to the time-varying parameters in system \eqref{2.2}, the probabilistic model \eqref{3.6} is extended to a Gaussian mixture model (GMM) for observations
	\begin{equation}
		\Pr(y|\vartheta)=\sum_{k=1}^K \alpha_k f_{\mathcal{N}}(y;\, \mu_k, \sigma_k^2),
	\end{equation}
	where $\vartheta=\{\mu_k,\sigma_k,\alpha_k,\,k=1,2,\cdots,K\}$ is model unidentified parameters with proportional factor $\sum_{k=1}^K \alpha_k=1$. Define the latent variable $D_{tk}$ is a $0/1$ encoding of the assignment of the observation $y_t$ to $k$ subgroup of the mixture model.
	\begin{equation}
		\gamma_{tk}=
		\begin{cases}
			1, \quad & \text{$y_t$ is from $k$ subgroup}, \\
			0, \quad & \text{otherwise},
		\end{cases}
	\end{equation}
	with its responsive estimation 
	\begin{equation}
		\hat{\gamma}_{tk} = \mathrm{E}(\gamma_{tk}|\mathrm{Y},\vartheta)=\frac{\alpha_k f_{\mathcal{N}}(y_t;\, \mu_k, \sigma_k^2)}{\sum_{k=1}^K \alpha_k f_{\mathcal{N}}(y_t;\, \mu_k, \sigma_k^2)}.
	\end{equation}
	The expectation step uses the parameter estimations of the model from the previous step to calculate the conditional expectation of the log-likelihood function for the observation data
	\begin{equation}
		\mathrm{E} [\mathrm{log}\Pr (y,\gamma | \mathrm{Y} ,\vartheta)]=\sum_{k=1}^{K}\Big\{ (\mathrm{log}\,\alpha_k)\sum_{t=1}^N \gamma_{tk} + \Big[ \mathrm{log} (1/\sqrt{2\pi})- \mathrm{log}\,\sigma_k - \frac{(y_t-\mu_k)^2}{2\sigma^2_k} \Big] \sum_{t=1}^N \hat{\gamma}_{tk}  \Big\}.
	\end{equation}
	The maximization step determines the parameters $\hat{\vartheta}_j^{\{m-1\}}$ for maximizing the log-likelihood function of the complete data obtained in the expectation step
	\begin{equation}
		\hat{\vartheta}^{\mathrm{new}}= \arg\mathop{\max}\limits_{\vartheta}\mathrm{E} [\mathrm{log}\Pr (y,\gamma | \mathrm{Y} ,\vartheta)].
	\end{equation}
	By Lagrange constrained optimization method, the updates of model parameters in each iteration are
	\begin{equation}
		\{\hat{\mu}_k,(\hat{\sigma}_k)^2,\hat{\eta}_k\}^{\mathrm{new}}=\Big\{\frac{\sum_{t=1}^N \hat{\gamma}_{tk} y_j} {\sum_{t=1}^N \hat{\gamma}_{tk}},\frac {\sum_{t=1}^N \hat{\gamma}_{tk} (y_j-\mu_k)^2} {\sum_{j=1}^N \hat{\gamma}_{tk}},\frac{\sum_{t=1}^{N} \hat{\gamma}_{tk}} {N}\Big\}.
	\end{equation}
	Then in continuous iterations, until the algorithm converges, the final two-state GMM parameter estimates
	\begin{equation}
		\hat{\vartheta}=\{\hat{\mu}_k,\hat{\sigma}_k,\hat{\eta}_k,\:,k=1,2,\cdots,K\},
	\end{equation}
	are generated.
	Thus the soft classification probability results based on GMM of observations $\mathrm{Y}$ can be obtained as a $ N \times K $ matrix
	\begin{equation}
		\mathbf{G} = \{ g_{i,j} \}_{1 \leq i \leq N,\, 1 \leq j \leq K},
	\end{equation}
	where 
	\begin{equation}
		g_{i,j} = \frac{\eta_j f_{\mathcal{N}}(y_i; \mu_j, \sigma_j)}{ \sum_{k=1}^K \eta_k f_{\mathcal{N}}(y_i; \mu_k, \sigma_k)},
	\end{equation}
	which is deduced from the Bayes theorem, and it reveals the magnitude of the probability that the $i$-th sample belongs to the $j$-th mixture component of the GMM model.
	Hence for the observation data $\mathrm{Y}=\{y_1,y_2,\cdots,y_N\}$. For 1D Burgers' equation with time-varying parameters, we can calculate a corresponding sequence of change-point probabilities in time interval $[t-1,t+1]$
	\begin{equation}
		\mathbf{P}^{\mathrm{change}}=\Big\{p_t = 1- \sum_{k=1}^K (g_{t-1,k} \cdot g_{t,k} \cdot g_{t+1,k}), \:  2 \leq t \leq N-1 \Big\}.
	\end{equation}
	For a 2D space-varying wave equation with a space-varying parameter $\alpha$, we need to consider the Gaussian distribution in a high dimension
	\begin{equation}
		f_{\mathcal{N}}(\mathbf{x}; \boldsymbol{\mu},\boldsymbol{\Sigma})=\sum^K_{k=1} \frac{1}{(2\pi)^{\frac{d}{2}} {\lvert \boldsymbol{\Sigma} \rvert}^{\frac{1}{2}}} \exp\Big\{-{\frac{1}{2}(\mathbf{x}-\boldsymbol{\mu})^{\mathrm{T}}\,\boldsymbol{\Sigma}^{-1}\,(\mathbf{x}-\boldsymbol{\mu})}\Big\},
	\end{equation}
	Similarly, after getting the two-dimensional GMM, we have the soft classification probability results for space point $(x,y)$ in the domain
	\begin{equation}
		g_{x,y,j} = \frac{\eta_j f_{\mathcal{N}}((x,y); \boldsymbol{\mu}_j,\boldsymbol{\Sigma}_j)}{ \sum_{k=1}^K \eta_k f_{\mathcal{N}}((x,y); \boldsymbol{\mu}_k,\boldsymbol{\Sigma}_k)},
	\end{equation}
	Then we give a similar calculation of change-point probabilities in a cross-shaped five-point region $\{(x,y), (x-1,y), (x+1,y), (x,y-1), (x,y+1)\}$
	\begin{equation}
		\begin{split}
			&\mathbf{P}^{\mathrm{change}}=\Big\{p_{x,y} = 1 - \sum_{k=1}^K (g_{x,y,j} \cdot g_{x-1,y,j} \cdot g_{x+1,y,j} \cdot g_{x,y-1,j} \cdot g_{x,y+1,j}) , \:  2 \leq x \leq N_x-1, \: 2 \leq y \leq N_y-1 \Big\}.
		\end{split}
	\end{equation}
	Finally, the peaks of this time series could be regarded as the detected state change-points of systems \eqref{2.2} and \eqref{2.5} in global time.
	
	\section{1D Burgers' Equation with Time-varying Parameter}\label{forth_sec}
	In this section, we use three distinctive types of numerical cases to test the performance of our framework. Moreover, those three category cases represent different evolutionary models of the time-varying 1D parametric Burgers' equation, and their hidden state transition paths can be discovered via our framework.
	
	To better identify parameter $\lambda_1(t)$, a sub-network with the input $t$ and the output $\lambda_1(t; \phi)$ is used to model the dynamics of the parameter, where $\phi$ denotes all trainable parameters of the network and could be optimized during training with the time-varying parameters $\lambda_{1}$ and $\lambda_{2}$. The loss function is denoted as 
	
	(a). Mean squared error on the observed data
	\begin{equation}
		\mathrm{MSE}_o=\frac{1}{N_o} \sum_{k=1}^{N_o}\left(\hat{u}\left(\boldsymbol{x}_k^o, t_k^o\right)-u_k^o\right)^2, \quad\left(\boldsymbol{x}_k^o, t_k^o\right) \in \Omega \times T.
	\end{equation}
	
	(b). Mean squared error of the residual points
	\begin{equation}
		\begin{aligned}
			\mathrm{MSE}_{R}&=\frac{1}{N_r} \sum_{k=1}^{N_r}\left(R\left(\boldsymbol{x}_k^r, t_k^r\right)\right)^2, \quad\left(\boldsymbol{x}_k^r, t_k^r\right) \in \Omega \times T, \\
			R :&= \hat{u}_t-\lambda_1(t) \hat{u} \hat{u}_x-\lambda_2(t) \hat{u}_{x x}. 
		\end{aligned}
	\end{equation}
	
	(c). Total mean squared error for inverse
	\begin{equation}
		\mathrm{MSE} = \lambda_o \mathrm{MSE}_o + \lambda_{R} \mathrm{MSE}_{R}.
	\end{equation}
	
	In the following numerical experiments, we will get $N_o = 4,000$ observed data, and $N_r = 64,000$ residual points randomly sampled from the computational domain with $\Omega = [-8,8], T = 10$. We let the weights of the PDE loss term and the residual loss term $\lambda_o = \lambda_{R} = 1$. We use the modified multilayer perceptron (MLP) \cite{wang2021understanding} with a depth of 6, a width of 128, and the tahn activation function as the Neural Network 1 for solving the inverse problem. As for Neural Network 2, a modified MLP is used here, which has 1 input neuron and consists of 4 hidden layers with 40 neurons in each layer, and the activation function is chosen as tanh. The Adam optimizer is used here to minimize the loss function with $N_e = 200,000$ epochs. Meanwhile, We set the batch size of residual points $N_{bs} = 4,000$ to reduce the memory requirement of hardware. The initial learning rate is 0.001, and the exponential learning rate annealing method is applied here with hyper-parameter $\gamma = 0.9$ during training. The total time-domain $[0,10]$ of the parametric Burgers' equation has been discretized into 256 times steps uniformly. To identify the parameters $\lambda_{1}$ and $\lambda_{2}$, all the observed data within five steps segment has been chosen. Prediction errors of identifying parameters via modified cSPINNs are shown in the appendix \ref{Error} while the statistical inferring results and the $L^2$ error for learning Burgers' equation are shown in the table \ref{Appendix Table}. Next, we start to exhibit our results. 
	
	\subsection{Case 1: Burgers' Equation with Single Change Point}
\begin{figure}[H]
		\centering
		\subfigure{
			\begin{minipage}{0.32\textwidth}
				\centering
				\includegraphics[width=1\textwidth]{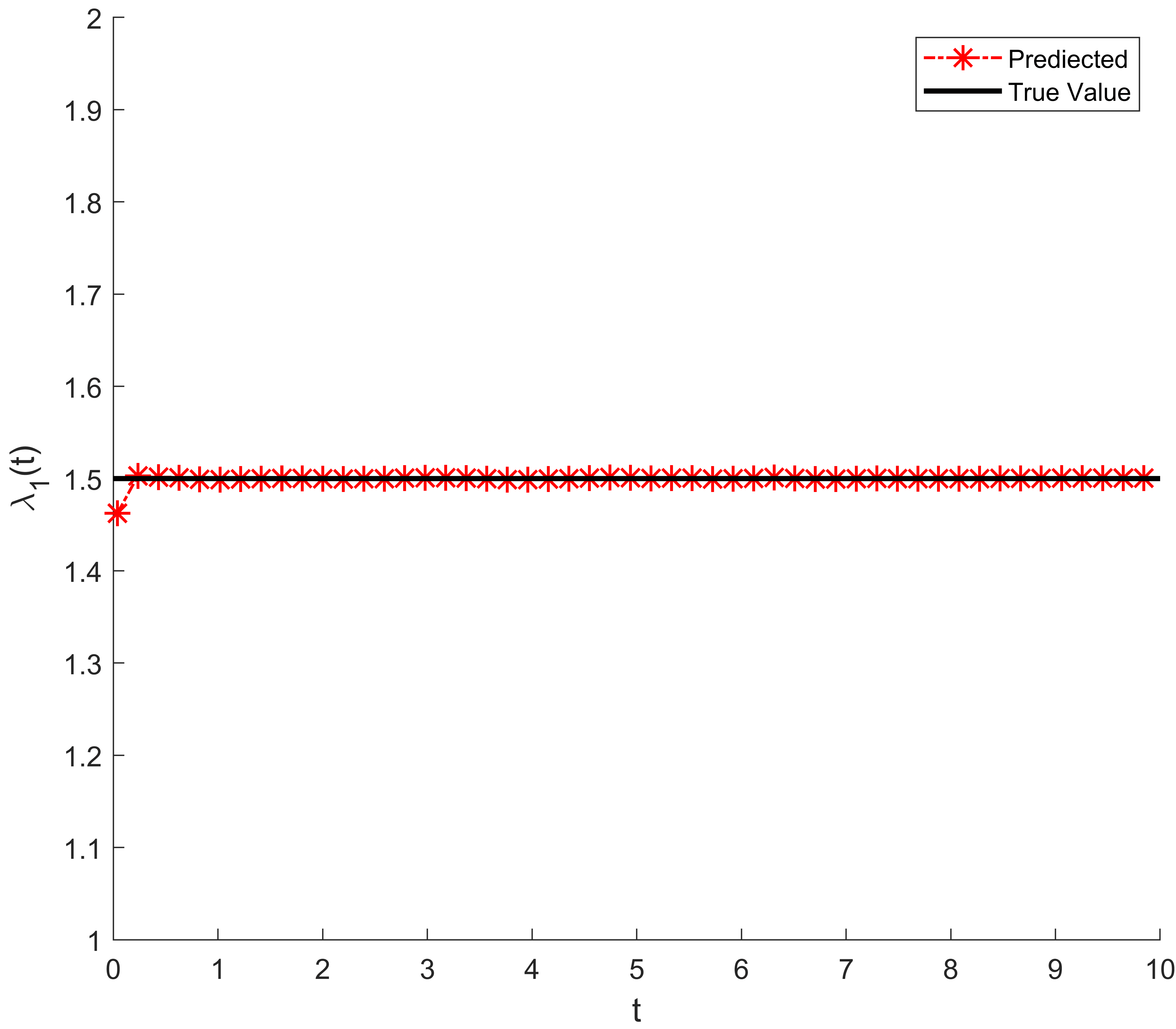}
			\end{minipage}
			\begin{minipage}{0.32\textwidth}
				\centering
				\includegraphics[width=1\textwidth]{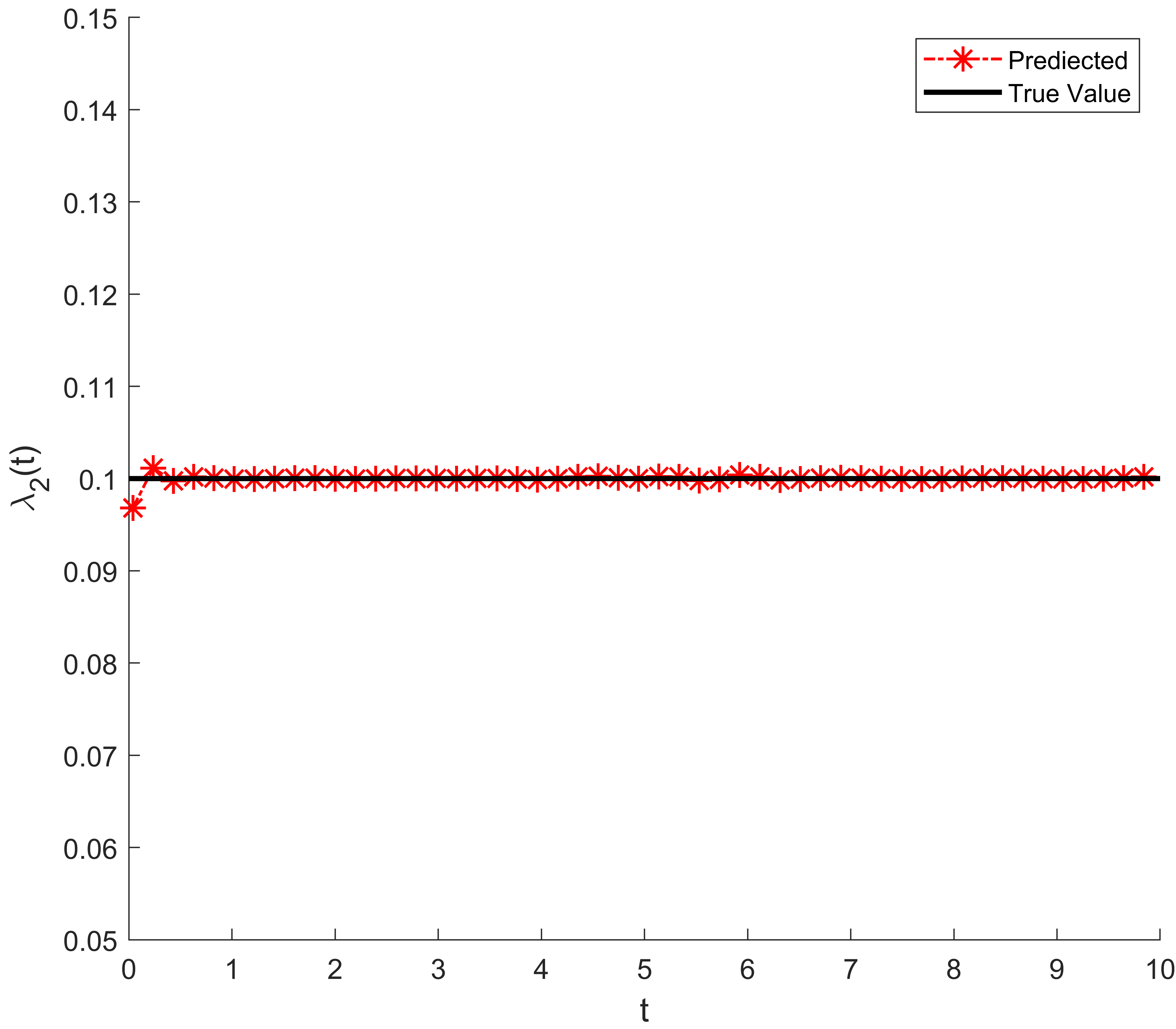}
			\end{minipage}
			\begin{minipage}{0.32\textwidth}
				\centering
				\includegraphics[width=1\textwidth]{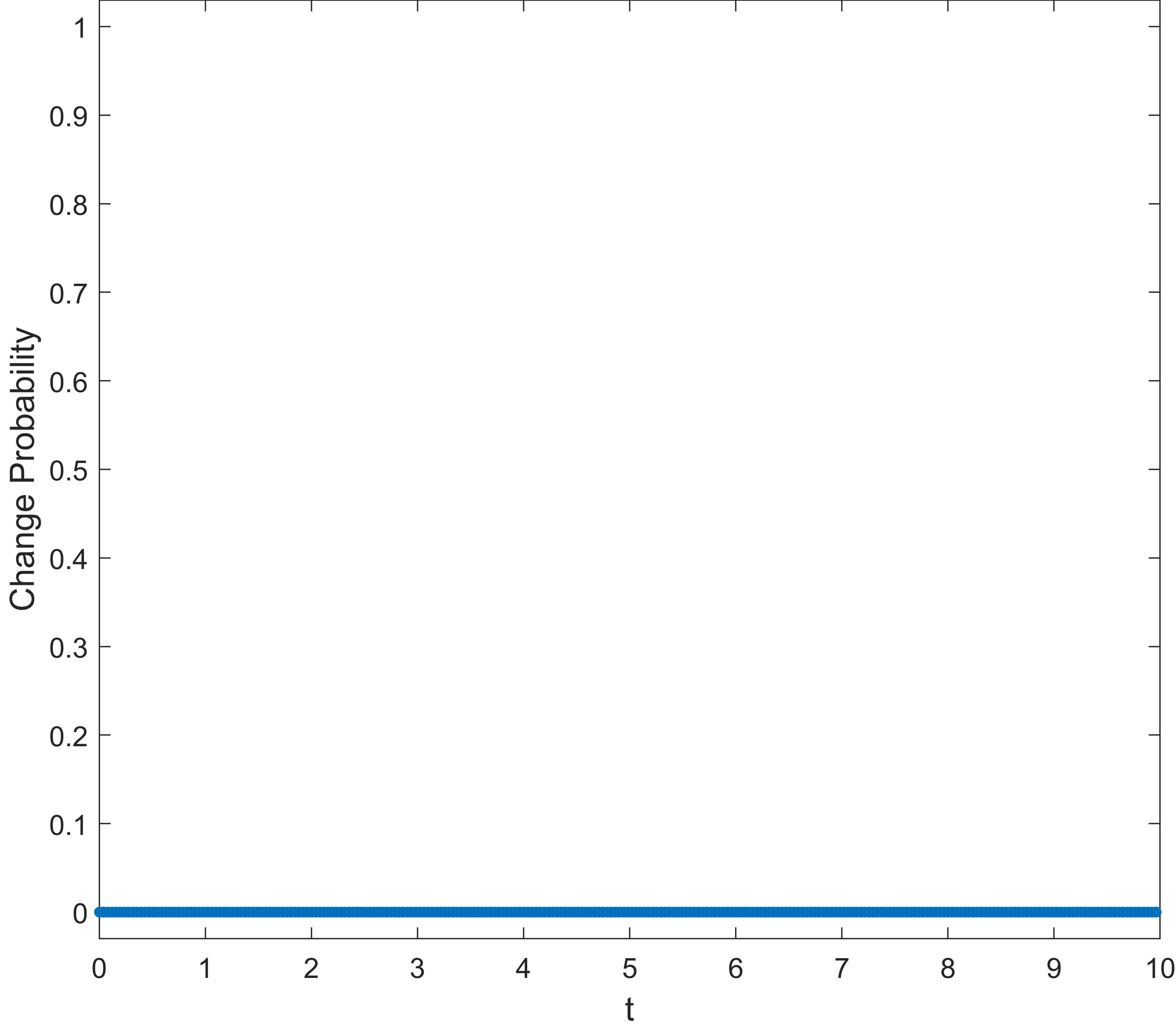}
			\end{minipage}
		}
		\subfigure{
			\begin{minipage}{0.32\textwidth}
				\centering
				\includegraphics[width=1\textwidth]{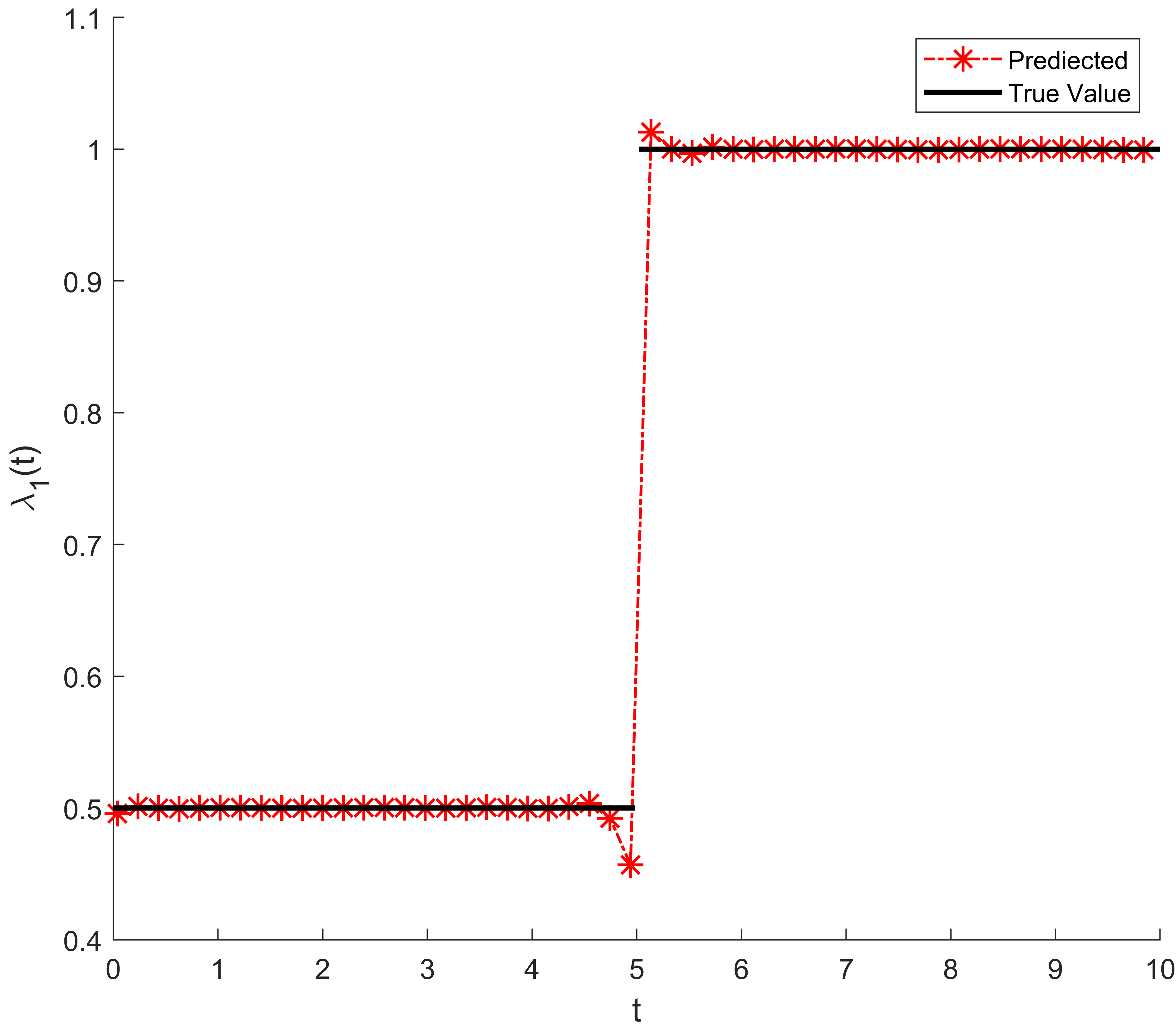}
			\end{minipage}
			\begin{minipage}{0.32\textwidth}
				\centering
				\includegraphics[width=1\textwidth]{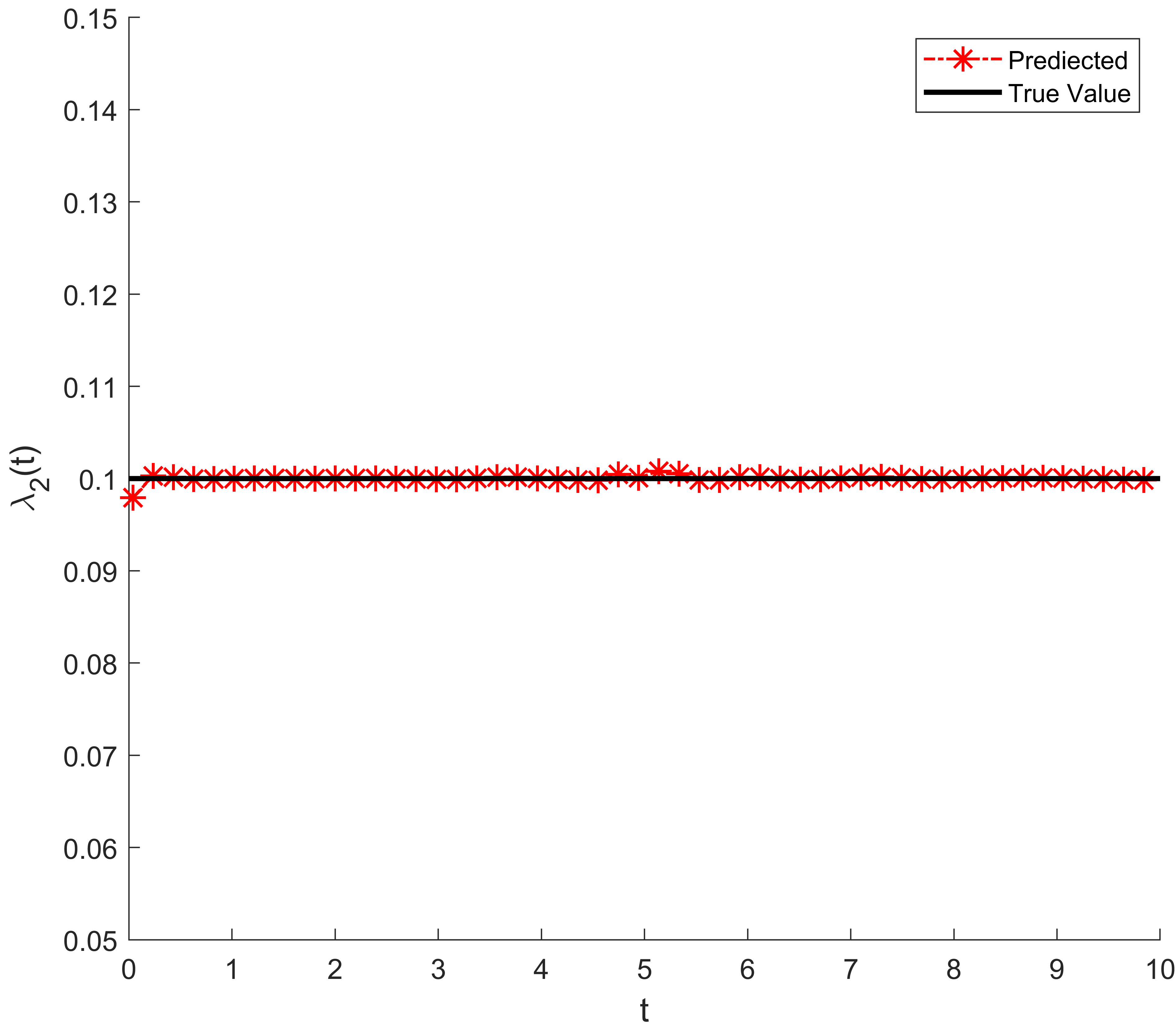}
			\end{minipage}
			\begin{minipage}{0.32\textwidth}
				\centering
				\includegraphics[width=1\textwidth]{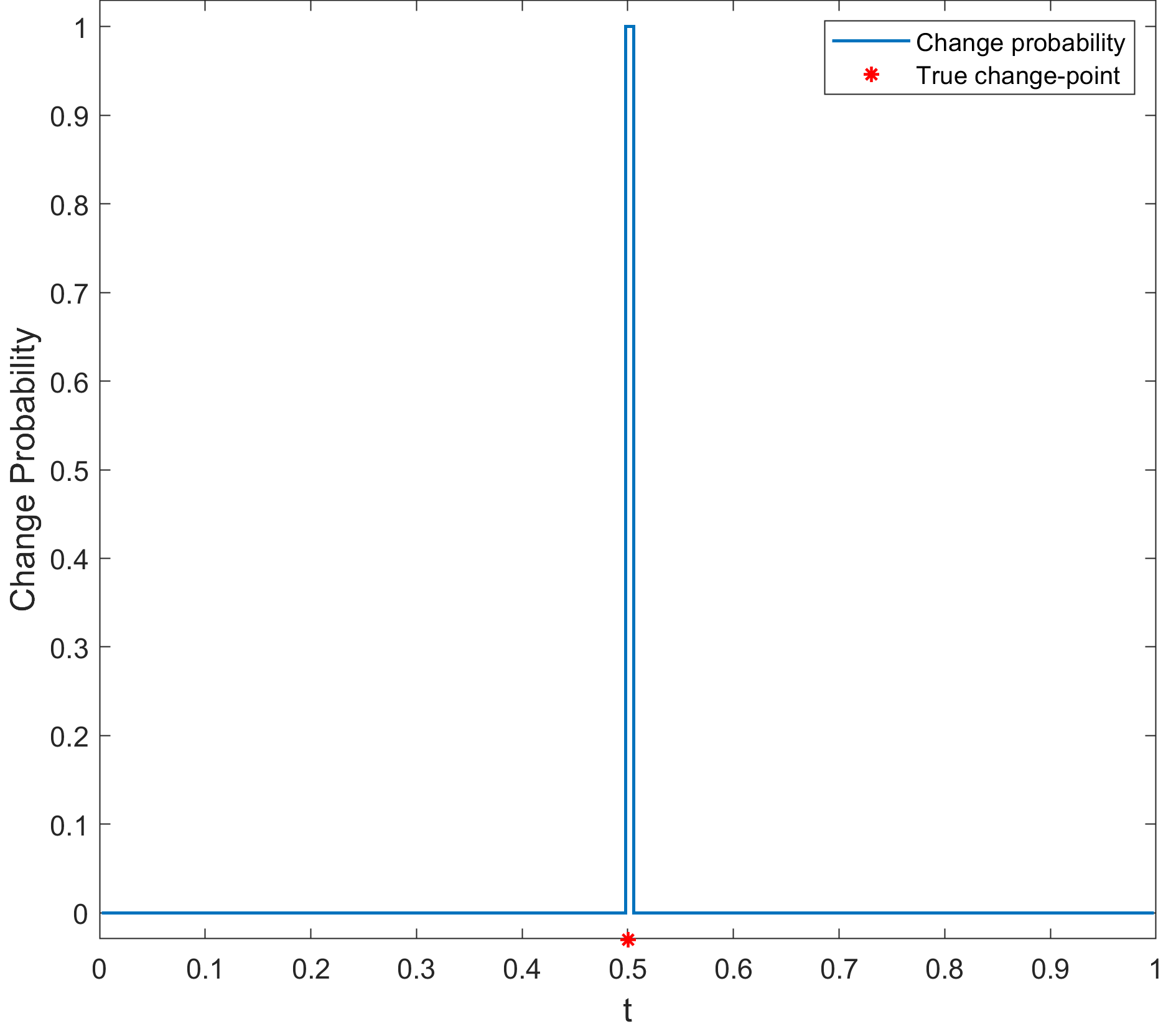}
			\end{minipage}
		}
		\subfigure{
			\begin{minipage}{0.32\textwidth}
				\centering
				\includegraphics[width=1\textwidth]{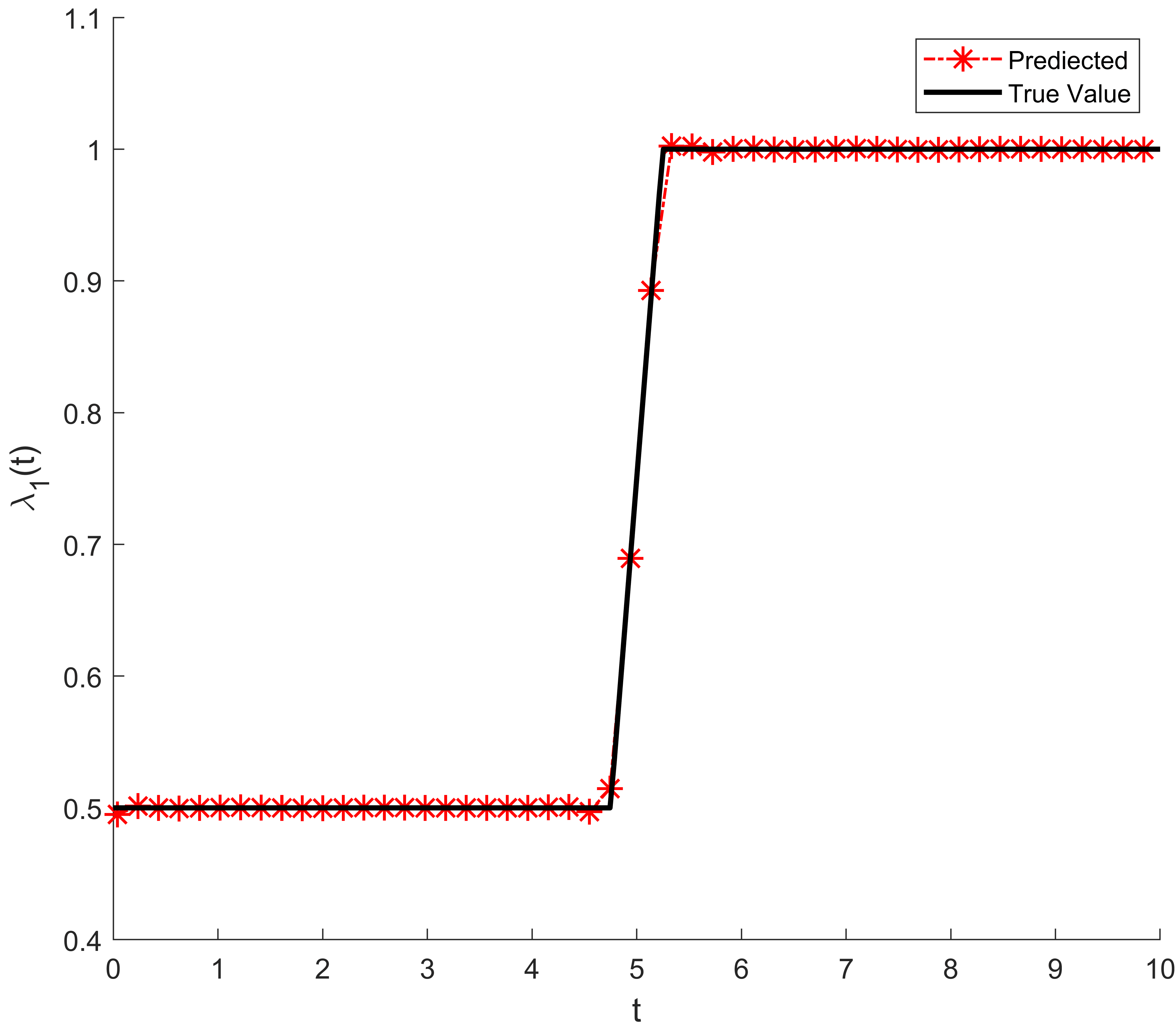}
			\end{minipage}
			\begin{minipage}{0.32\textwidth}
				\centering
				\includegraphics[width=1\textwidth]{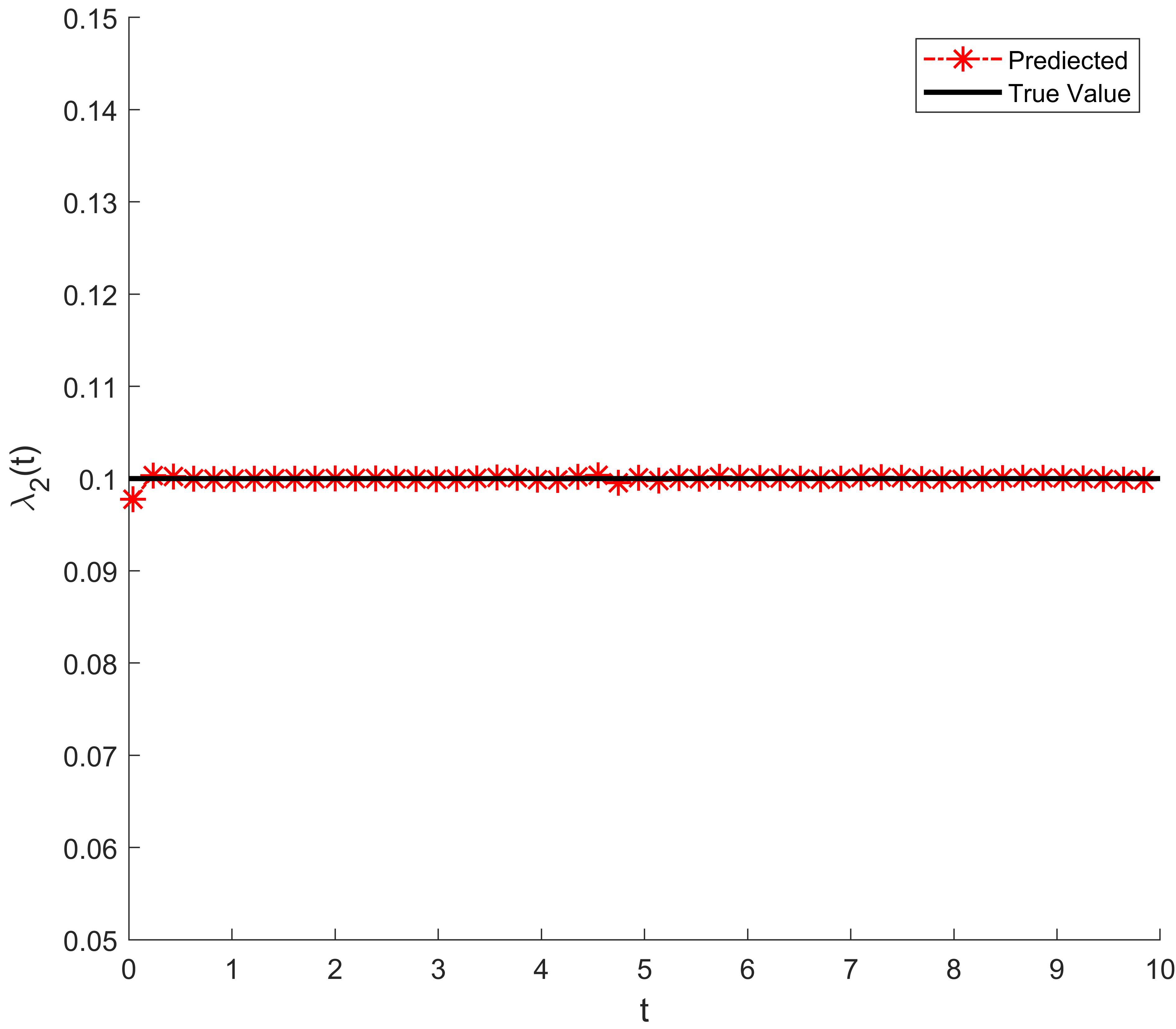}
			\end{minipage}
			\begin{minipage}{0.32\textwidth}
				\centering
				\includegraphics[width=1\textwidth]{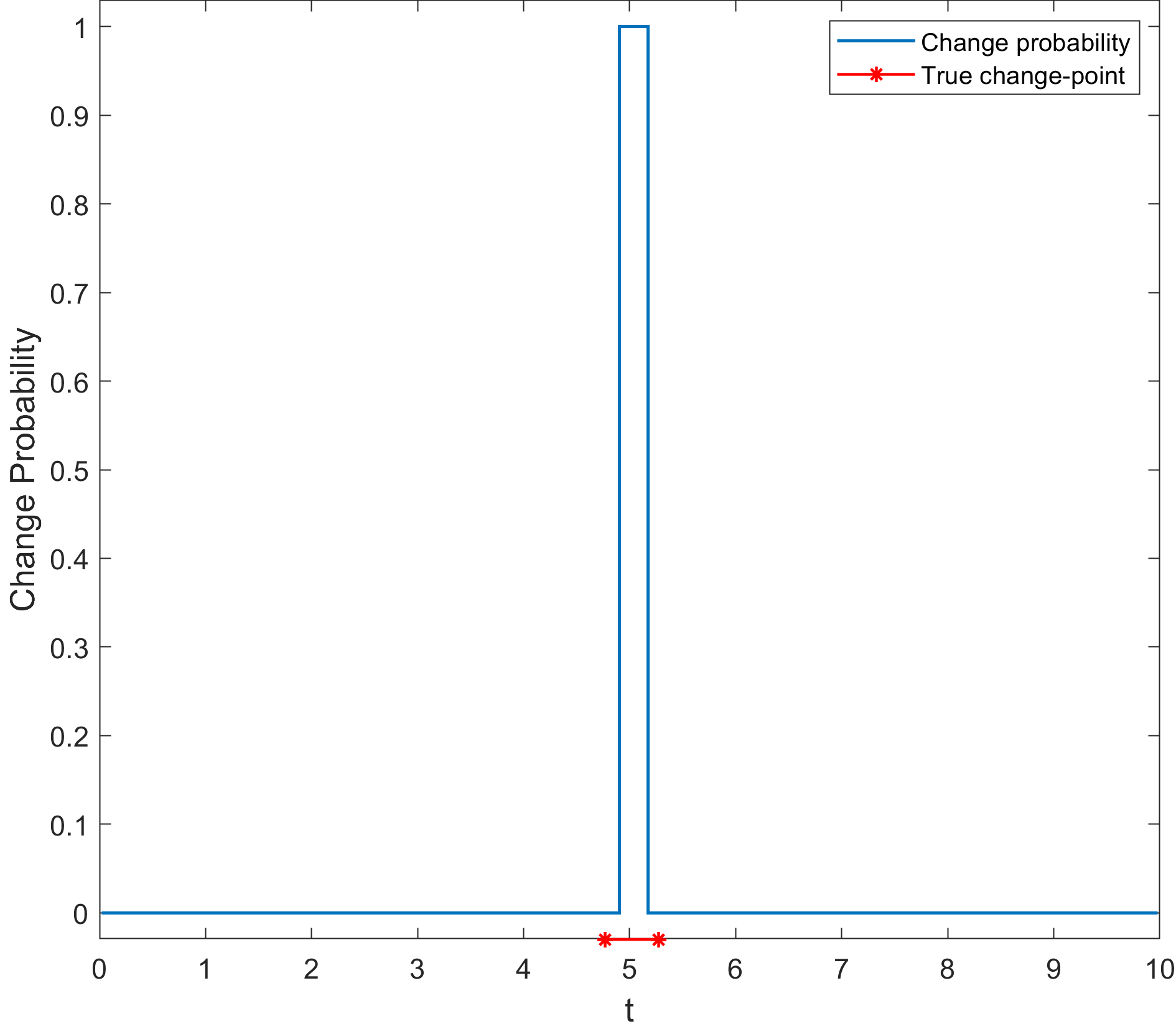}
			\end{minipage}
		}
		\caption{\label{Result 1} Figures from the first and second columns represent $\lambda_{1}$ and $\lambda_{2}$ learned using the modified cSPINNs approach for the time-varying parametric Burgers' equation. Figures from the last column illustrate change point detection results of $\lambda_{1}$ learned. The first, second, and third rows correspond with cases 1.1, 1.2, and 1.3.}
	\end{figure}

	The fundamental evolutionary model for a parametric PDE-governed time-varying system necessitates that one time-varying parameter contains one change point throughout the entire process. In this study, we explore three conditions: the first is the trivial case with no change point; the second and third are cases that feature a single varying parameter with one abrupt shift or one gradual change, respectively.
	\begin{equation}
		\begin{split}
			&{\rm case \: 1.1}: \lambda_1(t)= 1.5.\quad {\rm case 1.2}: \lambda_1(t)= \begin{cases} 0.5, & 0 \leq t<5. \\ 1, & 5 \leq t \leq 10. \\ \end{cases} \quad\\ 
			&{\rm case \: 1.3}: \lambda_1(t)= \begin{cases}0.5, & 0 \leq t< 4.77, \\ 0.98x-4.19 & 4.77 \leq t<5.27, \\ 1, & 5.27 \leq t \leq 10. \end{cases} 
		\end{split}
	\end{equation}
	Follow the proposed framework mentioned in Section \ref{Third_sec}, 
	we apply modified cSPINNs with a sub-network to learn the time-varying parameter $\lambda_{1}$ of the parametric Burgers' equation, then we detect the change points by a finite mixture model. Through the results attained above, the transition path could be discovered. Figure \ref{Result 1} shows the time-varying parameter values obtained using modified cSPINNs and the results of our change point detection scheme. Sub-figures in the first column and the second column illustrate values of $\lambda_{1}$ learned and $\lambda_{2}$ learned using modified cSPINNs, separately. And the last row of sub-figures is the results of the finite mixture model. It demonstrates that our framework performs well for all three cases. The main advantage of our framework is that we can discover the transition path of a time-varying system governed by a parametric PDE without any prior information. More specifically, we can predict the values of parameters(constant or time-varying) and the locations of change points without any prior information about time segments. 
	
	From figure \ref{Result 1}, we can observe that the predicted parameters accurately fit the reference solution for both constant and time-varying cases where the predicted errors mainly appear at the location with discontinuity. Moreover, the error of case 1.2 shown in the second row with an abrupt change is larger than case 1.3 which the time-varying parameter $\lambda_{1}$ evolves gradually. This phenomenon seems reasonable since it is always hard for PINNs to tackle problems with discontinuities \cite{jagtap2022physics}. To better identify the change points, we prefer to use the probability method to finish our change point detection task. Our criterion of detection is measured by probability through the finite mixture model. It successfully captures the same change point in case 1.2 as the reference solution which has properties of low variance and high confidence. In this way, we managed to find out all the change points in the evolutionary process in case 1.3.
	
	The total discretized time points for all experiments is 256 such that we may get 255 probability results through the change point detection method. For better analysis, we set a threshold as 1e-6. The results show that there exists one change point at $t=5$ for case 1.2 and 9 change points for case 1.3. Thus the transition path of ($\lambda_{1}$,$\lambda_{2}$) for cases above are $(1.5,0.1)\rightarrow(1.5,0.1)$, $(0.5,0.1)\rightarrow(1,0.1)$ and $(0.5,0.1)\rightarrow(1,0.1)$ with sequential gradual change points.
	
	\subsection{Case 2: One Time-varying Parameter with Multiple Change Points}
\begin{figure}[ht]
		\centering
		\subfigure{
			\begin{minipage}{0.32\textwidth}
				\centering
				\includegraphics[width=1\textwidth]{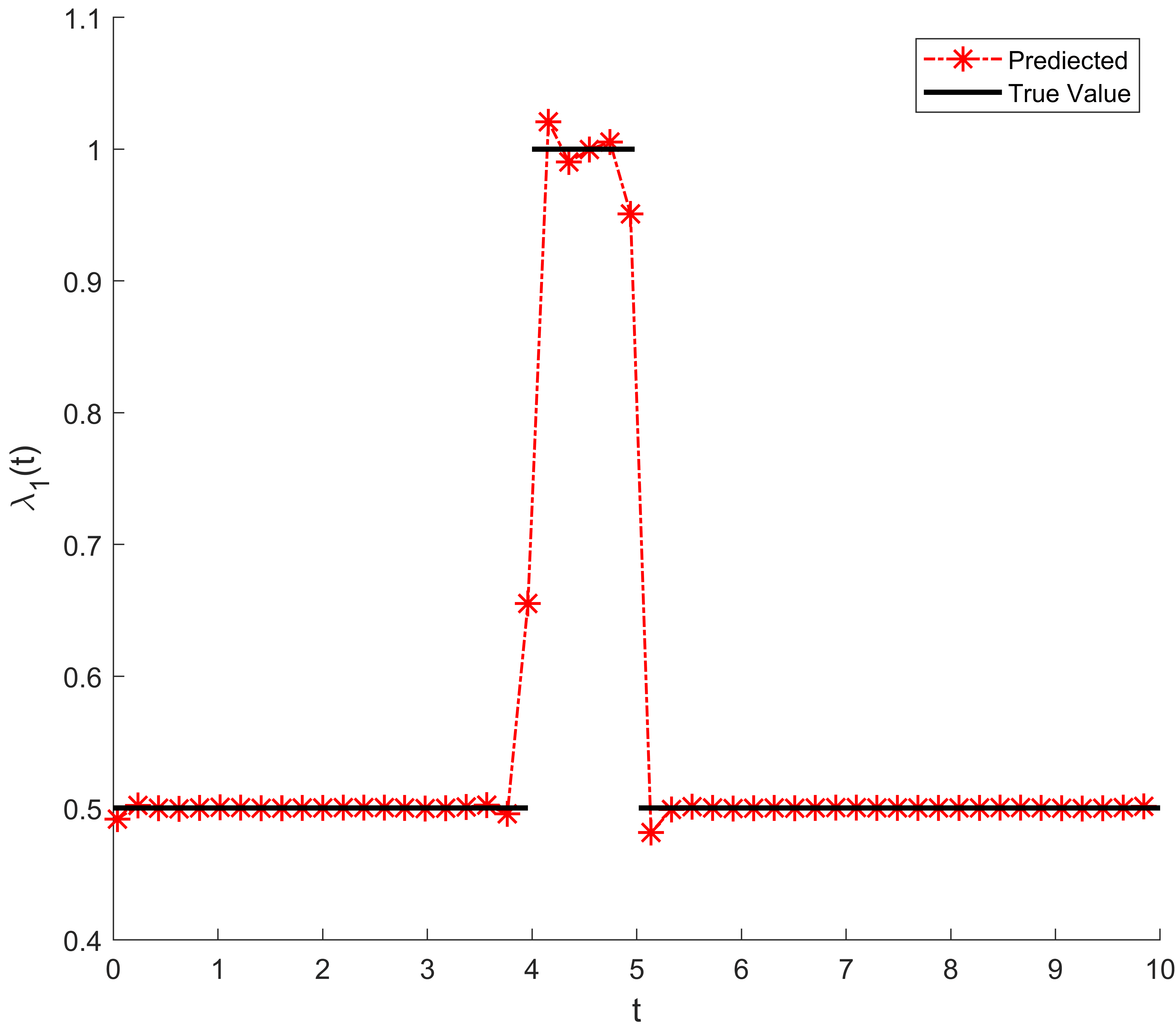}
			\end{minipage}
			\begin{minipage}{0.32\textwidth}
				\centering
				\includegraphics[width=1\textwidth]{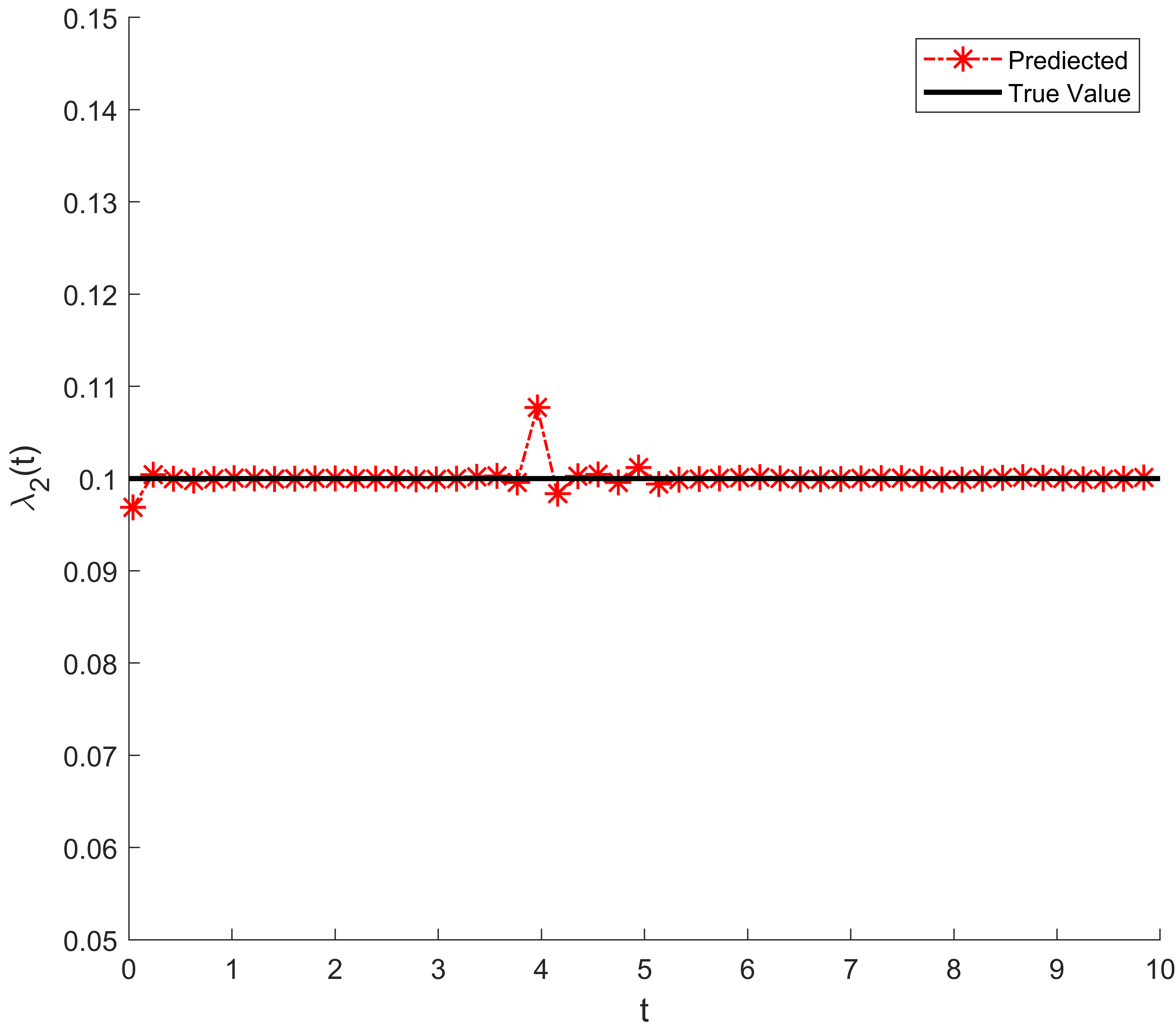}
			\end{minipage}
			\begin{minipage}{0.32\textwidth}
				\centering
				\includegraphics[width=1\textwidth]{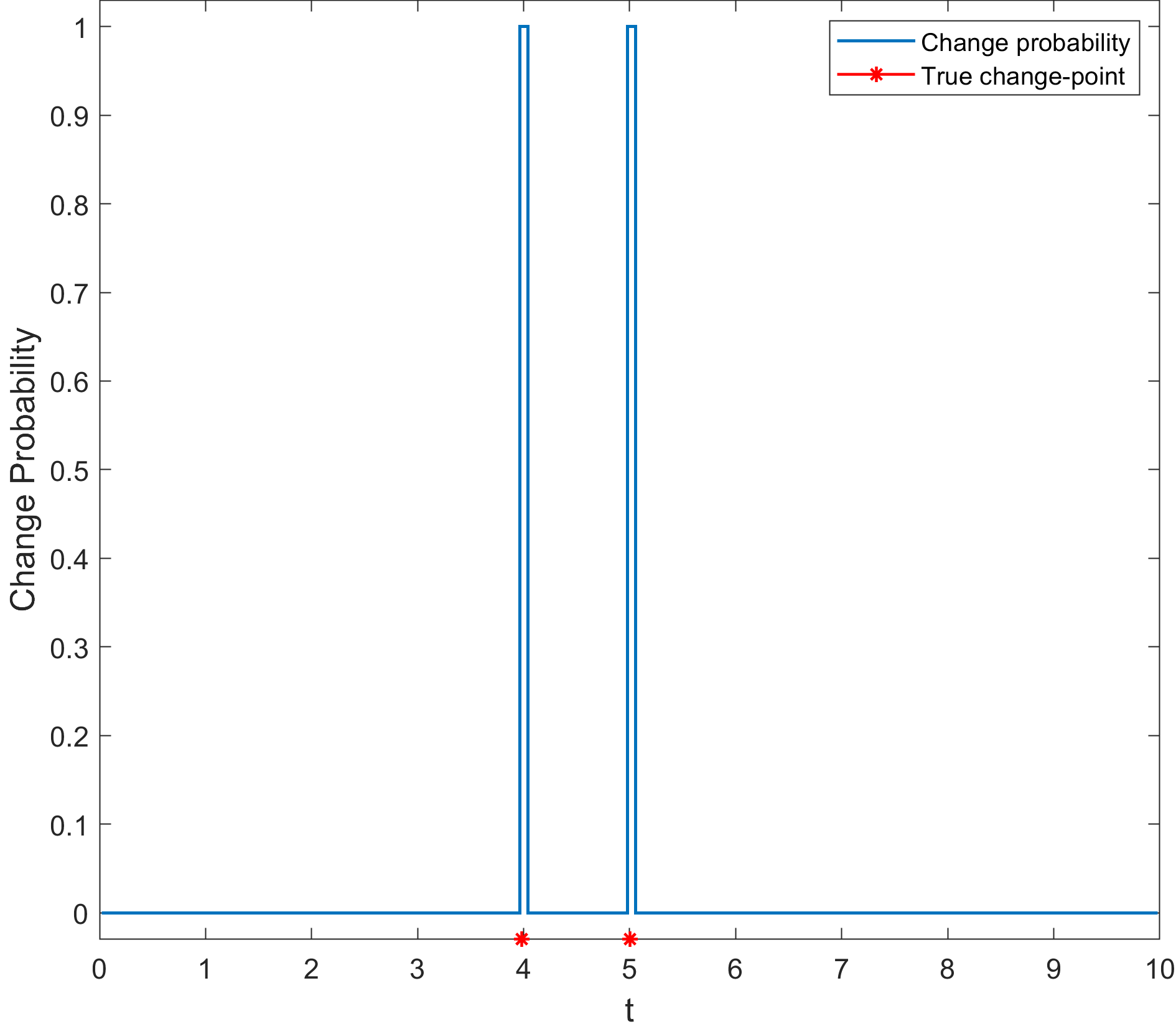}
			\end{minipage}
		}
		\subfigure{
			\begin{minipage}{0.32\textwidth}
				\centering
				\includegraphics[width=1\textwidth]{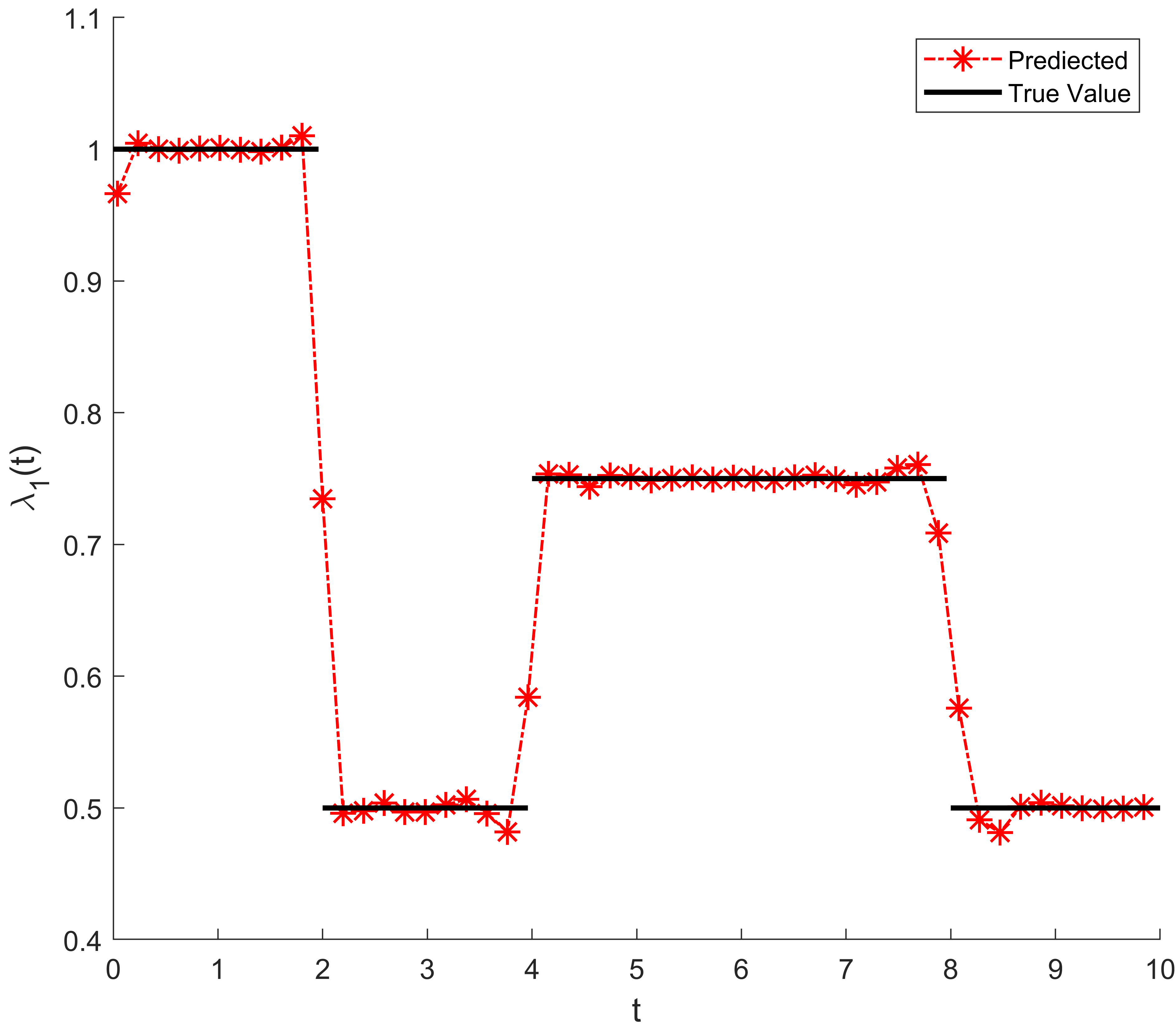}
			\end{minipage}
			\begin{minipage}{0.32\textwidth}
				\centering
				\includegraphics[width=1\textwidth]{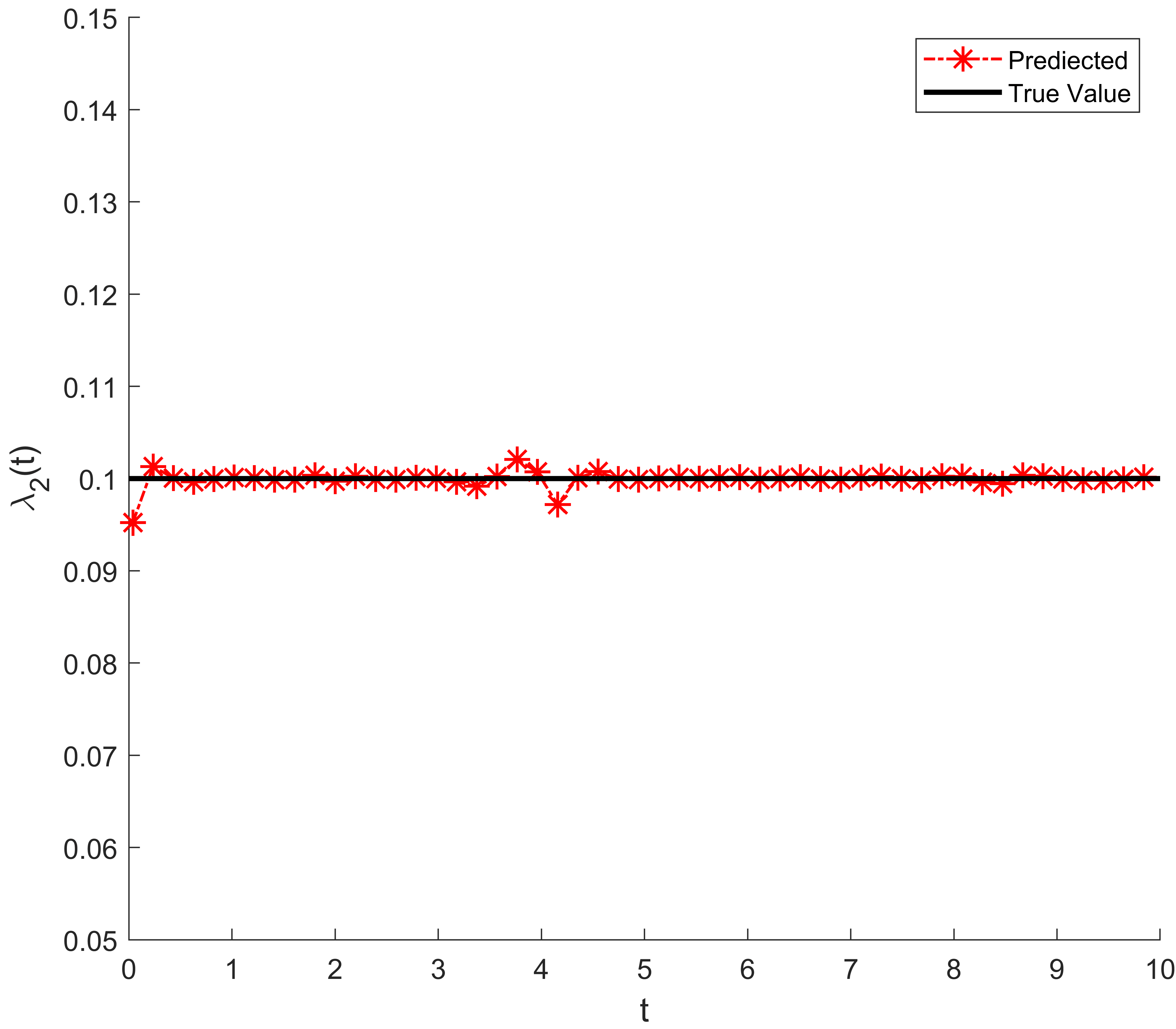}
			\end{minipage}
			\begin{minipage}{0.32\textwidth}
				\centering
				\includegraphics[width=1\textwidth]{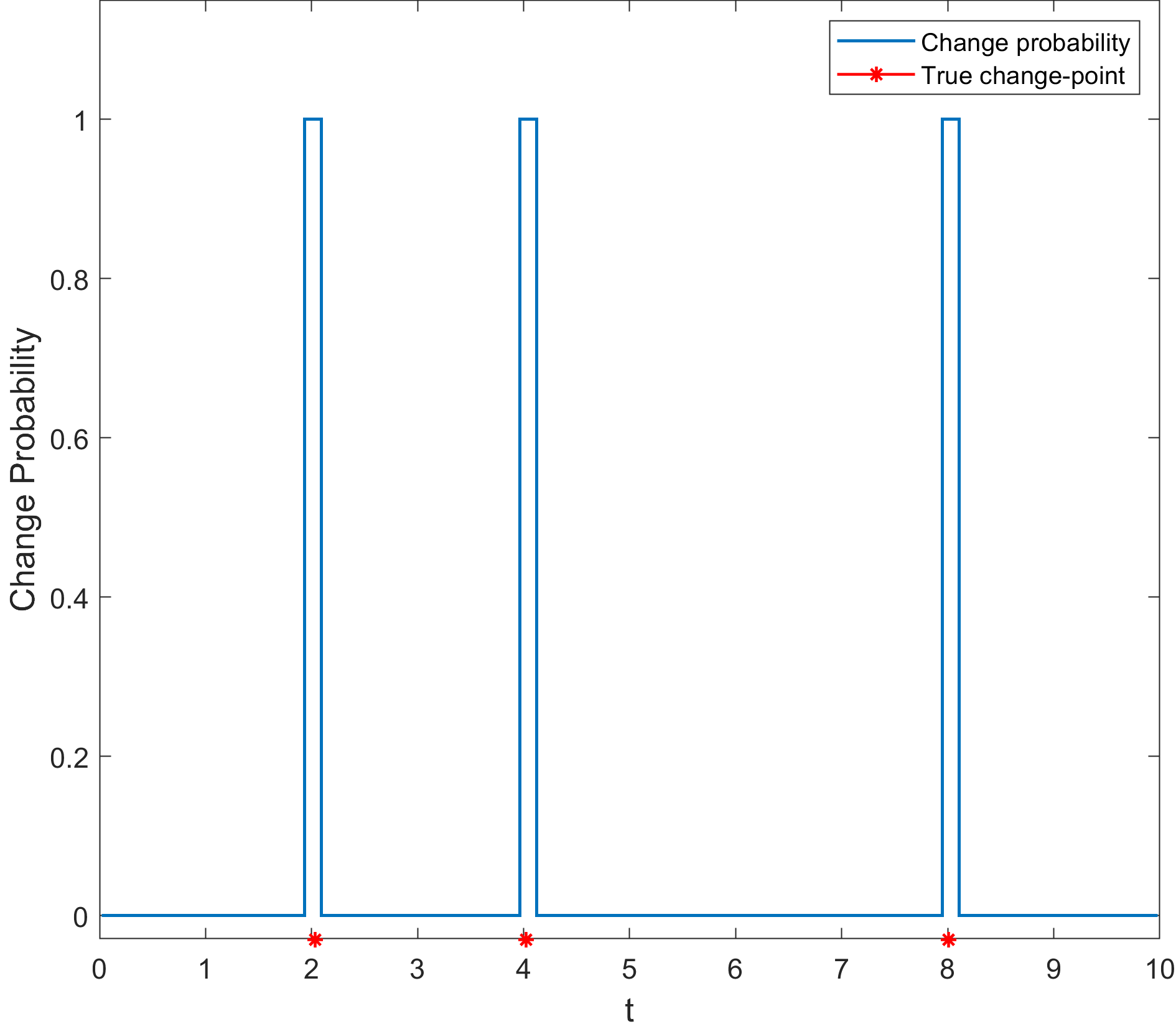}
			\end{minipage}
		}
		\caption{\label{Result 2}Figures from the first and second columns represent $\lambda_{1}$ and $\lambda_{2}$ learned results. And the last column illustrates the change point detection results of the time-varying parameter $\lambda_{1}$. The first and second rows correspond with cases 2.1 and 2.2.}
	\end{figure}

	The second type of evolutionary model is one time-varying parameter with multiple change points. Here, the time-varying parameter is $\lambda_{1}$ and another parameter $\lambda_{2}$ is constantly 0.1. In this scenario, we test the performance of our framework through two cases. The first is the time-varying parameter $\lambda_{1}$ takes two values with two change points, and the second is the time-varying parameter $\lambda_{1}$ takes two values with three change points. The reference solution has been obtained as follows: 
	\begin{equation}
		\begin{split}
			{\rm case\: 2.1}:\lambda_1(t)= \begin{cases}0.5, & 0 \leq t< 4, \\ 1, & 4 \leq t< 5,\\ 0.5, & 5 \leq t \leq 10. \end{cases}\quad
			{\rm case\: 2.2}:\lambda_1(t)= \begin{cases}1, & 0 \leq t< 2, \\ 0.5, & 2 \leq t< 4,\\ 0.75, & 4 \leq t< 8,\\ 0.5, & 8 \leq t \leq 10.\\\end{cases}
		\end{split}
	\end{equation}
	Figures \ref{Result 2} illustrate results in the same way as the cases discussed above and errors mainly locate at positions where the discontinuity occurs. For all three cases, our framework successfully identifies the time-varying parameter $\lambda_{1}$ precisely and captures all change points which are consistent with the reference solution. In this way, the transition path has been discovered. 
	
	For case 2.1, the time-varying parameter $\lambda_{1}$ has the same state for the beginning and the end while mixing with a small ratio of the difference in the middle. Based on the results of modified cSPINNs, the change point detection method detects the two change points precisely. And our framework also performs well on case 2.2, a more complex three-state mixing time-varying system with three change points.
	As we mentioned above, the transition path of ($\lambda_{1}$,$\lambda_{2}$) for those two cases are $(0.5,0.1)\rightarrow(1,0.1)\rightarrow(0.5,0.1)$ and $(1,0.1)\rightarrow(0.5,0.1)\rightarrow(0.75,0.1)\rightarrow(0.5,0.1)$.
	
	\subsection{Case 3: Multiple Time-varying Parameters with Multiple Change Points}
	
	This type of case describes a more complicated time-varying system with multiple time-varying parameters and multiple change points. More precisely, a mixing time-varying 1D parametric Burgers' Equation with multiple change points. And the time-varying parameters $\lambda_{1}$ and $\lambda_{2}$ vary simultaneously in different paths. The reference solution of this case has been calculated as follows:
	\begin{equation}
		{\rm case \: 3}:\Big(\lambda_1(t),\lambda_2(t)\Big)= \begin{cases}(1.00,1.00), & 0 \leq t<2, \\ (0.75,1.33), & 2 \leq t<4, \\ (0.50,2.00), & 4 \leq t<6, \\ (0.75,1.33) & 6 \leq t<8, \\ (1.00,1.00), & 8 \leq t \leq 10.\end{cases} \quad
	\end{equation}
	The results of modified cSPINNs fit the reference solution well and the detection method successfully captures all four change points within the evolutionary process. In this case, the transition path of $(\lambda_{1},\lambda_{2})$ is $(1,1)\rightarrow(0.75,1.33)\rightarrow(0.5,2)\rightarrow(0.75,1.33)\rightarrow(1,1)$. The values of parameters represent the corresponding phases of the system.
	
	\begin{figure}[H]
		\centering
		\begin{minipage}{0.32\textwidth}
			\centering
			\includegraphics[width=1\textwidth]{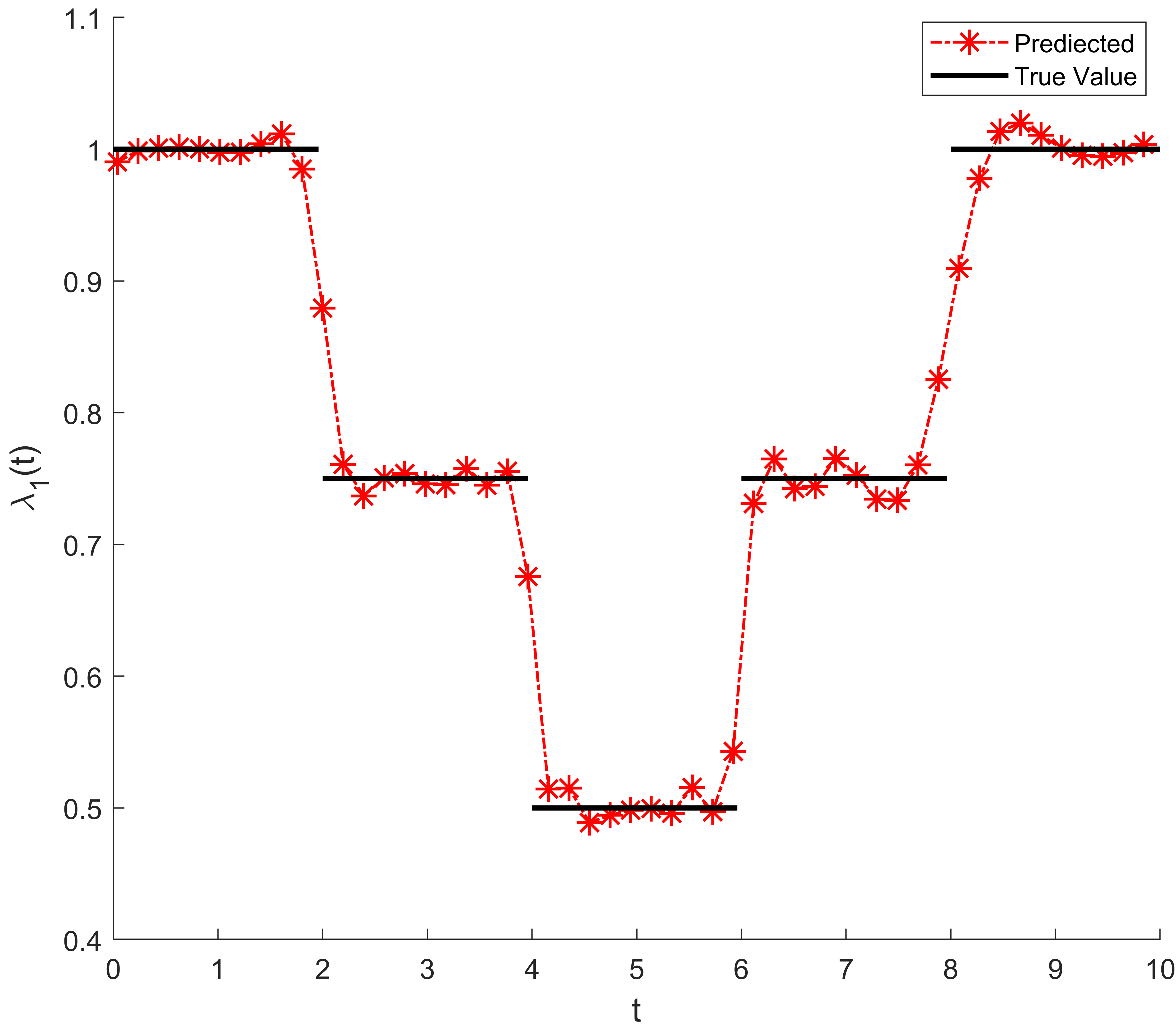}
		\end{minipage}
		\begin{minipage}{0.32\textwidth}
			\centering
			\includegraphics[width=1\textwidth]{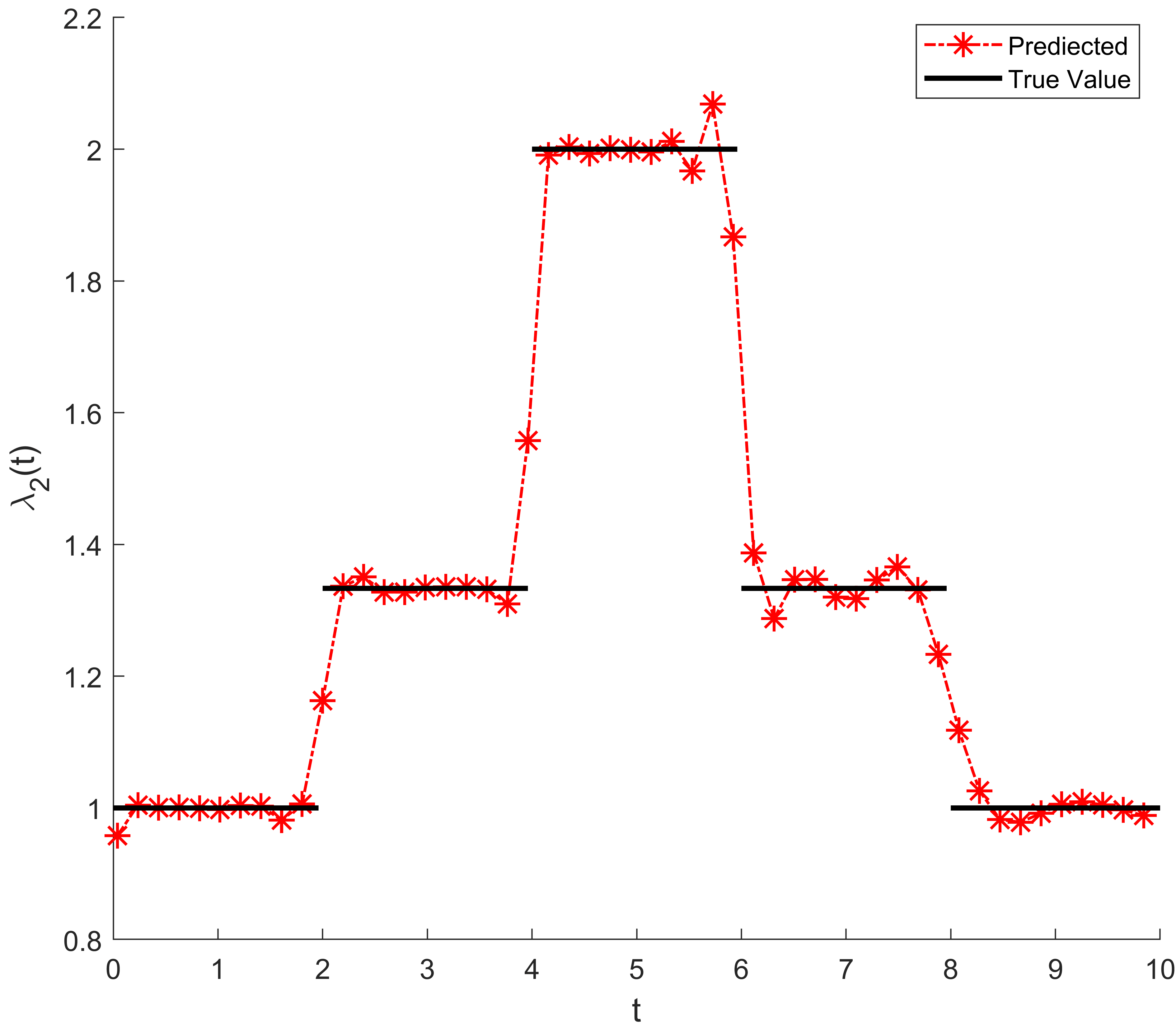}
		\end{minipage}
		\subfigure{
			\begin{minipage}{0.32\textwidth}
				\centering
				\includegraphics[width=1\textwidth]{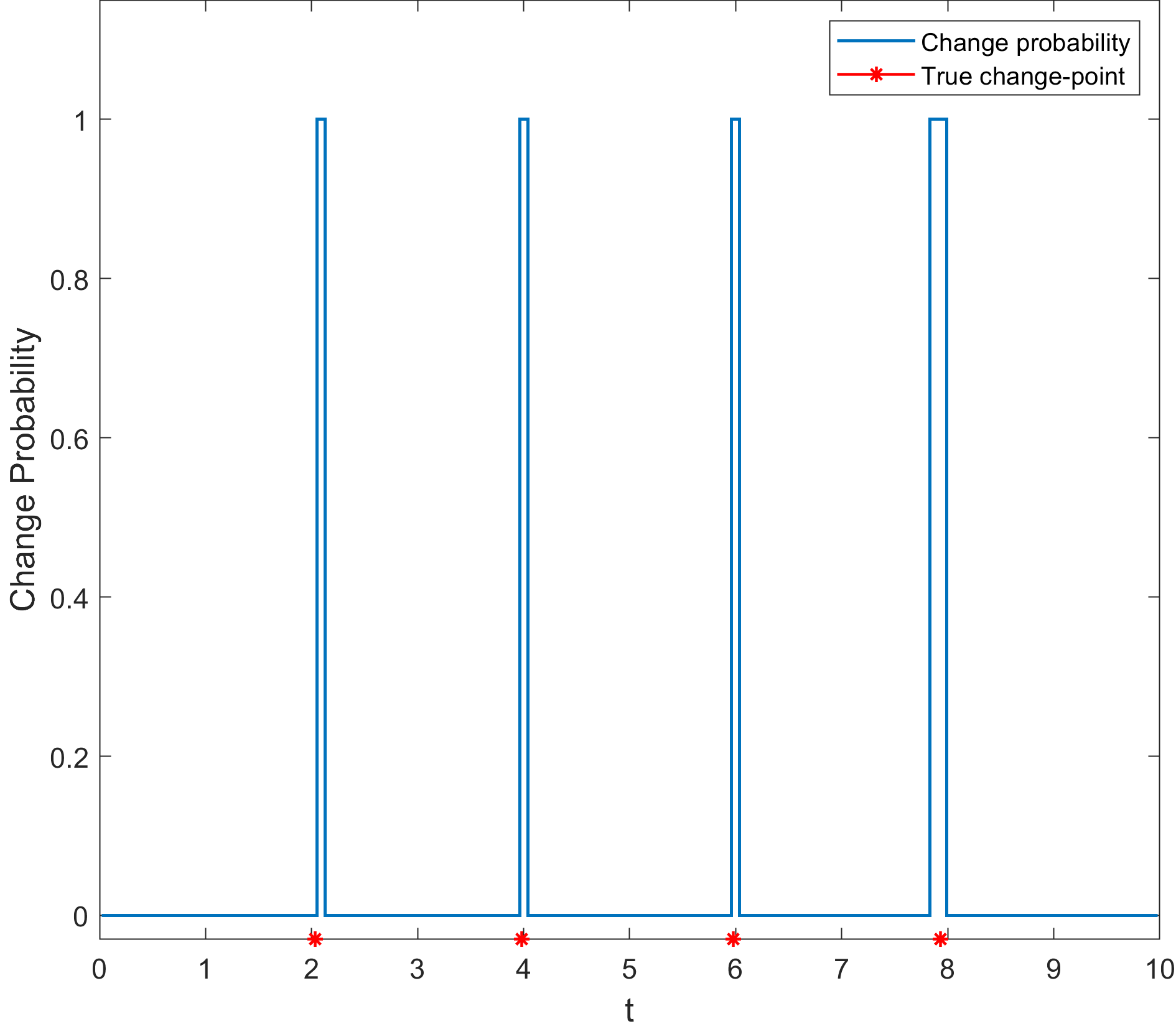}
			\end{minipage}
		}
		\caption{\label{Result 3} The first and second figure shows the $\lambda_{1}$, $\lambda_{2}$ learned results of the case 3. And the last picture shows the result of the change point detection method.}
	\end{figure}
	
	\section{Comparison with Existing Methods}\label{fifth_sec}
	
	In this section, we compare the proposed methods with traditional approaches for change-point detection and existing neural network models. The aim is to assess the effectiveness and advantages of our proposed techniques in addressing the respective research problems. By examining these comparisons, we can gain insights into the performance improvements and novel features offered by our proposed methods.
	
	\subsection{Comparison of Change-point Detection by Finite Mixture Method with Traditional Approach}
	
	Traditional research focuses on the consistency and convergence rates of CUSUM-type estimators for detecting change points in the mean of dependent observations \cite{kokoszka1998change}. The results obtained in this study hold under weak assumptions on the dependence structure, allowing for non-linear and non-stationary sequences. The consistency of CUSUM-type estimators is proven for detecting shifts in the mean of a sequence of observations, and the rates of convergence are derived. The analysis considers a broad range of dependence structures, making the findings applicable to various scenarios.
	The estimator of change points is defined as 
	\begin{equation}
		\hat{k}_{n}(\alpha)=\underset{1 \leq k \leq n-1}{\operatorname{argmax}}\left|U_{k}(\alpha)\right|,
	\end{equation}
	where
	\begin{equation}
		U_{k}(\alpha)=\left(\frac{k(n-k)}{n}\right)^{1-\alpha}\left(\frac{1}{k} \sum_{i=1}^{k} X_{i}-\frac{1}{n-k} \sum_{i=k+1}^{n} X_{i}\right), \quad 1 \leq k \leq n-1.
	\end{equation}
	
	Our tool enables the comprehensive detection of all four change points in a sequence, encompassing their precise positions and distinctive attributes. Conversely, traditional methods are limited to identifying solely the final change point, which is 0.8242s. Thus failing to capture the other change points. This discrepancy arises from the sequence's limited length, which impairs the accuracy of change point detection using conventional methods. Traditional approaches heavily rely on specific statistical models and assumptions to facilitate change point detection. However, in shorter sequences, these methods often struggle to identify early change points. This limitation stems from the constrained sensitivity and accuracy of traditional approaches when confronted with shorter sequences. In contrast, our tool employs a flexible and adaptive approach to detect change points, effectively adjusting to the data's unique features and patterns. By leveraging additional information, it accurately determines the presence and characteristics of change points, granting our tool superior detection capabilities even in shorter sequences.

	\begin{figure}[ht]
		\centering
		\subfigure{
			\begin{minipage}{0.42\textwidth}
				\centering
				\includegraphics[width=1\textwidth]{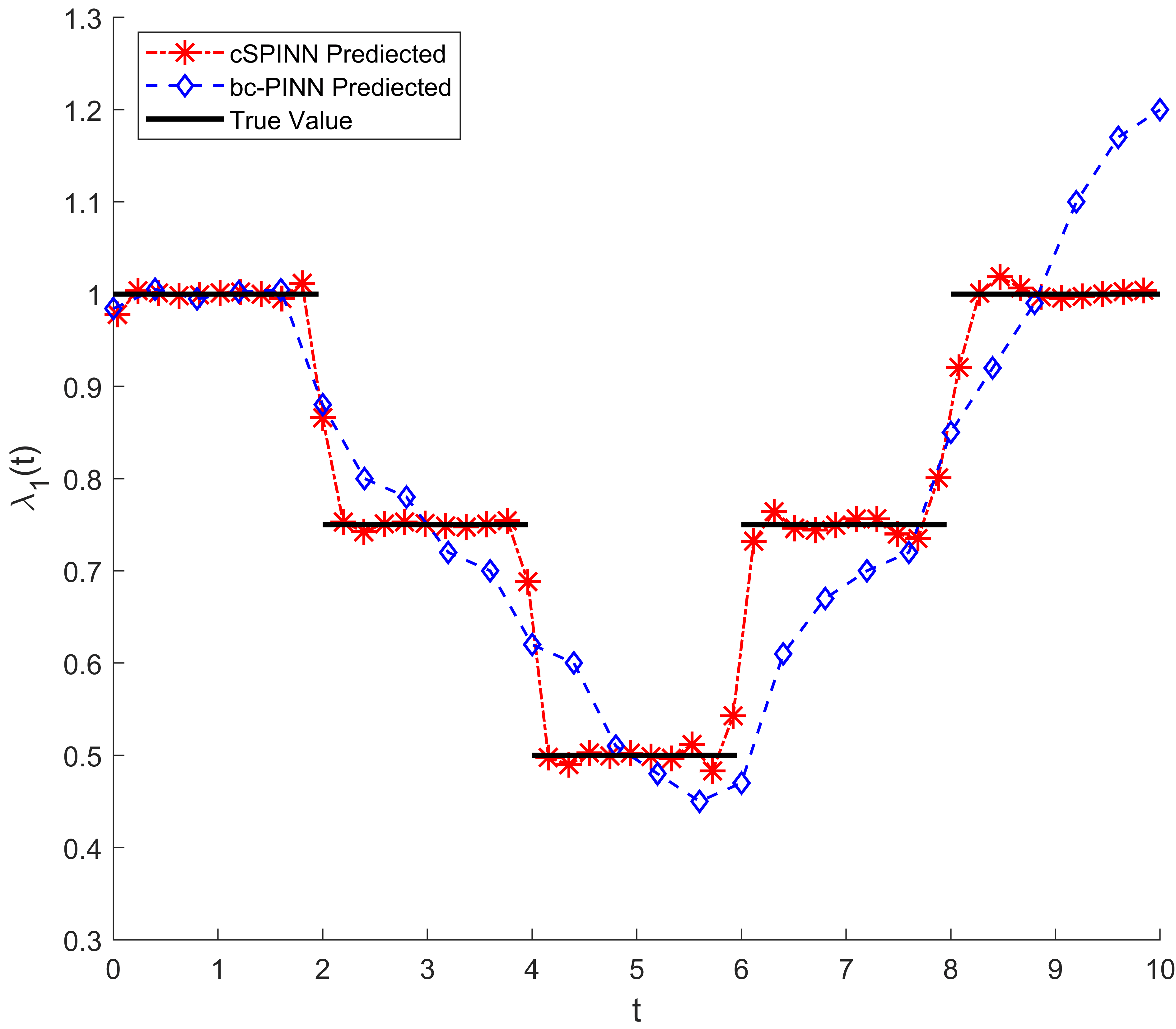}
			\end{minipage}
			\begin{minipage}{0.42\textwidth}
				\centering
				\includegraphics[width=1\textwidth]{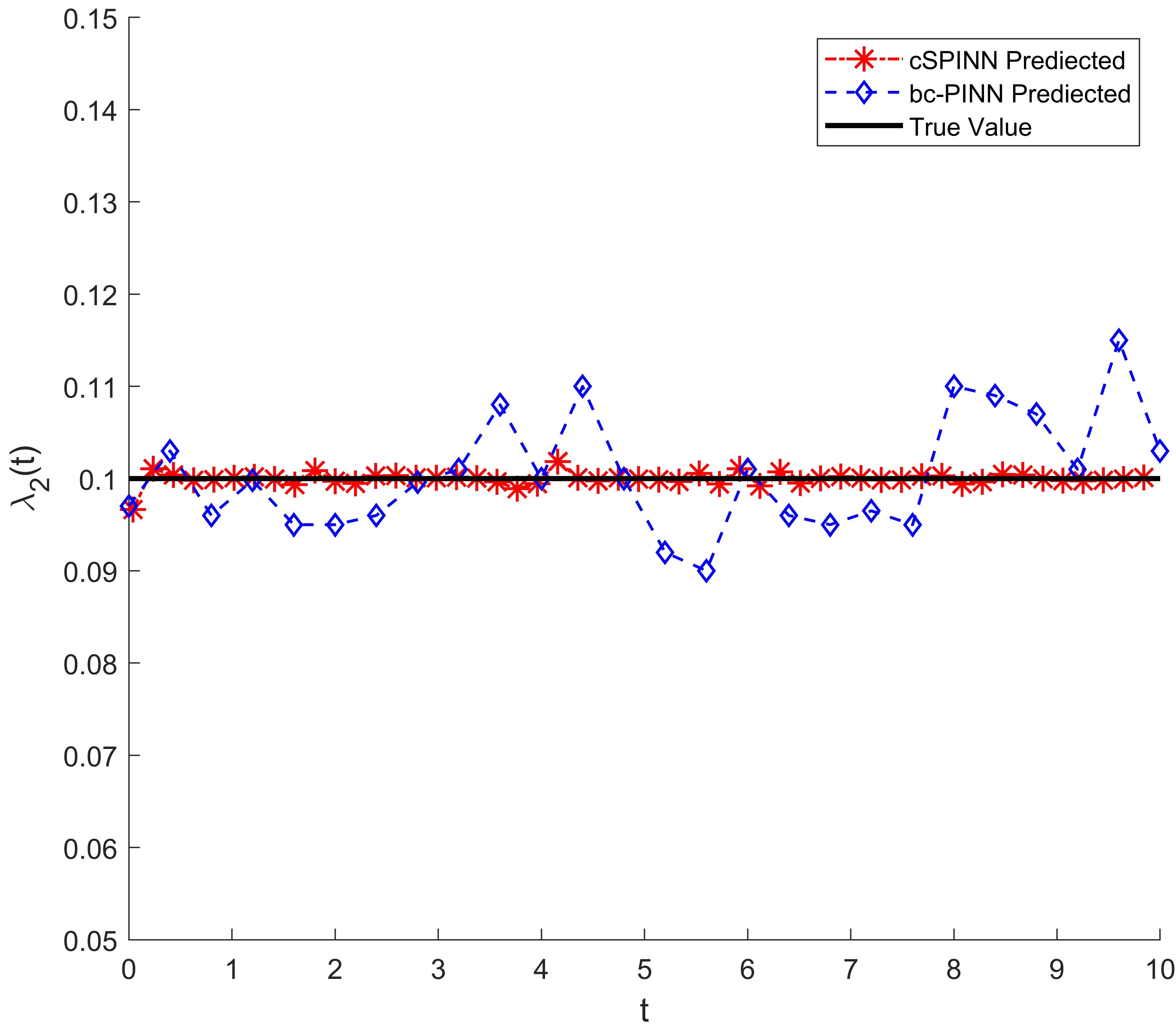}
			\end{minipage}
		}
		\subfigure{
			\begin{minipage}{0.42\textwidth}
				\centering
				\includegraphics[width=1\textwidth]{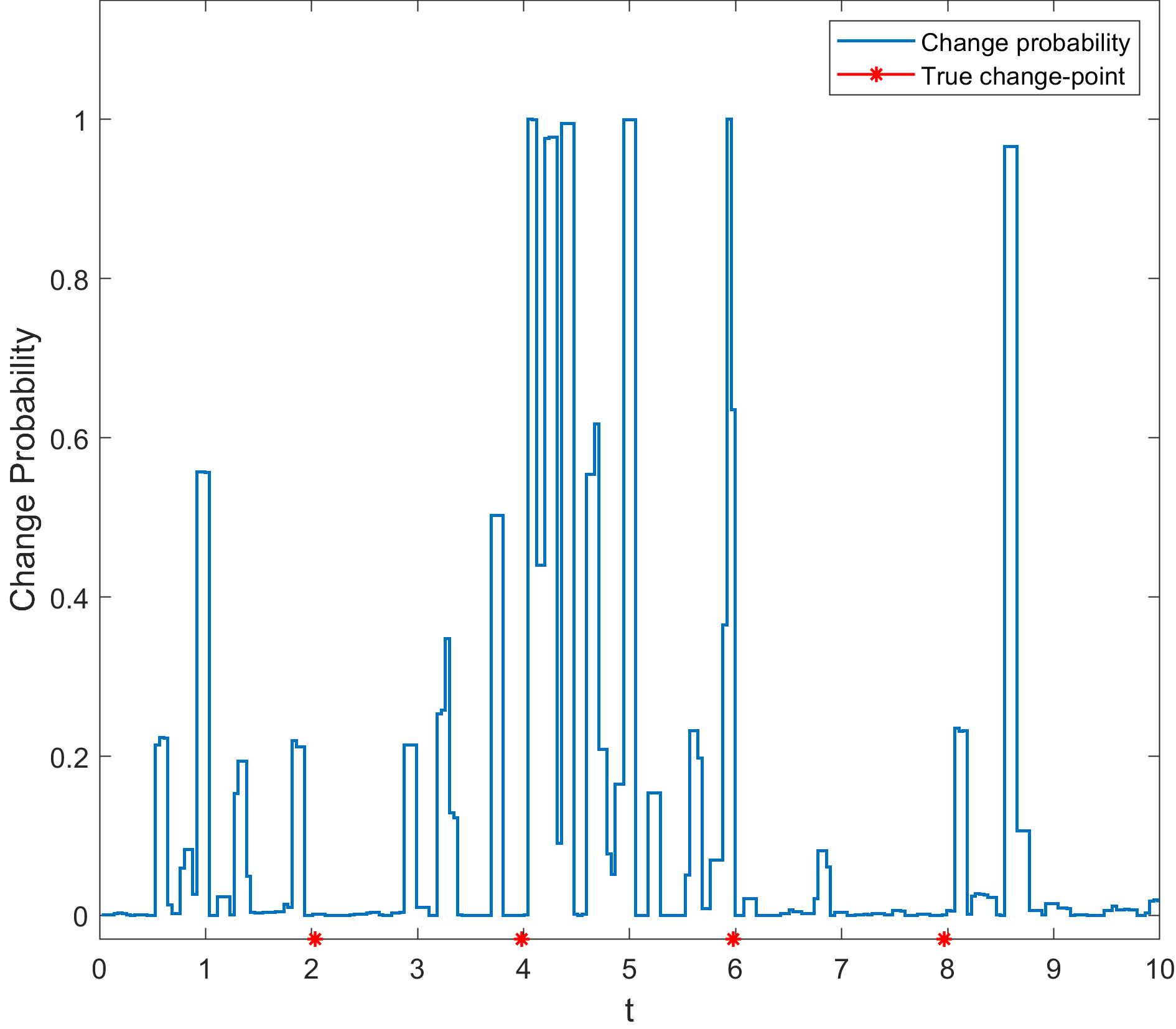}
			\end{minipage}
			\begin{minipage}{0.42\textwidth}
				\centering
				\includegraphics[width=1\textwidth]{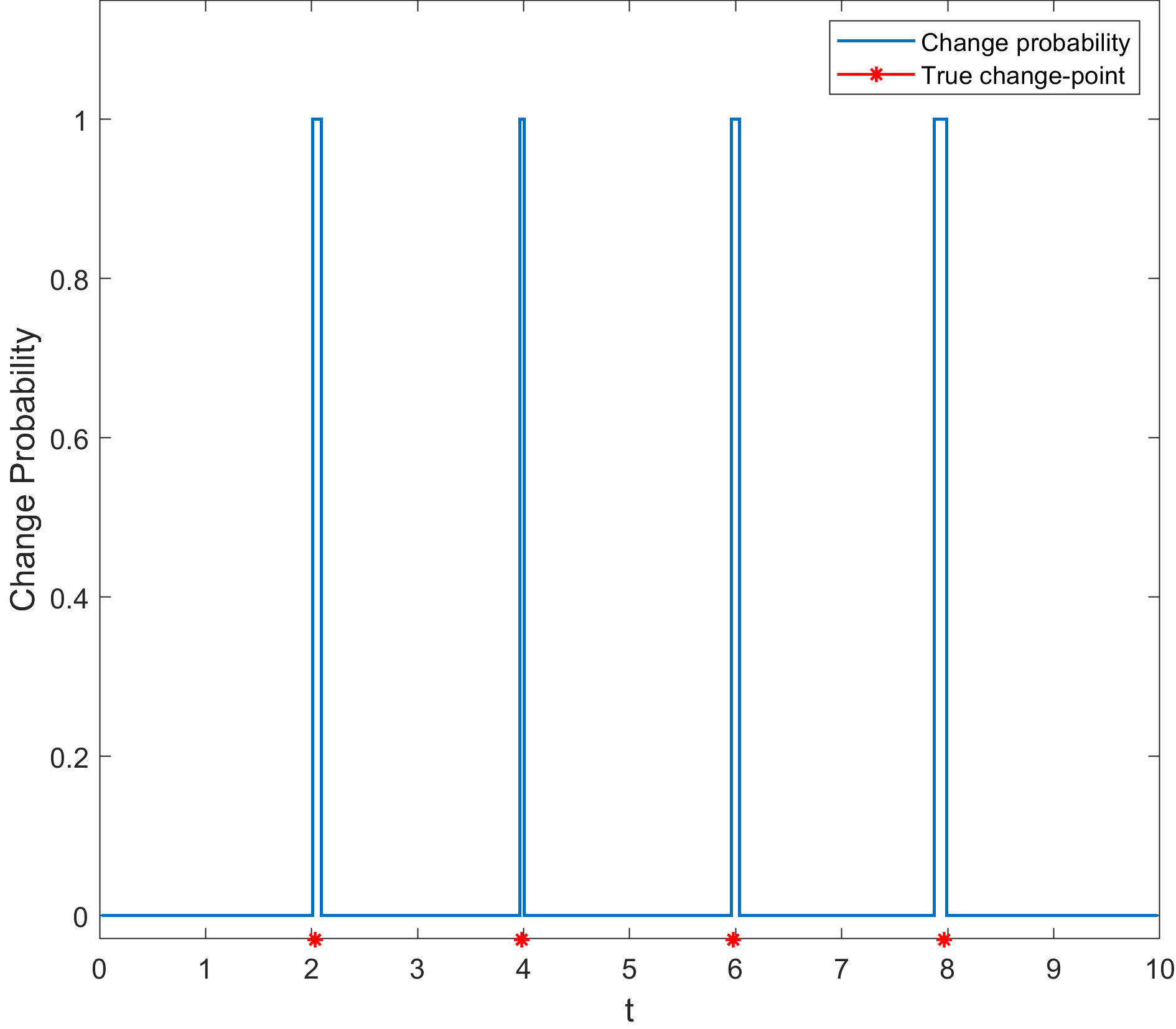}
			\end{minipage}
		}
		\caption{\label{Result 4}Figures from the first row represent $\lambda_{1}$ and $\lambda_{2}$ learned results. The second column illustrates the change point detection results of $\lambda_{1}$ learned. The first picture in the second row shows the four detected change points based on modified cSPINNs and the second picture shows the bad results based on bc-PINNs.}
	\end{figure} 
	
	\subsection{Comparison of Modified cSPINNs with bc-PINNs}
	
	In this part, we will compare the results of modified cSPINNs and bc-PINNs. We draw the predicted results of $\lambda_{1}$ learned from bc-PINNs \cite{mattey2022novel} and modified cSPINNs together with their detected change points in the following figure \ref{Result 4}. The shape of bc-PINNs' result is more like a smooth parabola with no apparent cut-off points for three different steps. Consequently, the parameter identification result of bc-PINNs will obtain wrong detected change points resulting in a larger variance statistical result. In contrast, the result of modified cSPINNs fits the reference solution better. Based on it, the finite mixture model can detect four change points precisely. In this case, the transition path of Burgers' equation parameter $\lambda$ is
	\begin{equation}
		\begin{split}
			{\rm Comparison \: case}:\lambda_1(t)= \begin{cases}1.00, & 0 \leq t<2, \\ 0.75, & 2 \leq t<4, \\ 0.50, & 4 \leq t<6, \\ 0.75 & 6 \leq t<8, \\ 1.00, & 8 \leq t<10.\end{cases}\quad
			\lambda_2(t)= 0.1,\: \:  0 \leq t< 10.
		\end{split}
	\end{equation}
	
	The bc-PINNs algorithm has been previously used to solve PDE inverse problems with time-varying parameters. However, this method was found to have limited accuracy, and it was unable to accurately detect the system's change points. In contrast, the cSPINNs algorithm has been developed as a new and improved approach for solving these types of problems. With significantly higher accuracy than bc-PINNs, cSPINNs can accurately identify the turning points in a system, which is essential for many scientific applications. By using cSPINNs, we can gain deeper insights into complex systems and develop more accurate models to describe their behavior. As a result, the cSPINNs algorithm is a powerful tool for scientific computing and can be used to accurately detect change points in a wide range of complex systems in combination with the finite mixture model.

	\section{2D Wave Equation with Space-varying Parameter}\label{six_sec}
	
	Here, we consider the 2D space-varying acoustic wave equation as another test case for our framework. The parametric wave equation is \ref{2.5} and the space-varying parameter is $\alpha(x,y)$. Similarly, we firstly use the modified cSPINNs to infer the space-varying parameter $\alpha(x,y)$, whose loss function could be defined as: 
	\begin{subequations}
		\begin{align}
			& R_{PDE} := \alpha^2 \nabla^2 \phi - \frac{\partial^2 \phi}{\partial t^2},\\
			& R_{P.C} := \rho \alpha^2 \nabla^2 \phi(x, t, z=0),\\
			& R_{S_1} := \nabla \phi(x, z, t=t_1^0) - U_1^0(x, z),\\
			& R_{S_2} := \nabla \phi(x, z, t=t_2^0) - U_2^0(x, z),\\
			& R_{obs} := \nabla \phi(x, z, t) - U_{obs}(x, z, t),
		\end{align}
	\end{subequations}
	with the domain is $\{(x, z, t) | (x, z, t) \in [0, 1.2]\times[0, 0.45]\times[0, 0.5]\}$. We construct a 2D domain with a certain distributed wave speed $\alpha(x,y)$ and obtain the result by using the package specfem2D\cite{song2021wavefield}. Moreover, we impose a free-surface condition for the 2D domain. The generated seismograms are the observed data for inferring, and we use two early-time snapshots of the displacement field for training, which are taken before the wave interacts with any heterogeneities in the ground truth model. The reference solution used for the training set can be found in the left part of figure \ref{space_varying}.
	
	The space-varying wave speed parameter $\alpha(x,y)$ is what we need to infer and for the direct comparison, we also set the weights of different loss terms as $\lambda_1=0.1, \lambda_2=1, \lambda_3=1, \lambda_4=0.1$ during training, which could be denoted as 
	\begin{equation}
		\begin{aligned}
			M S E(\Theta)=\lambda_1 M S E_{P D E}+\lambda_2 M S E_S+\lambda_3 M S E_{P . C}+\lambda_4 M S E_{O b s}.
		\end{aligned}
	\end{equation}
	
	For the following training, we select $N_r=40,000$ residual points from a mesh with size $200\times200$ and $N_b=5,000$ boundary points from each edge. Our architecture here is a fully-connected neural network, trained by using the modified cSPINN scheme with four corresponding stages. For the backbone network, an MLP with a depth of 8 and a width of 100 is used; for the sub-network, a fully-connected neural network with a depth of 5 and a width of 10 is used. Similar to the modified cSPINNs for the forward problems, we train the PINNs in the following four stages with $N_{S_1} = 5000,  N_{S_2} = 2000,  N_{S_3} = N_{S_4} = 30000$. In Stage 3 and Stage 4, an exponential learning rate decay method for the Adam optimizer is applied with a decay rate of 0.7 every 2500 iterations. Then, L-BFGS is used to optimize the backbone network with 1000 epochs further.

\begin{figure}[ht]
		\centering
		\begin{minipage}{0.31\textwidth}
			\centering
			\includegraphics[width=1\textwidth]{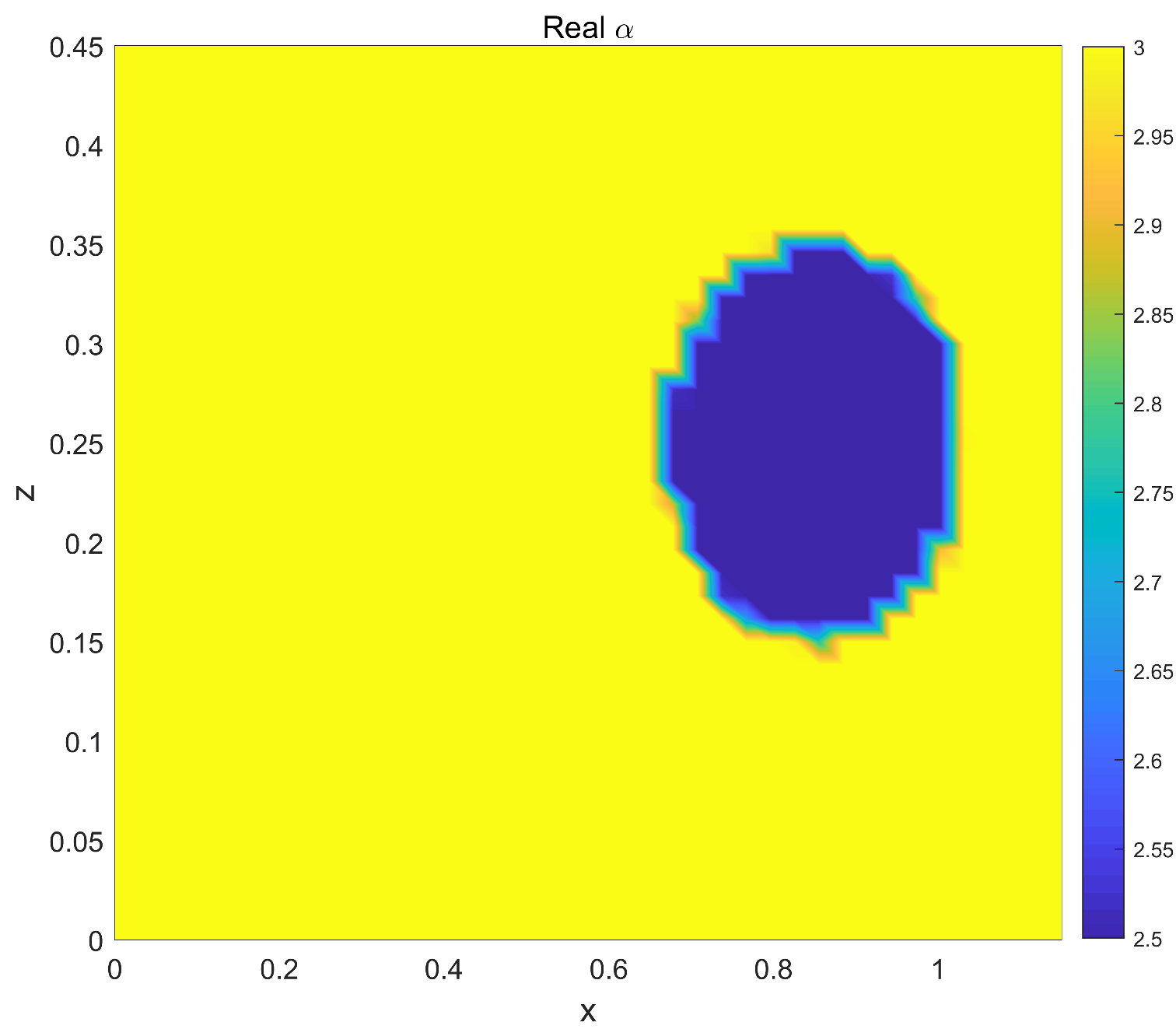}
		\end{minipage}
		\begin{minipage}{0.31\textwidth}
			\centering
			\includegraphics[width=1\textwidth]{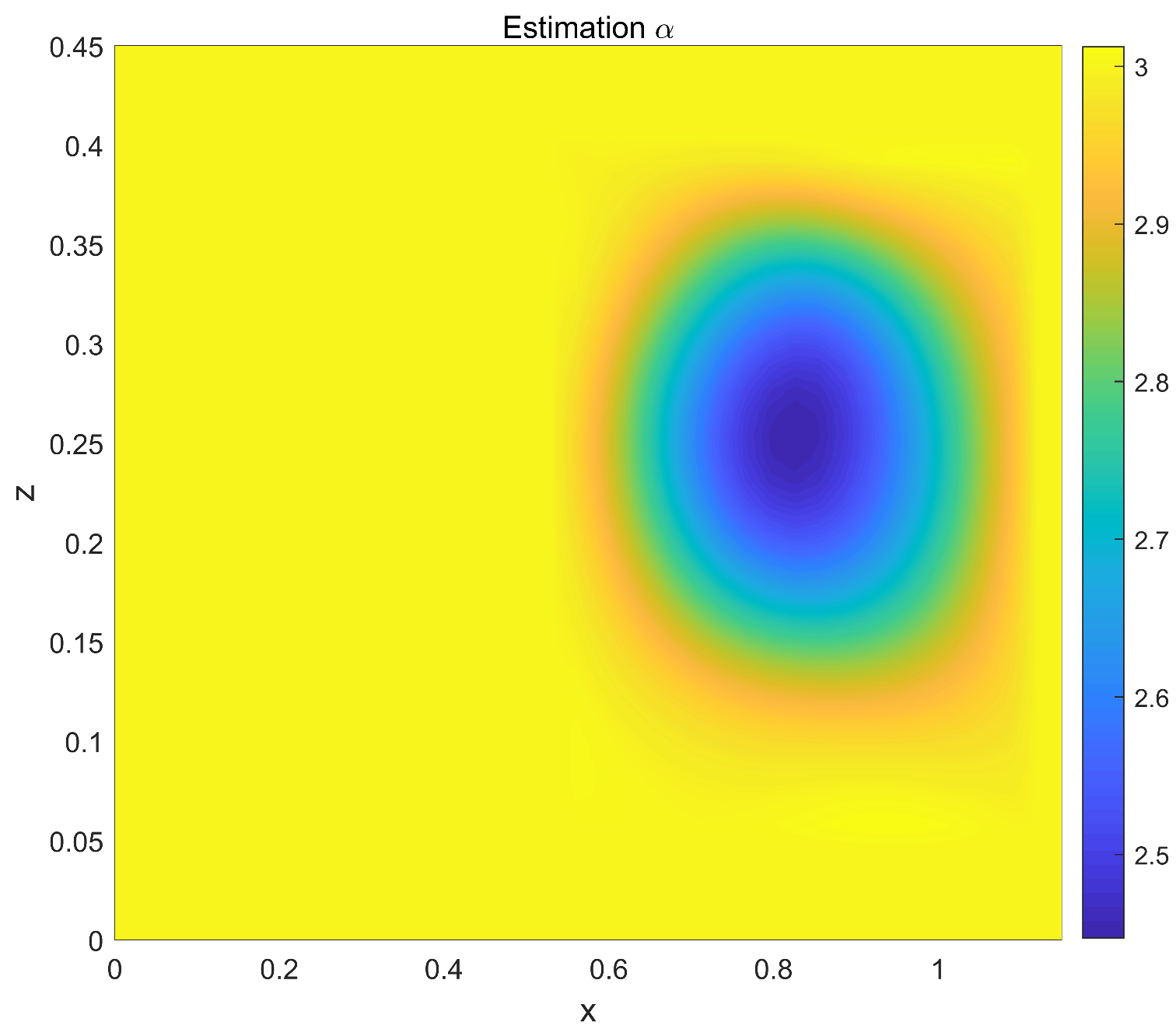}
		\end{minipage}
		\begin{minipage}{0.31\textwidth}
			\centering
			\includegraphics[width=1\textwidth]{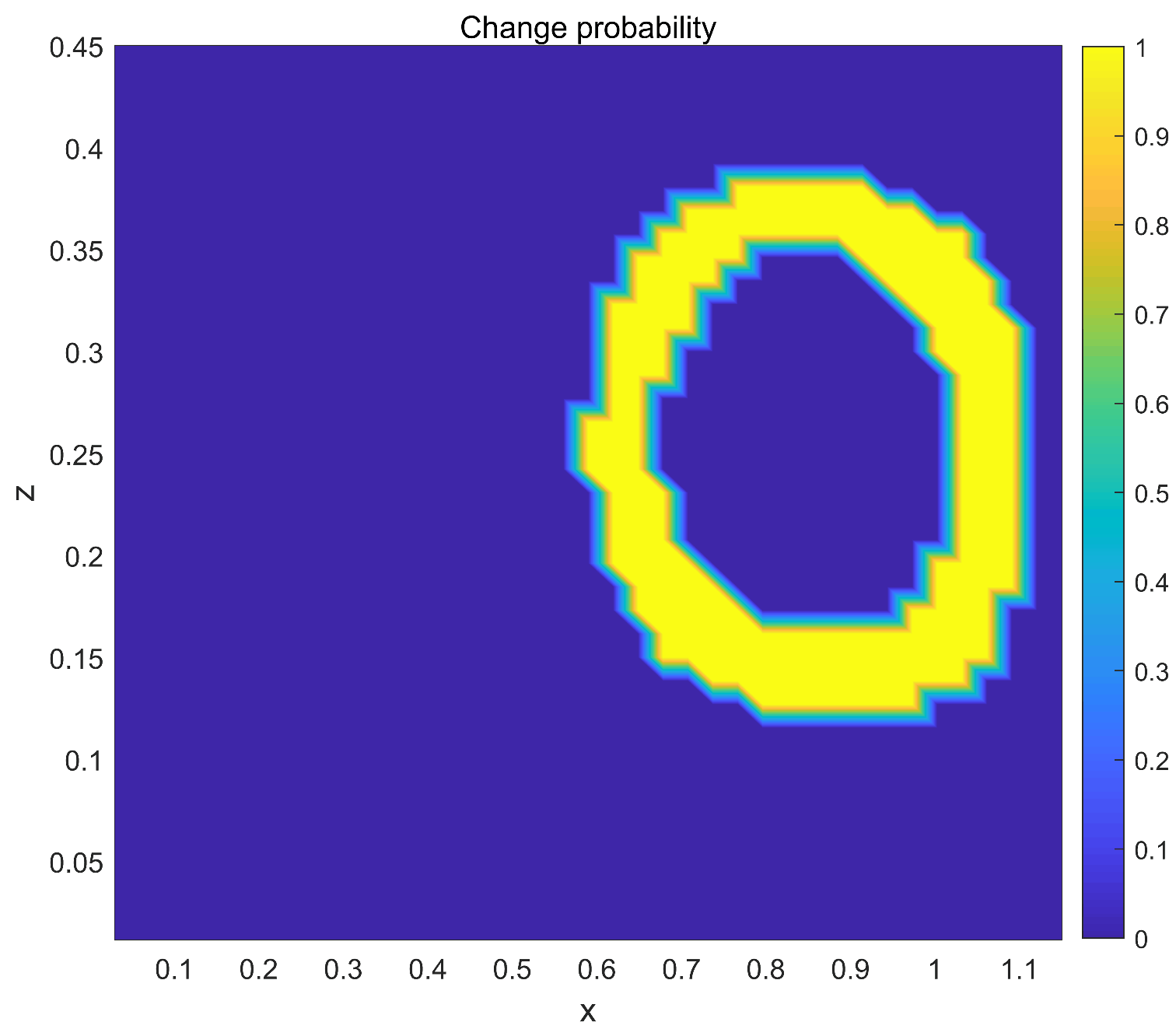}
		\end{minipage}
		\caption{\label{space_inverse}From the left to right, they are the numerical solution of specfem2D, inferring results of modified cSPINNs and change point detection results.}
	\end{figure}

	In this case, a low-velocity anomaly, taking the shape of an ellipsoid, with a wave speed of $2 km/s$, is situated within a uniform background model with a wave speed of $3 km/s$. The transition in velocity between the anomaly and the background is abrupt, resembling a sharp step function. The first one of figure \ref{space_inverse} shows a good match between the wave simulated with specfem2D applied to the wave speed model. The second one is the modified cSPINNs solution shows the inverted solution for the wave speed parameter $\alpha(x,y)$ for the 2D wave equation. Compared with the reference solution, the inverted solution is smoothed instead of the sharp discontinuous transition. The last picture is the result of the finite mixture model which corresponds well with the reference solution.  
	
	Deep learning statistical algorithms to infer the locations of parameter variations in spatial properties from the solution of equations has significant implications. It means that we can infer certain characteristics of a system from observation data without prior knowledge of all parameters and physical properties. This approach is particularly useful for practical problems, as real-world systems often contain numerous complex parameters and physical properties that may have intricate relationships with each other. With deep learning algorithms, we can learn these relationships from vast amounts of observation data and use them to make predictions and control the system. We hope that this method would have a wide range of applications in many fields, such as weather forecasting, climate modeling, environmental monitoring, and engineering design. By inferring the locations of parameter variations in spatial properties from equations, we can gain insights into the behavior of complex systems and make more accurate predictions and better control.
	
	\section{Conclusion and Discussion}\label{seven_sec}
	
	The rapid development of parameter identification methods for complex physical models has been enabled by the advancement of computational models. In this study, we propose a novel framework for discovering the hidden transition path of a time-varying complex system. Specifically, we introduce the combination of modified cSPINNs and finite mixture model as the change point detection method to identify change points in the system, then we can discover the transition path behind it. Our method has been tested by using the 1D time-varying parametric Burgers' equation in three different types of evolutionary models, and our framework performed well for all cases. The modified cSPINNs method has been proven to be an efficient approach for parameter identification in time-varying parametric Burgers' equations, and the change point detection method is also crucial for identifying change points in time-varying systems. We use finite mixture models as well as the EM algorithm to offer a straightforward way of describing a system's variation through discrete state space, making statistical computational algorithms to be widely applicable across various fields. Our future works will focus on more challenging models like the Naiver-Stokes equation, which leads to the goal of providing powerful computational tools for the applications of computer vision in detecting angiomas' location and diagnosing vascular aging.
	
	\section*{Appendix A: Parameter Estimation Results}\label{Appendix}
	Here, the following table \ref{Appendix Table} shows the statistical inferring results of the finite mixture model based on the results of modified cSPINNs. Moreover, we also give the inferring results based on the results of bc-PINNs. The results contain the parameter estimation results, the Gaussian variance, and the mixture ratio. Take case 3 as an example, the mixture ratio of $\lambda_{1}$ with values of 0.5052, 0.7663, and 1.0015 is 0.1908, 0.4479, and 0.3612. The last column is the L2 relative error for modified cSPINNs about the 1D time-varying parametric Burgers' Equation. The value of $\lambda_{k}$ represents the L2 error between the inferring results and the reference solution. And the value of $u(t,x)$ is the error between the reference solution and the result calculated by $\lambda_{k}$. 
	
	\begin{table}[H]
		\centering
		\caption{Estimates of parameters for the Burgers' Equation with time-varying parameters. \label{Appendix Table} }
		\scalebox{0.9}{
			\begin{tabular}{c|ccccccc}
				\hline
				\hline
				Numerical & Equation & True & Parameter & Gaussian & Mixture & \multicolumn{2}{c}{Relative $L^2$  Error of} \\
				Example & Coefficient & Value & Estimation & Variance & Ratio & $u(t,x)$ & $\lambda_k$ \\
				\hline
				\hline
				Case 1.1:  & $\lambda_1$ & 1.50 & 1.4996 & 1.9800$\mathrm{e}$-5 & 1.0000 & 2.455$\mathrm{e}$-04 & 2.978$\mathrm{e}$-03 \\
				\hline
				Non-change & $\lambda_2$ & 0.10 & 0.1000 & 1.6652$\mathrm{e}$-7 & 1.0000 &  & 4.081$\mathrm{e}$-03 \\
				\hline
				\hline
				Case 1.2: & $\lambda_1$ & 0.50 & 0.4988 & 8.9737$\mathrm{e}$-5 & 0.5000 & 1.348$\mathrm{e}$-04 & 1.709$\mathrm{e}$-02 \\
				Single-change  &  & 1.00 & 0.9985 & 2.7171$\mathrm{e}$-5 & 0.5000 &  &  \\
				\hline
				& $\lambda_2$ & 0.10 & 0.1000 & 9.1572$\mathrm{e}$-8 & 1.0000 & & 3.026$\mathrm{e}$-03 \\
				\hline
				\hline
				Case 1.3: & $\lambda_1$ & 0.50 & 0.5060 & 9.6023$\mathrm{e}$-4 & 0.5000 & 7.419e-05 & 3.472e-03 \\
				Gradual change &  & 1.00 & 0.9938 & 9.3191$\mathrm{e}$-4 & 0.5000 &  & \\
				\hline
				& $\lambda_2$ & 0.10 & 0.1000 & 0.3418$\mathrm{e}$-8 & 1.0000 &  & 2.897e-03  \\
				\hline 
				\hline
				Case 2.1: & $\lambda_1$ & 0.50 & 0.5001 & 3.0196$\mathrm{e}$-7 & 0.8253 & 2.110e-04 & 3.389e-02  \\
				Multi-change &  & 1.00 & 0.7897 & 0.0582 & 0.1747 &  &  \\
				\hline
				Two States& $\lambda_2$ & 0.10 & 0.1001 & 1.2926$\mathrm{e}$-6 & 1.0000 &  & 1.139e-02 \\
				\hline
				\hline
				Case 2.2: & $\lambda_1$ & 0.50 & 0.4987 & 6.6471$\mathrm{e}$-5 & 0.3693 & 3.514e-04 & 3.169e-02 \\
				Multi-change &  & 0.75 & 0.7570 & 0.0044 & 0.4511 &  &  \\
				Three States &  & 1.00 & 1.0010 & 4.3976$\mathrm{e}$-5 & 0.1796 &  &  \\
				\hline
				& $\lambda_2$ & 0.10 & 0.1000 & 3.9237$\mathrm{e}$-7 & 1.0000 &  & 6.264e-03 \\
				\hline
				\hline
				Case 3: & $\lambda_1$ & 0.50 & 0.5052  & 1.5886$\mathrm{e}$-4 & 0.1908 & 4.656e-04 & 3.451e-02 \\
				Multi-change  &  & 0.75 & 0.7663 & 0.0043 & 0.4479 &  &    \\
				Three States &  & 1.00 & 1.0015 & 6.7240$\mathrm{e}$-5 & 0.3612 &  &  \\
				\hline
				Two-Parameter & $\lambda_2$ & 1.00  & 0.9964 & 1.1749$\mathrm{e}$-4 & 0.3508 &  & 3.810e-02 \\
				Varying &  & 1.33 & 1.3201 & 0.0188 & 0.4656 &  &    \\
				&  & 2.00 & 1.9989 & 6.1425$\mathrm{e}$-4 & 0.1836 &  &  \\
				\hline 
				\hline
				Comparison case: & $\lambda_1$ & 0.50 & 0.4770 & 4.5955$\mathrm{e}$-4 & 0.1509 & 1.130e-02 & 1.057e-01  \\
				bc-PINNs &  & 0.75 & 0.6825 & 0.0041 & 0.3488 &   &  \\
				for Multi-change &  & 1.00 & 0.9895 & 0.0137 & 0.5003 &  &  \\
				\hline
				& $\lambda_2$ & 0.10 & 0.1004 & 3.7265$\mathrm{e}$-5 & 1.0000 &  & 5.477e-02 \\
				\hline     
				Comparison case: & $\lambda_1$ & 0.50 & 0.4982 & 6.0886$\mathrm{e}$-5 & 0.3593 & 4.627e-04 & 4.119e-02 \\  
				modified cSPINNs  &  & 0.75 & 0.7423 & 0.0048 & 0.4661 &  &   \\
				for Multi-change &  & 1.00 & 0.1007 & 1.2454$\mathrm{e}$-5 & 0.1746 &  &   \\
				\hline
				& $\lambda_2$ & 0.10 & 0.1000 & 6.2961$\mathrm{e}$-7 & 1.0000 &  & 7.949e-03 \\
				\hline     
				\hline
		\end{tabular}}
	\end{table}

	\section*{Appendix B: Absolute Error between Reference and Predicted Solution of 1D parametric Burgers' Equation}\label{Error}
	
\begin{figure}[H]
		\centering
		\subfigure{
			\begin{minipage}{0.32\textwidth}
				\centering
				\includegraphics[width=1\textwidth]{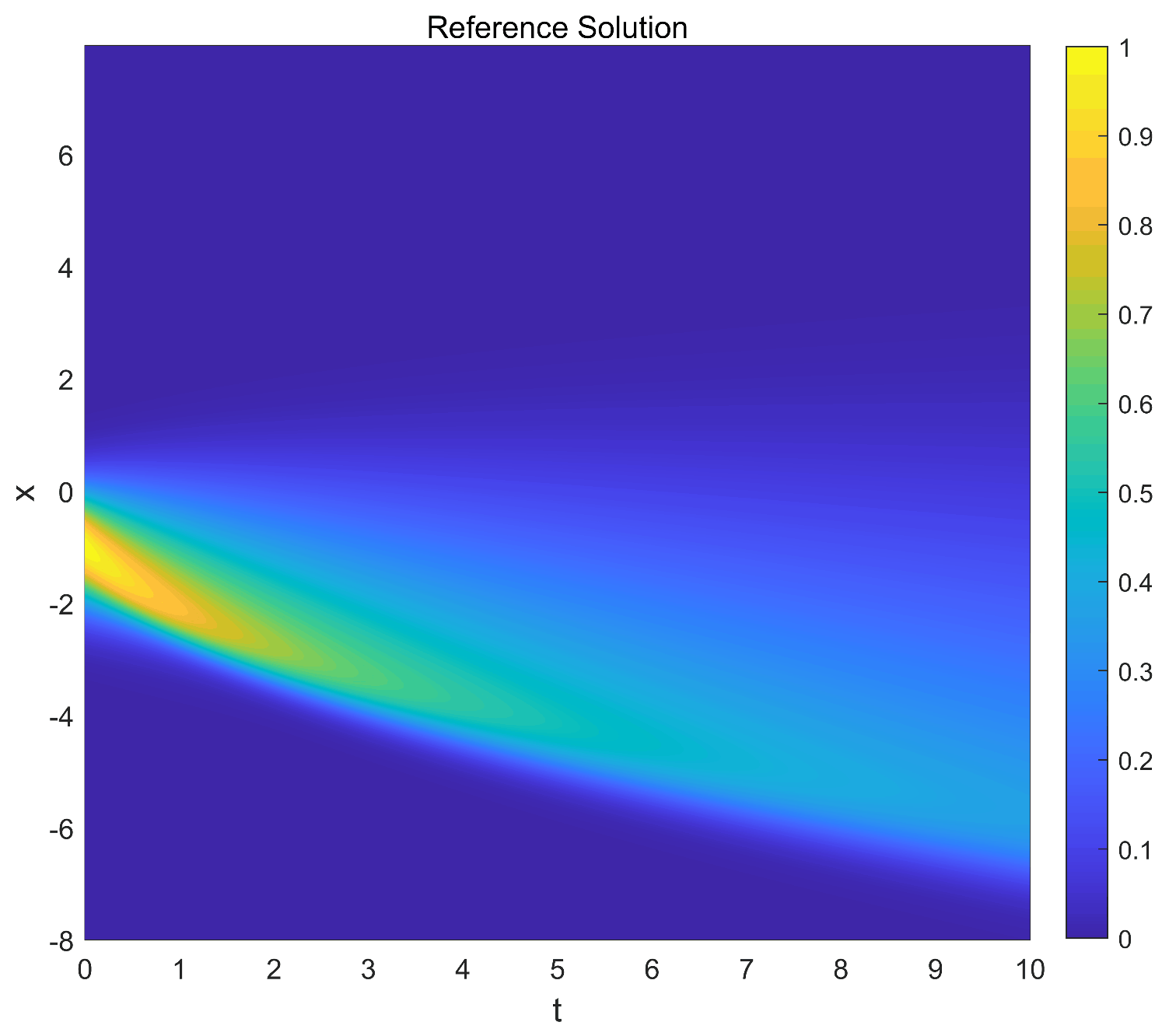}
			\end{minipage}
			\begin{minipage}{0.32\textwidth}
				\centering
				\includegraphics[width=1\textwidth]{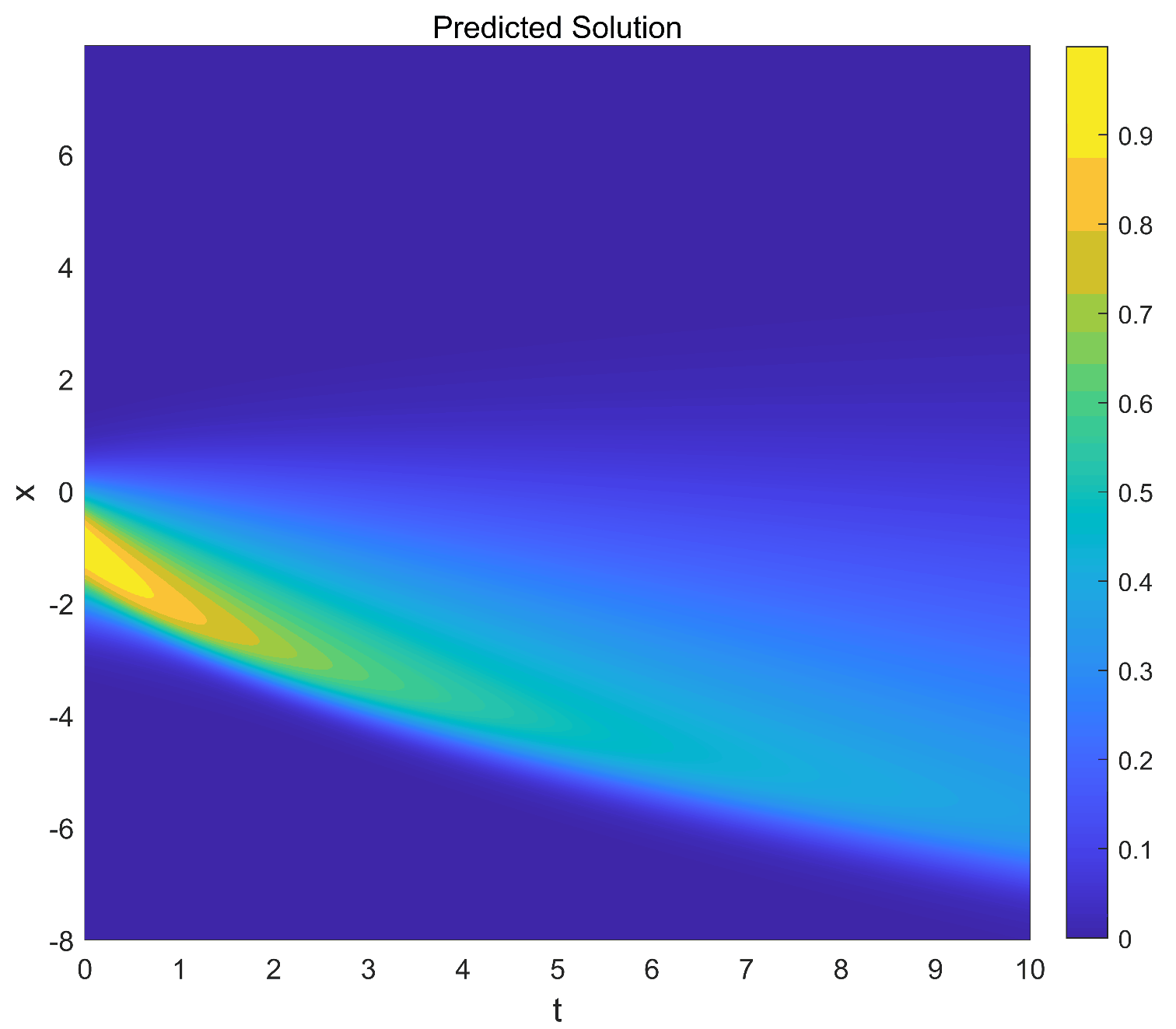}
			\end{minipage}
			\begin{minipage}{0.32\textwidth}
				\centering
				\includegraphics[width=1\textwidth]{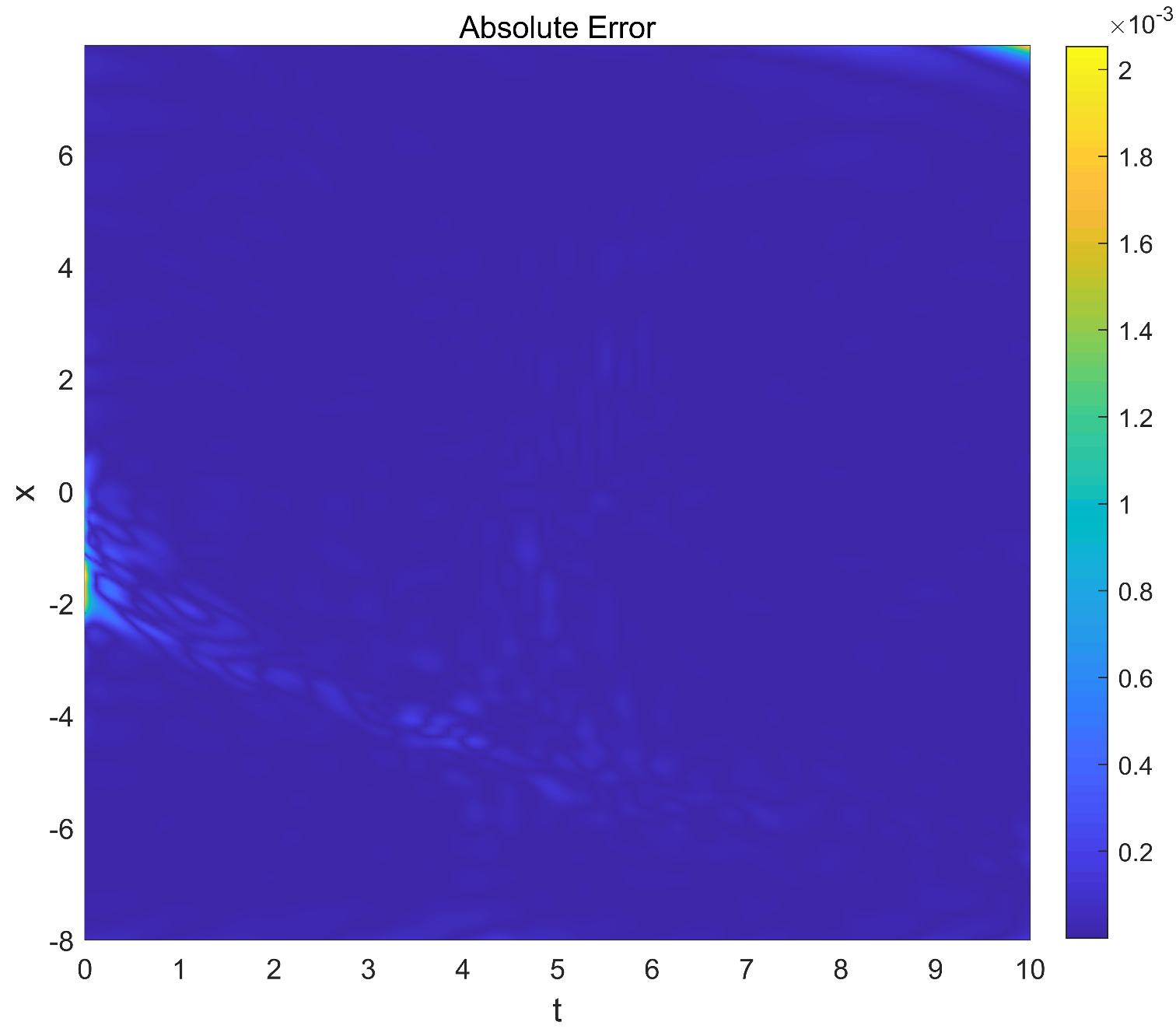}
			\end{minipage}
		}
		\subfigure{
			\begin{minipage}{0.32\textwidth}
				\centering
				\includegraphics[width=1\textwidth]{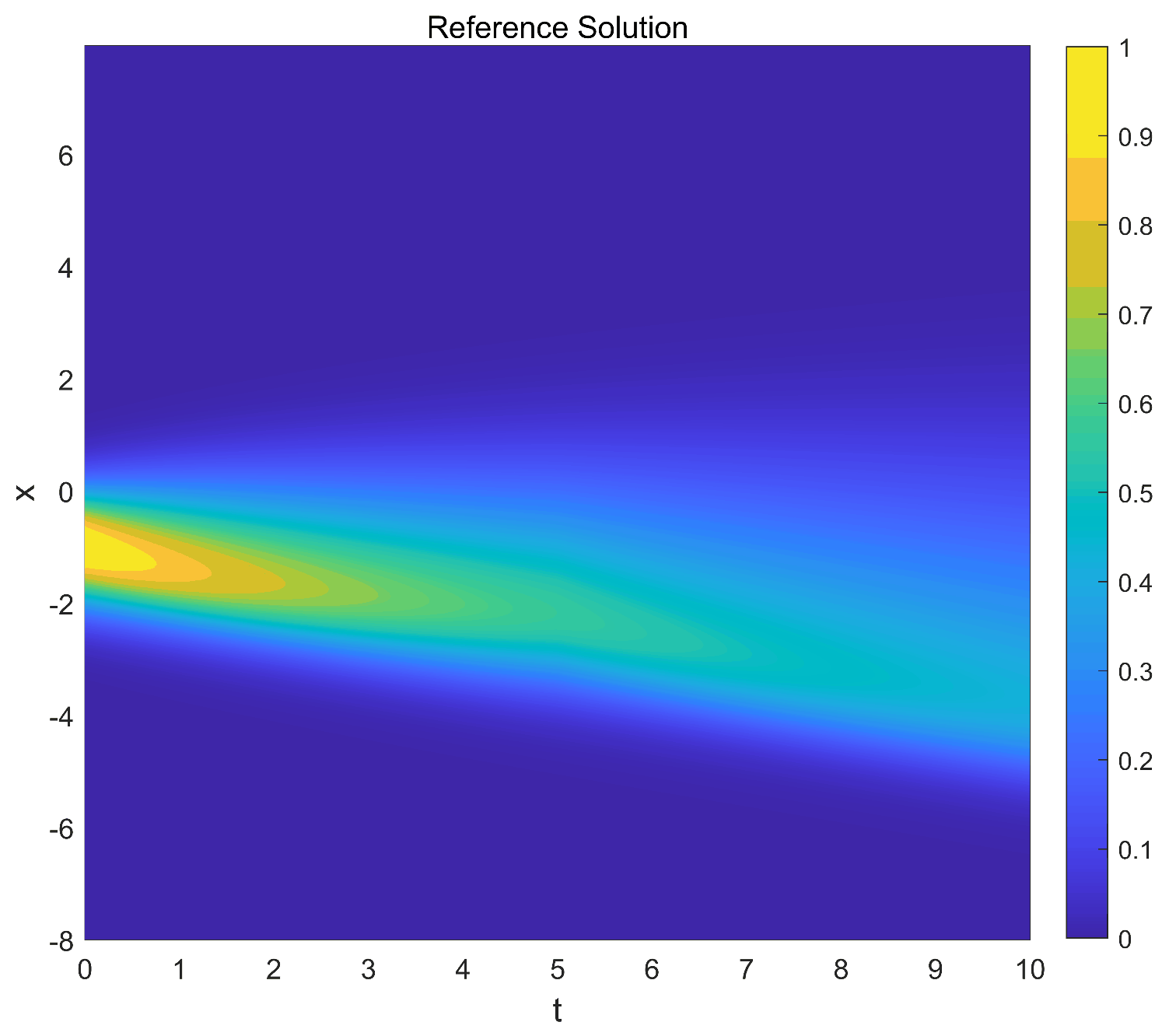}
			\end{minipage}
			\begin{minipage}{0.32\textwidth}
				\centering
				\includegraphics[width=1\textwidth]{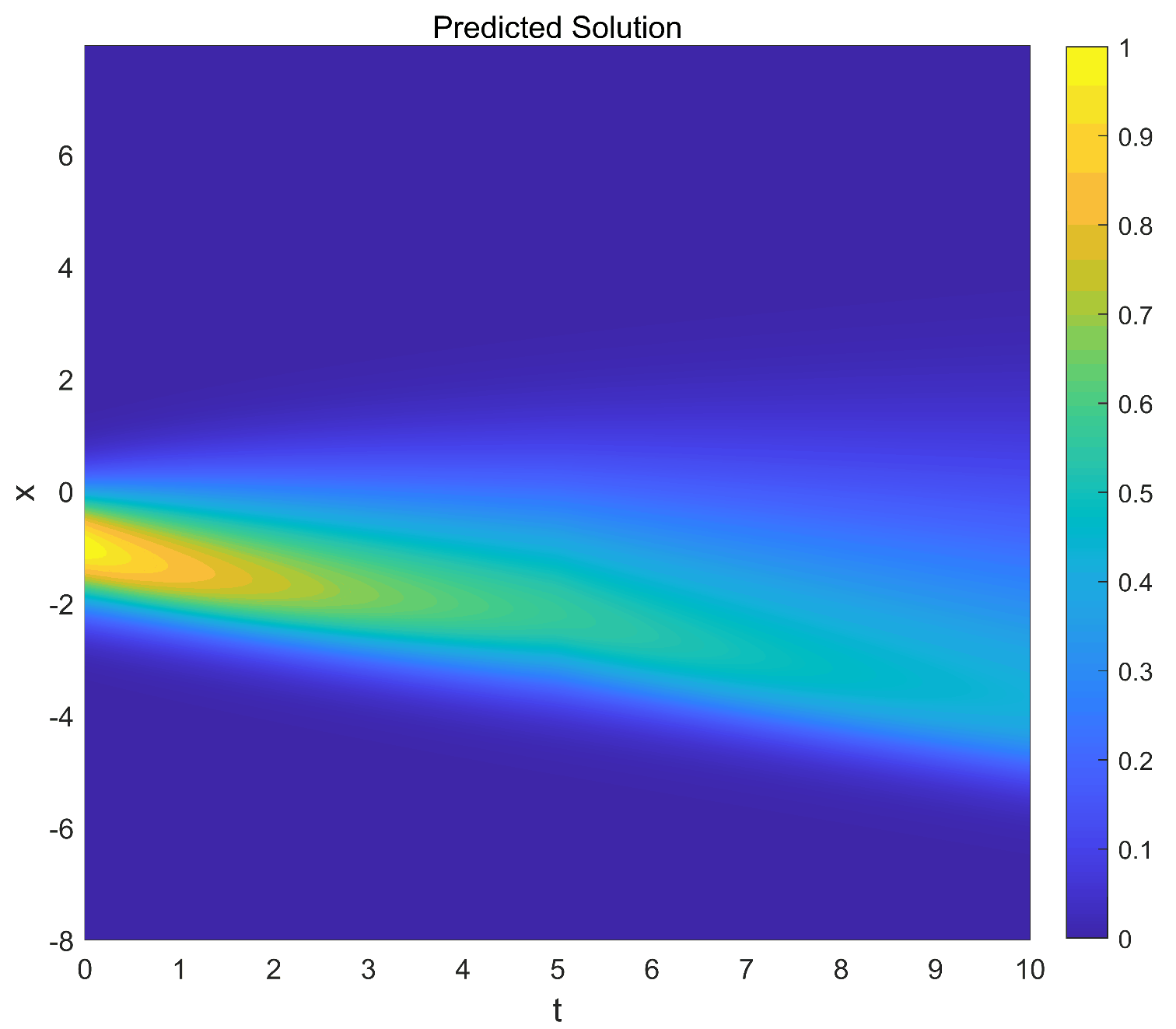}
			\end{minipage}
			\begin{minipage}{0.32\textwidth}
				\centering
				\includegraphics[width=1\textwidth]{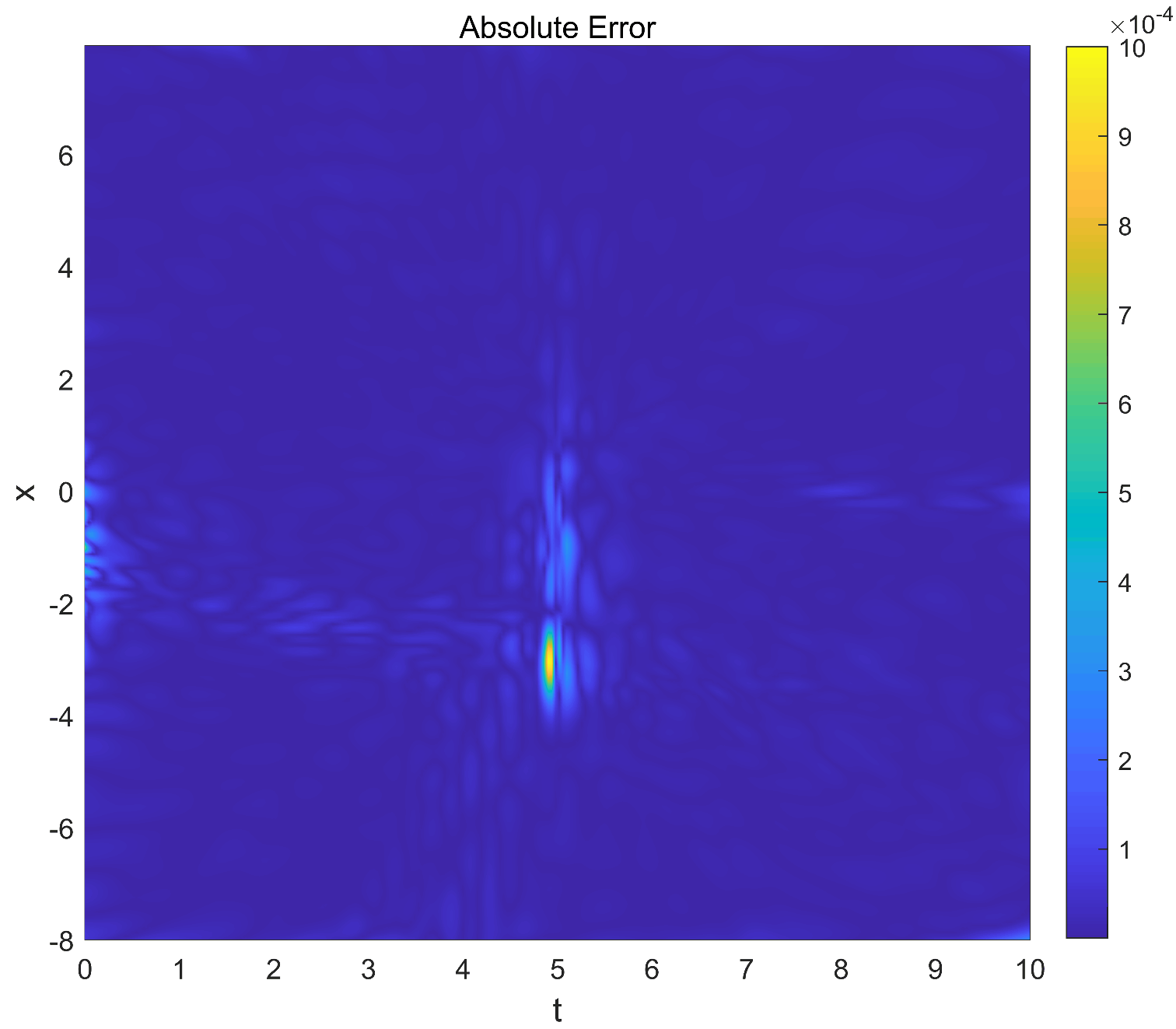}
			\end{minipage}
		}
		\subfigure{
			\begin{minipage}{0.32\textwidth}
				\centering
				\includegraphics[width=1\textwidth]{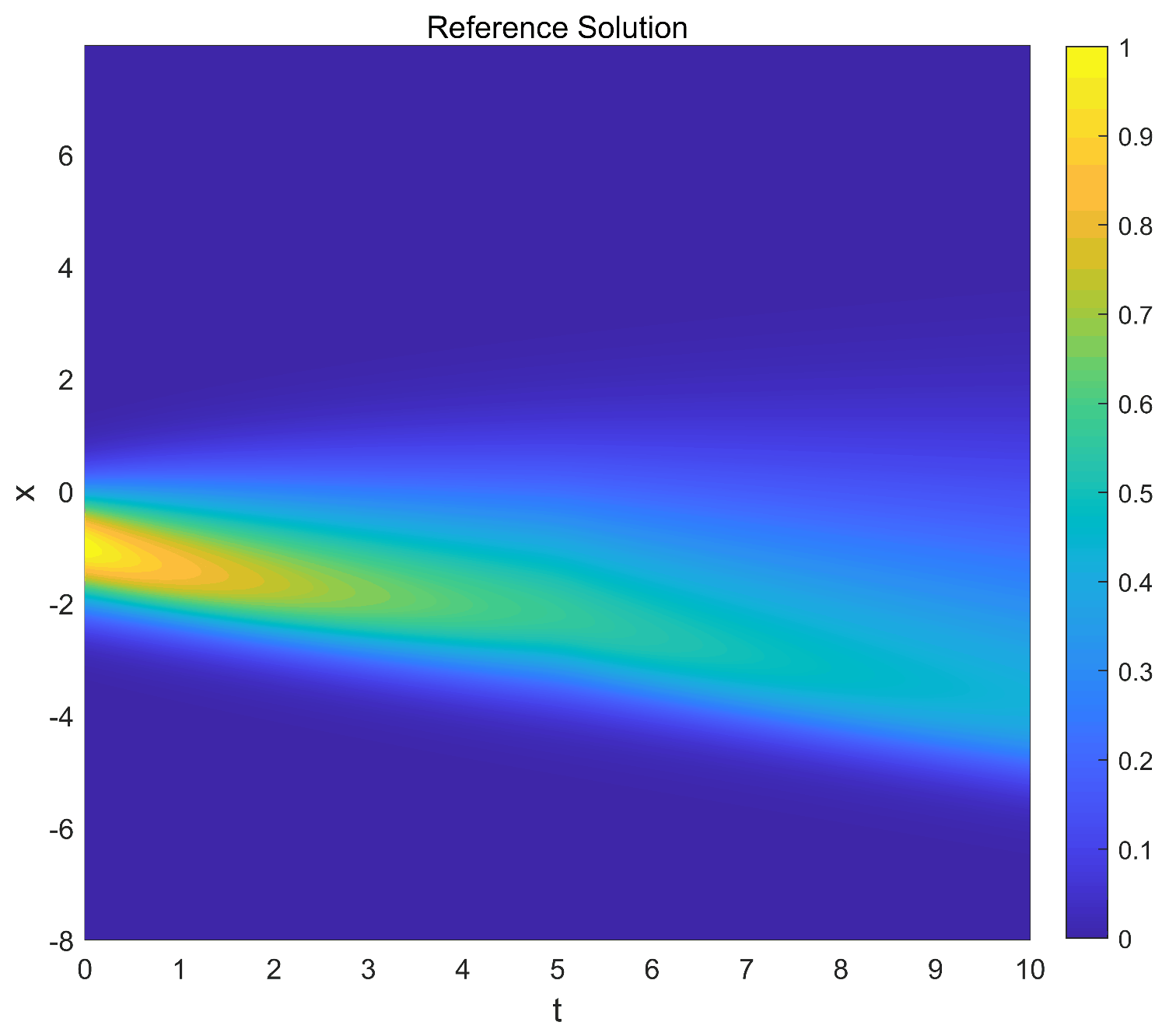}
			\end{minipage}
			\begin{minipage}{0.32\textwidth}
				\centering
				\includegraphics[width=1\textwidth]{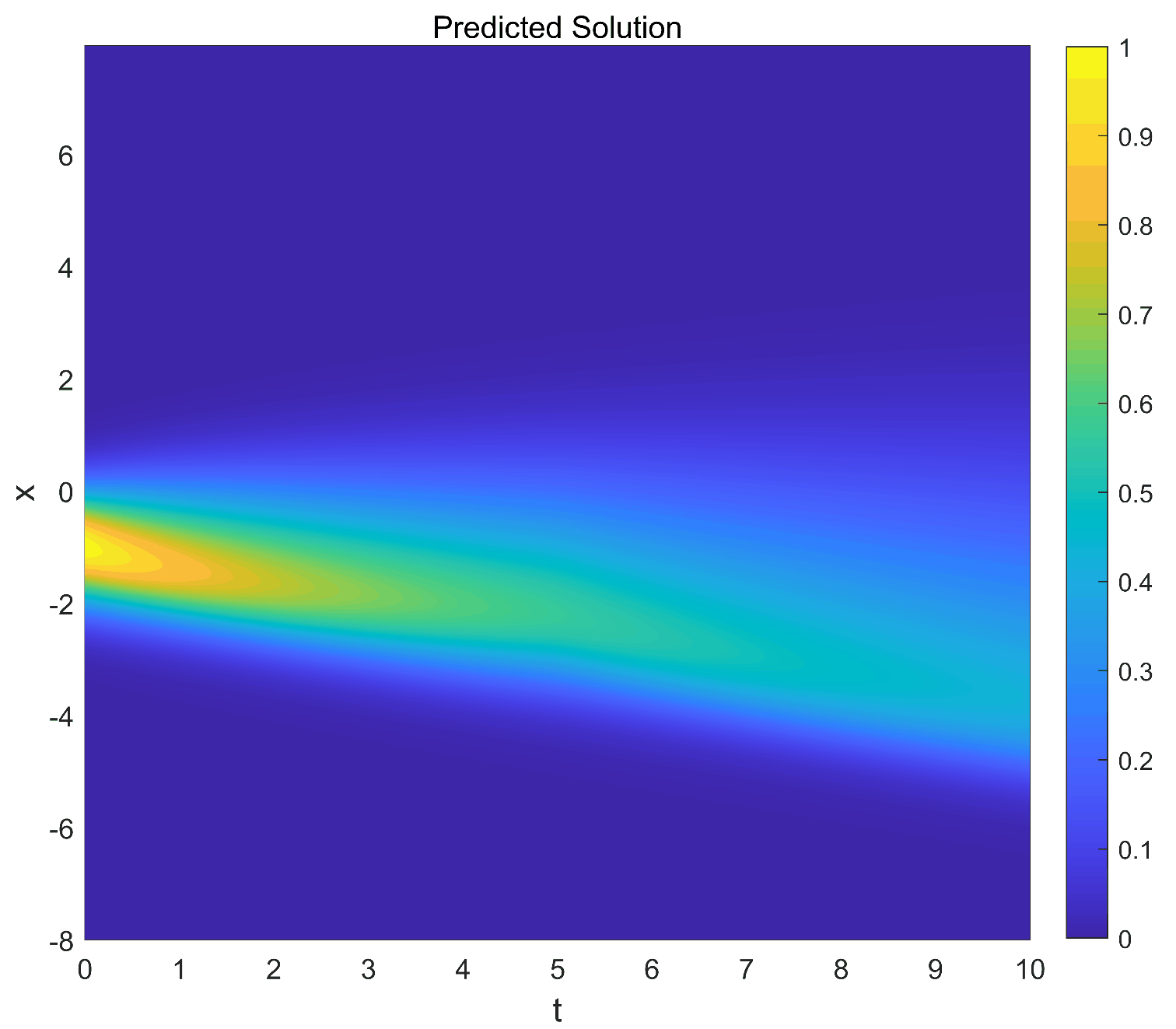}
			\end{minipage}
			\begin{minipage}{0.32\textwidth}
				\centering
				\includegraphics[width=1\textwidth]{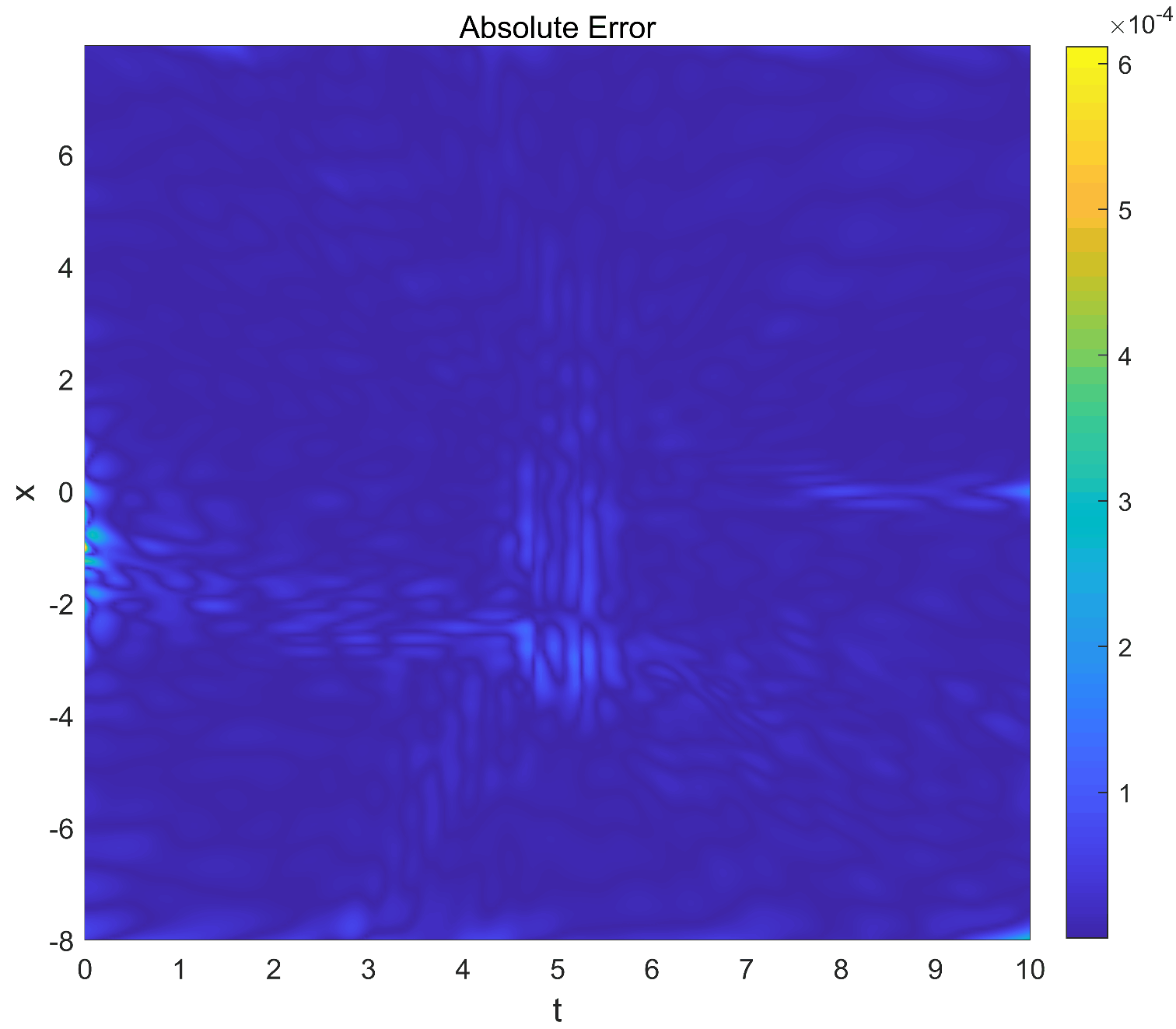}
			\end{minipage}
		}
		\caption{\label{Error for case1} From top to bottom, they are constant parameters with no change point(case 1.1), a single varying parameter with one abrupt shift(case 1.2), and one gradual shift(case 1.3).}
	\end{figure}

\begin{figure}[H]
		\centering
		\subfigure{
			\begin{minipage}{0.32\textwidth}
				\centering
				\includegraphics[width=1\textwidth]{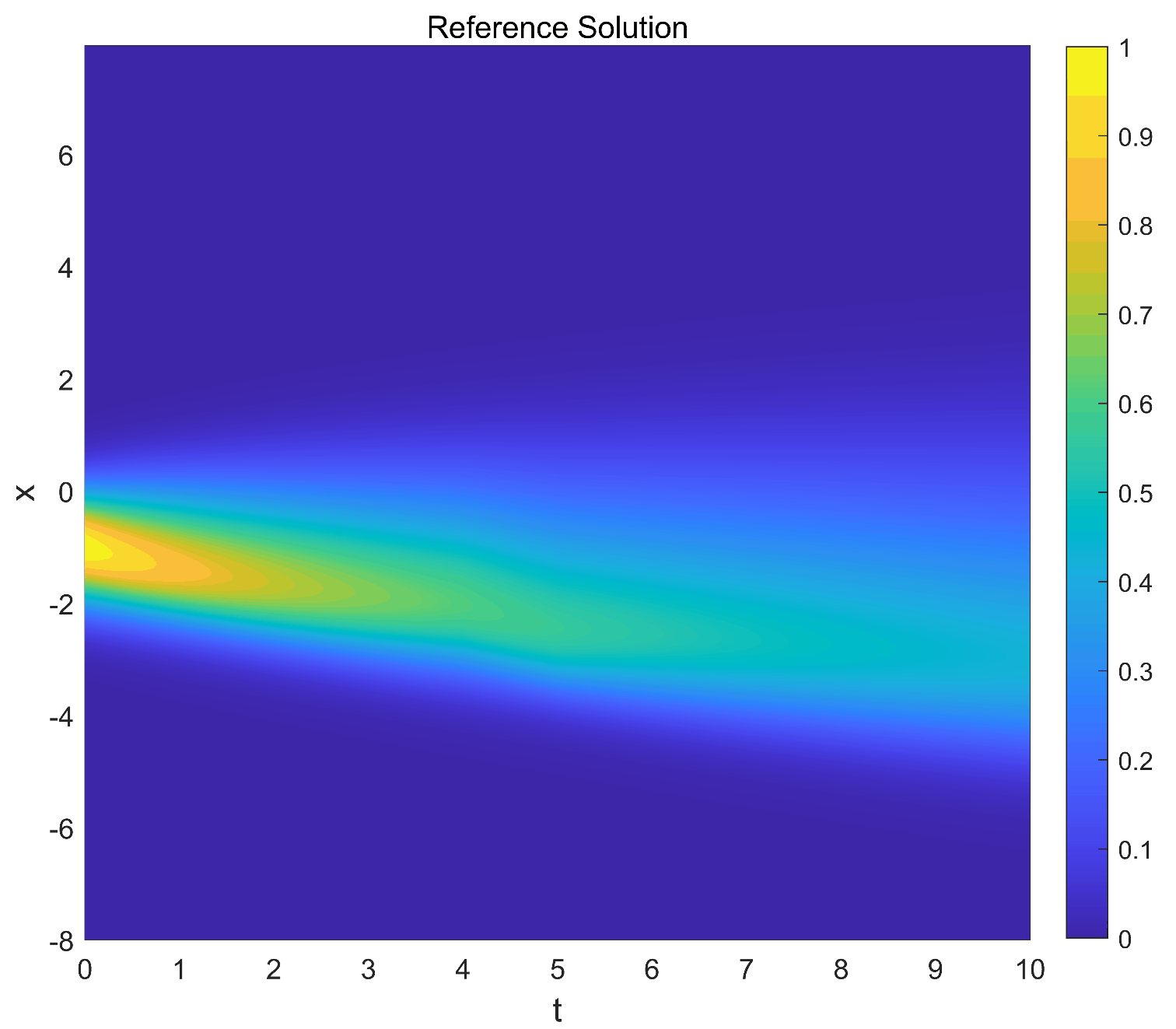}
			\end{minipage}
			\begin{minipage}{0.32\textwidth}
				\centering
				\includegraphics[width=1\textwidth]{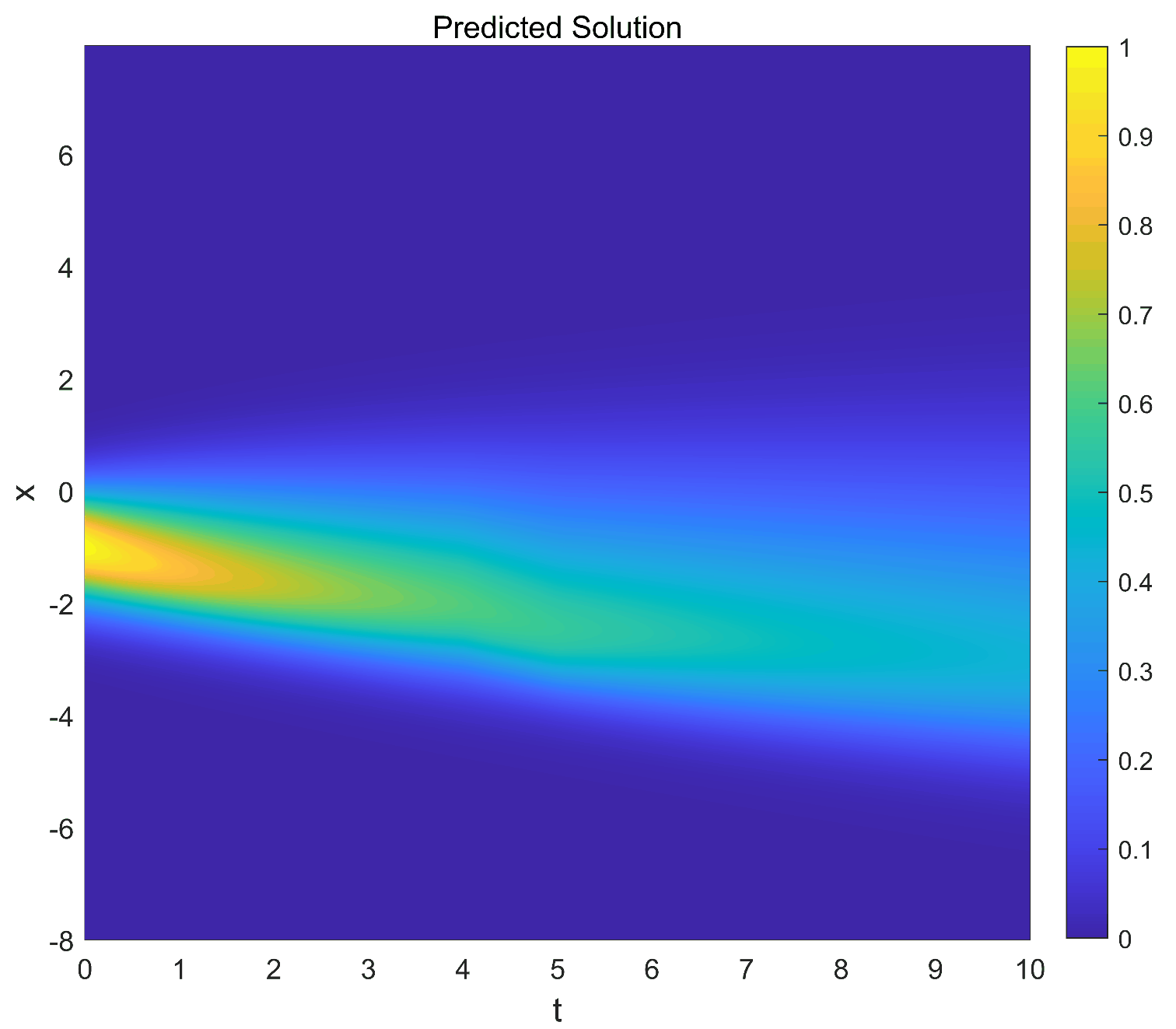}
			\end{minipage}
			\begin{minipage}{0.32\textwidth}
				\centering
				\includegraphics[width=1\textwidth]{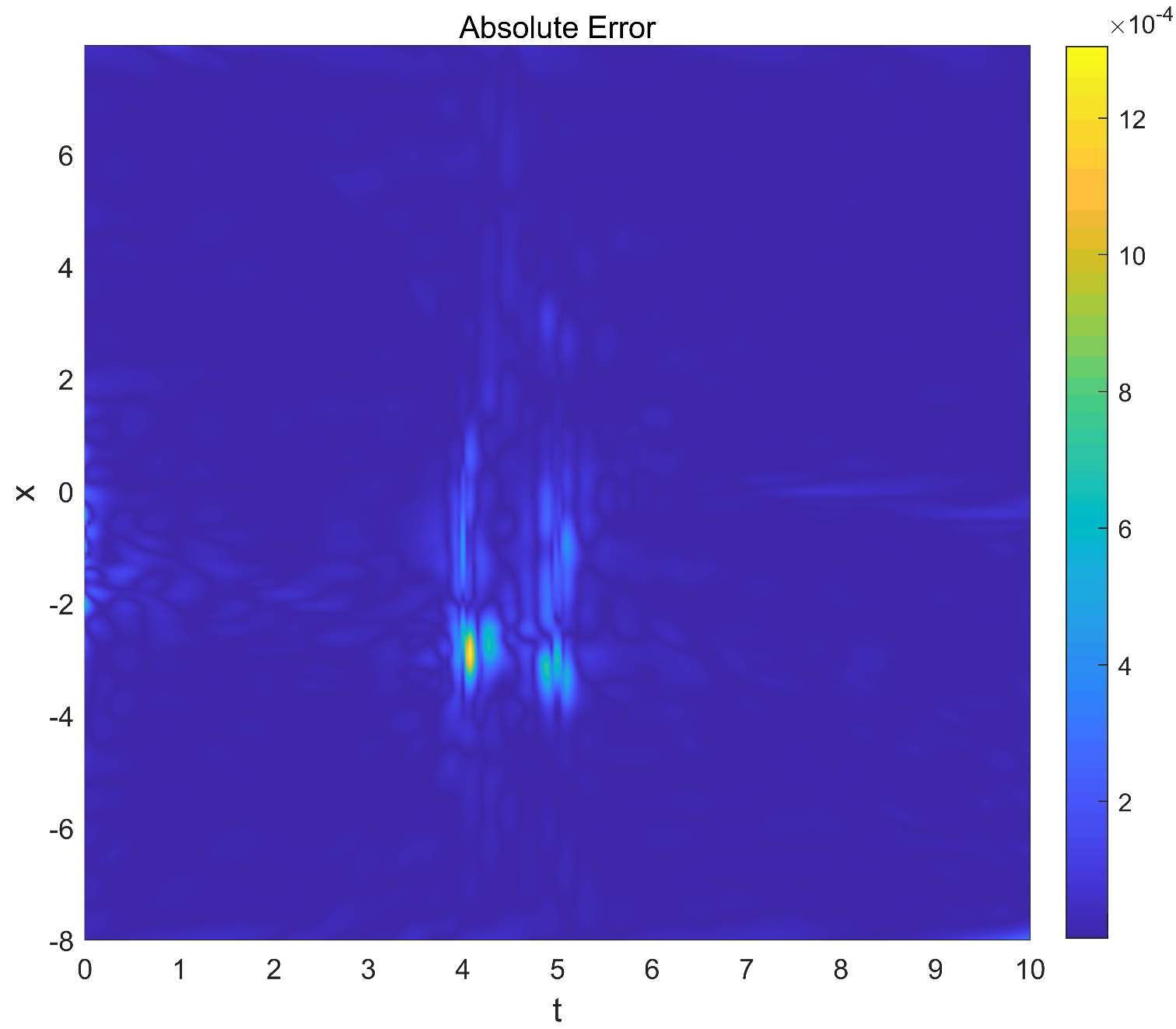}
			\end{minipage}
		}
		\subfigure{
			\begin{minipage}{0.32\textwidth}
				\centering
				\includegraphics[width=1\textwidth]{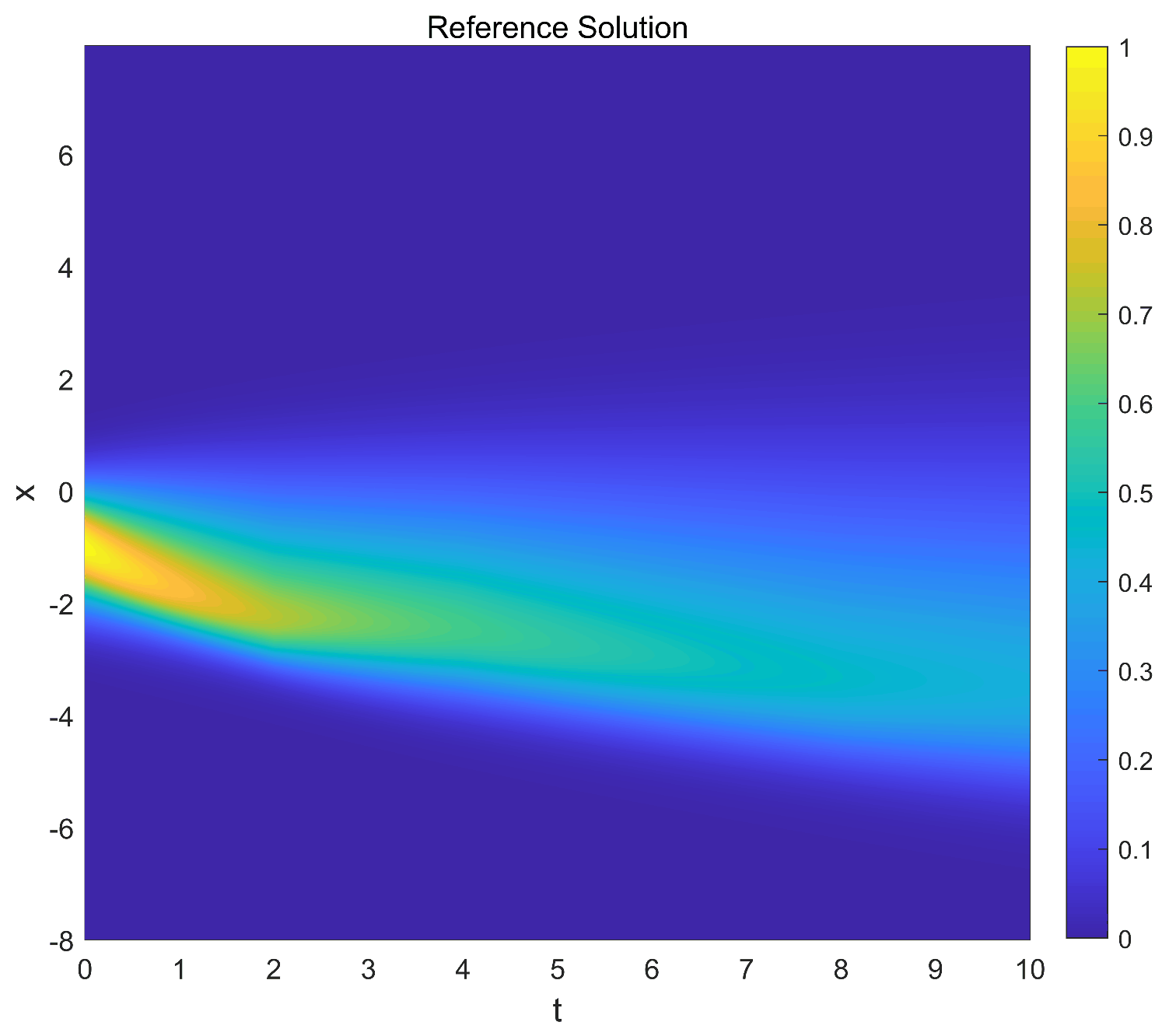}
			\end{minipage}
			\begin{minipage}{0.32\textwidth}
				\centering
				\includegraphics[width=1\textwidth]{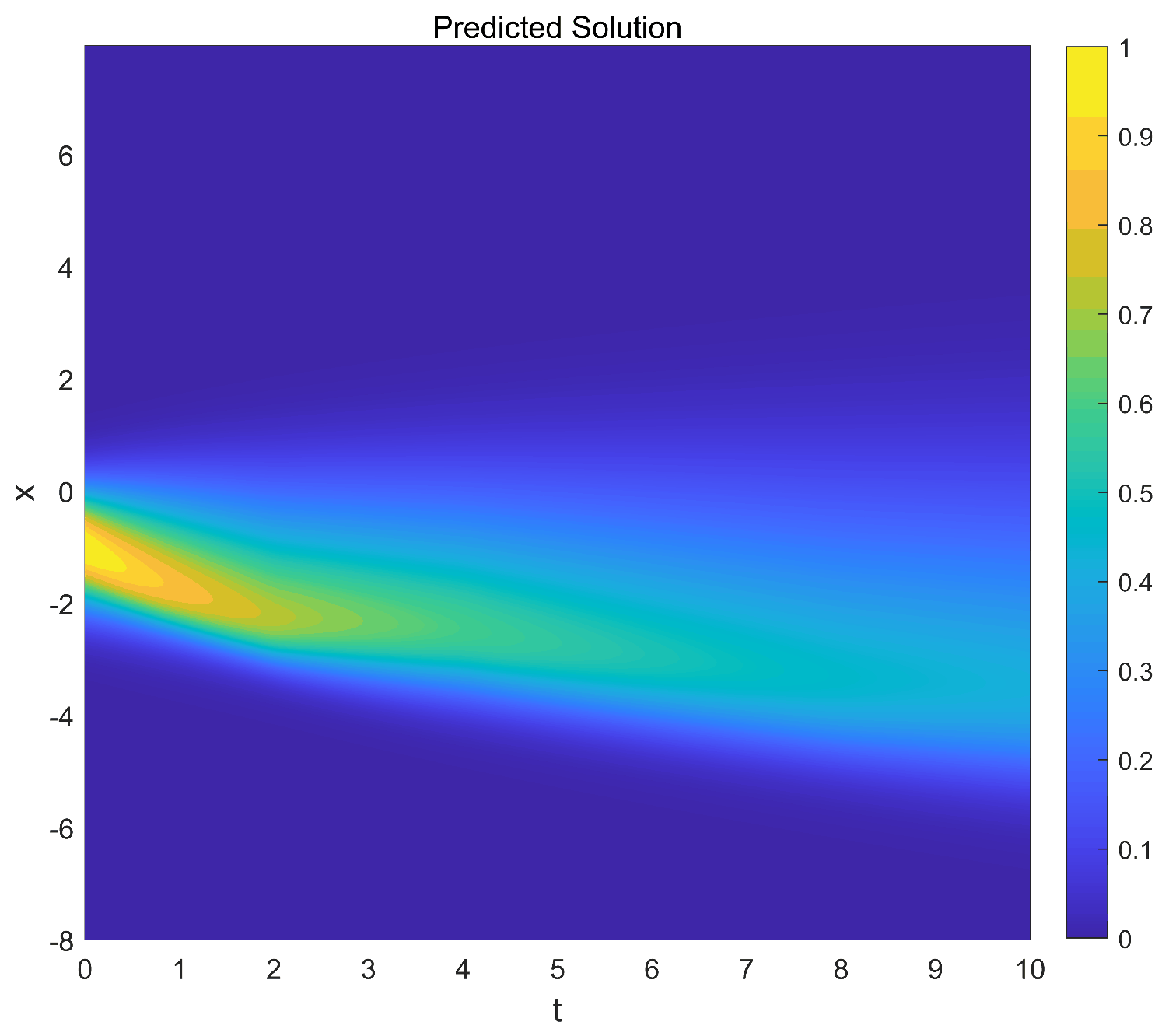}
			\end{minipage}
			\begin{minipage}{0.32\textwidth}
				\centering
				\includegraphics[width=1\textwidth]{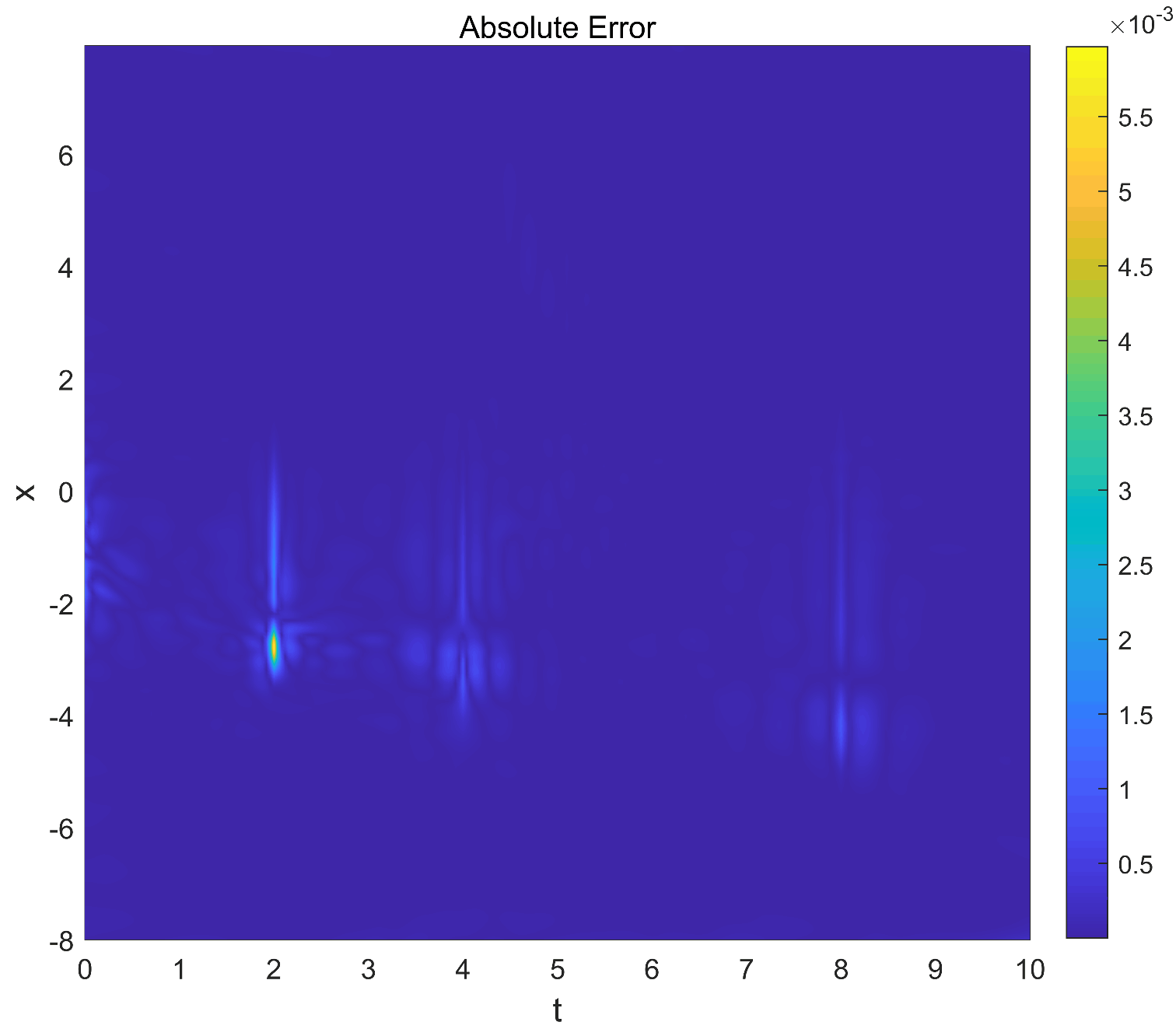}
			\end{minipage}
		}
		\caption{\label{Error for case2} From top to bottom, they are one time-varying parameter with two change point(case 2.1), modified cSPINNs and bc-PINNs for one time-varying parameter takes two values with three change points(case 2.2).}
	\end{figure}

\begin{figure}[H]
		\centering
		\begin{minipage}{0.32\textwidth}
			\centering
			\includegraphics[width=1\textwidth]{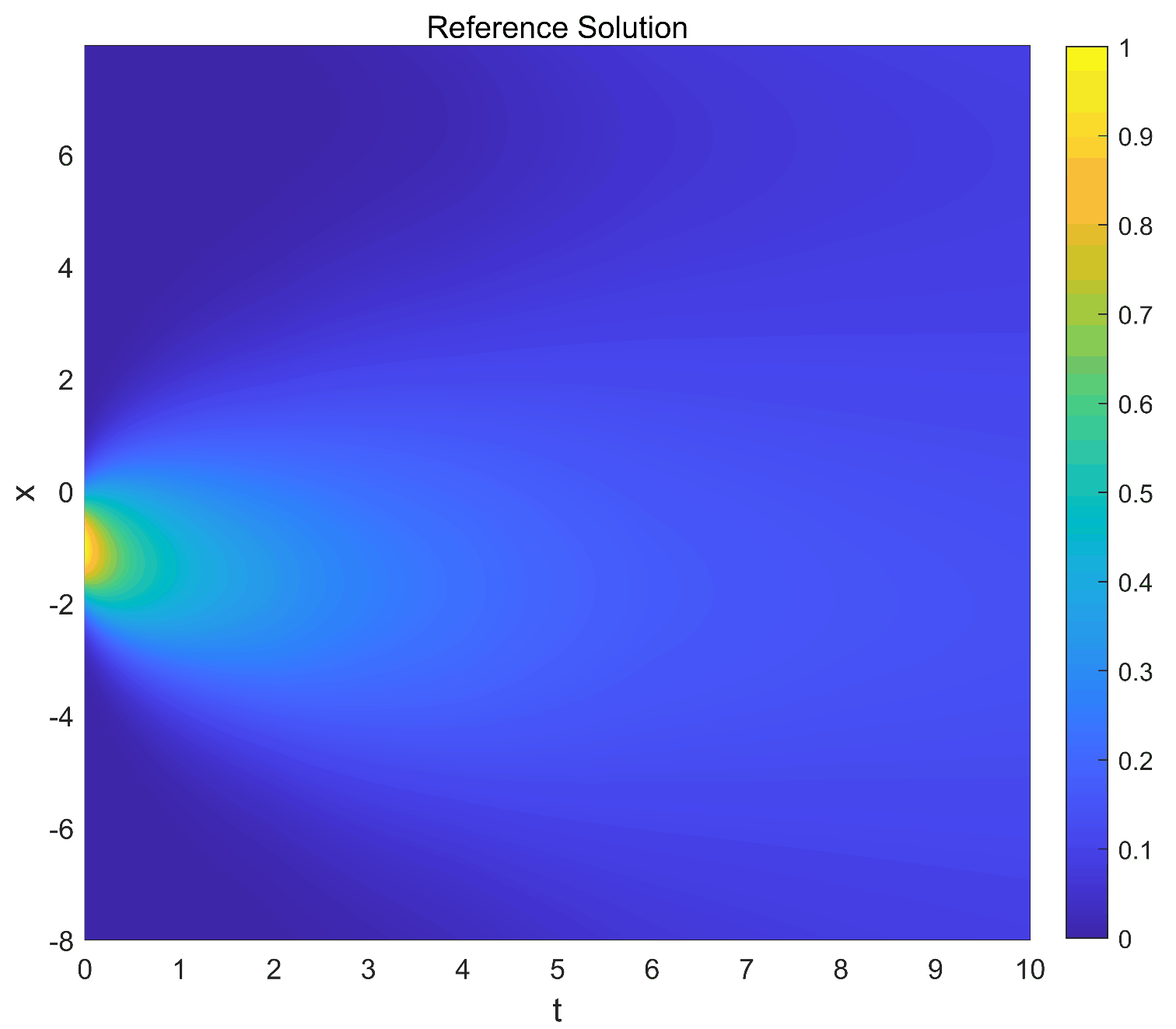}
		\end{minipage}
		\begin{minipage}{0.32\textwidth}
			\centering
			\includegraphics[width=1\textwidth]{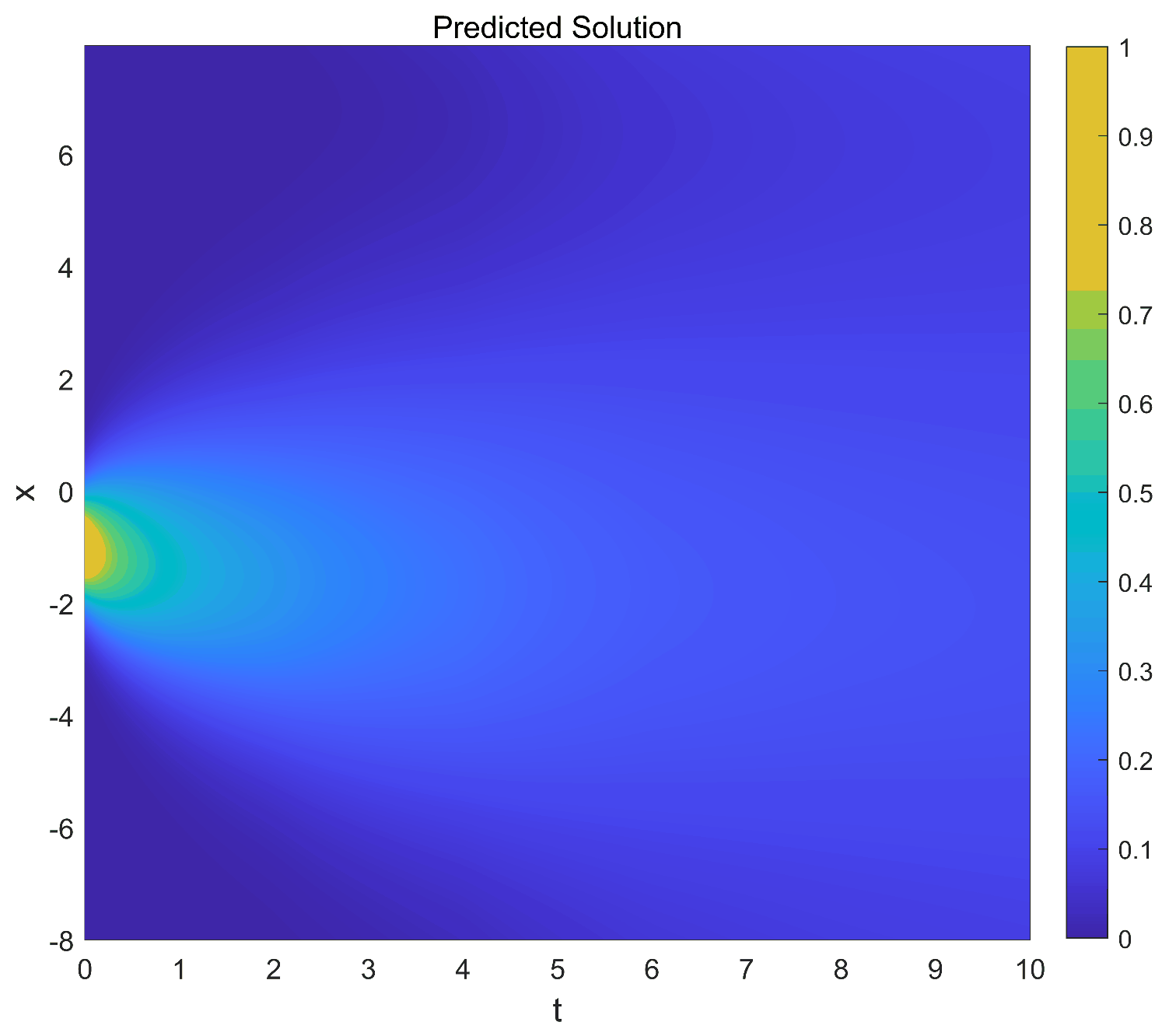}
		\end{minipage}
		\begin{minipage}{0.32\textwidth}
			\centering
			\includegraphics[width=1\textwidth]{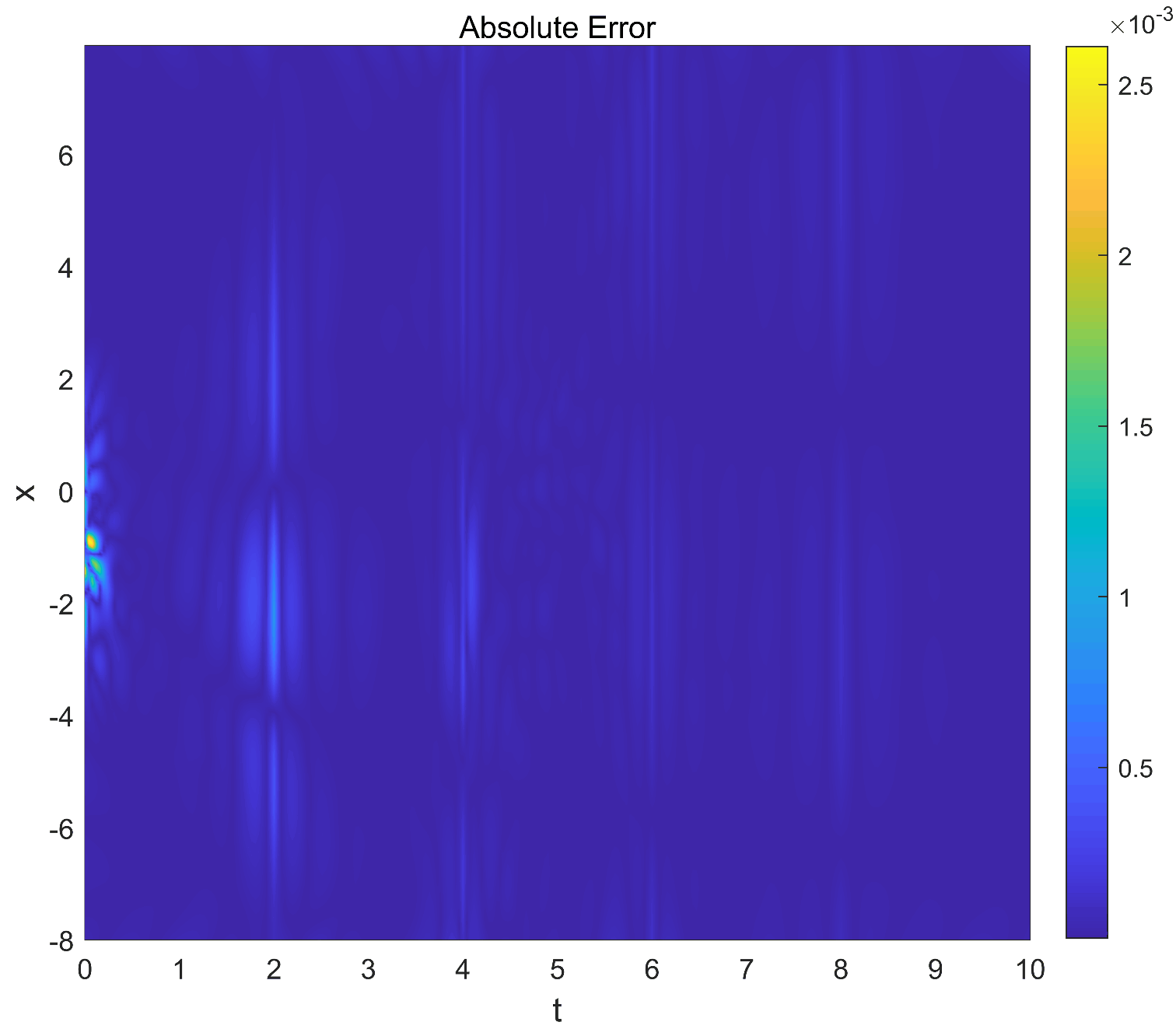}
		\end{minipage}
		\caption{\label{Error for case3}Case 3: Multiple time-varying parameters with multiple change points}
	\end{figure}

	We draw the errors of reference solution, predicted solution, and absolute error in the following three figures \ref{Error for case1}, \ref{Error for case2}, \ref{Error for case3}. 
	
	\begin{figure}[H]
		\centering
		\subfigure{
			\begin{minipage}{0.32\textwidth}
				\centering
				\includegraphics[width=1\textwidth]{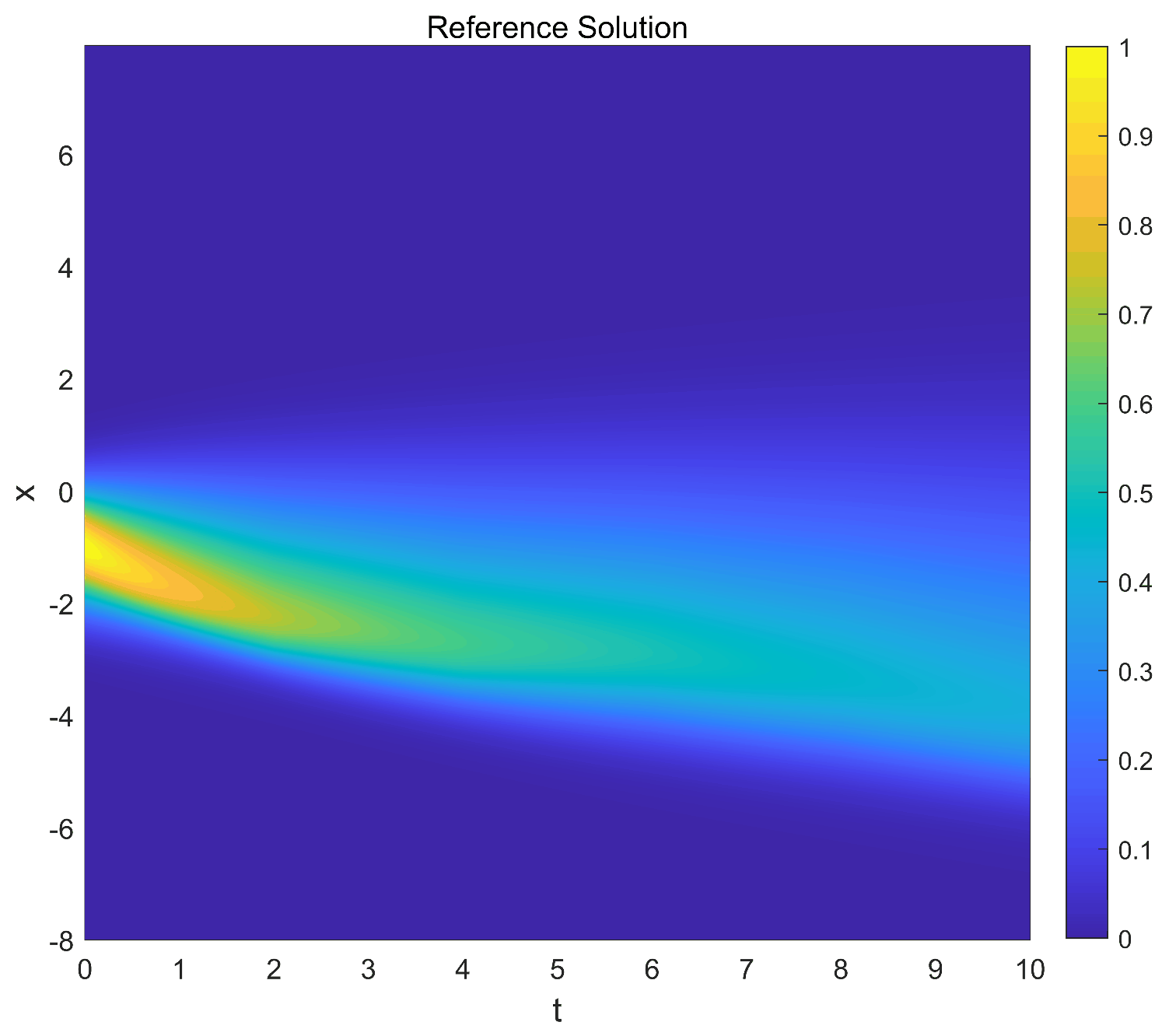}
			\end{minipage}
			\begin{minipage}{0.32\textwidth}
				\centering
				\includegraphics[width=1\textwidth]{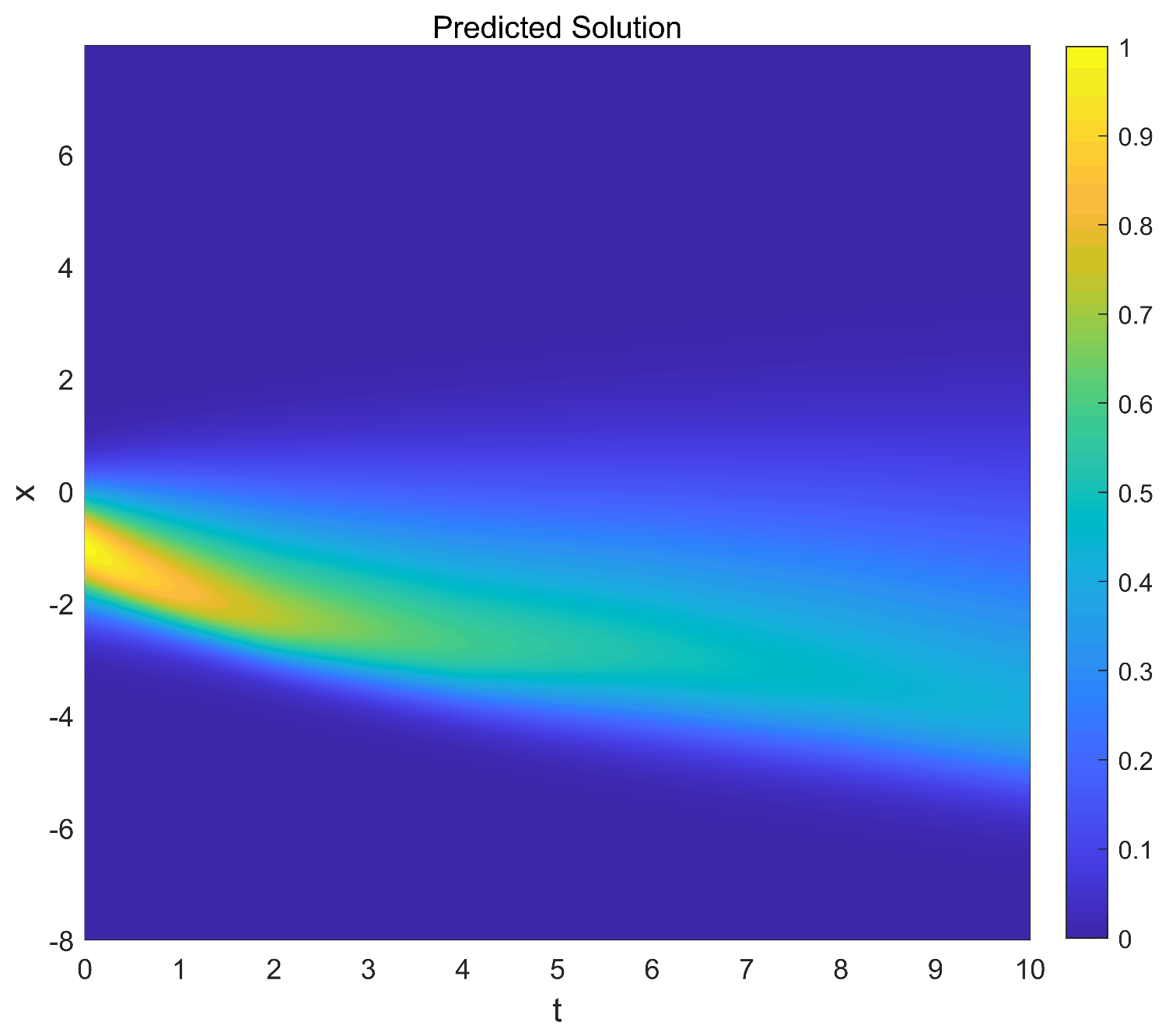}
			\end{minipage}
			\begin{minipage}{0.32\textwidth}
				\centering
				\includegraphics[width=1\textwidth]{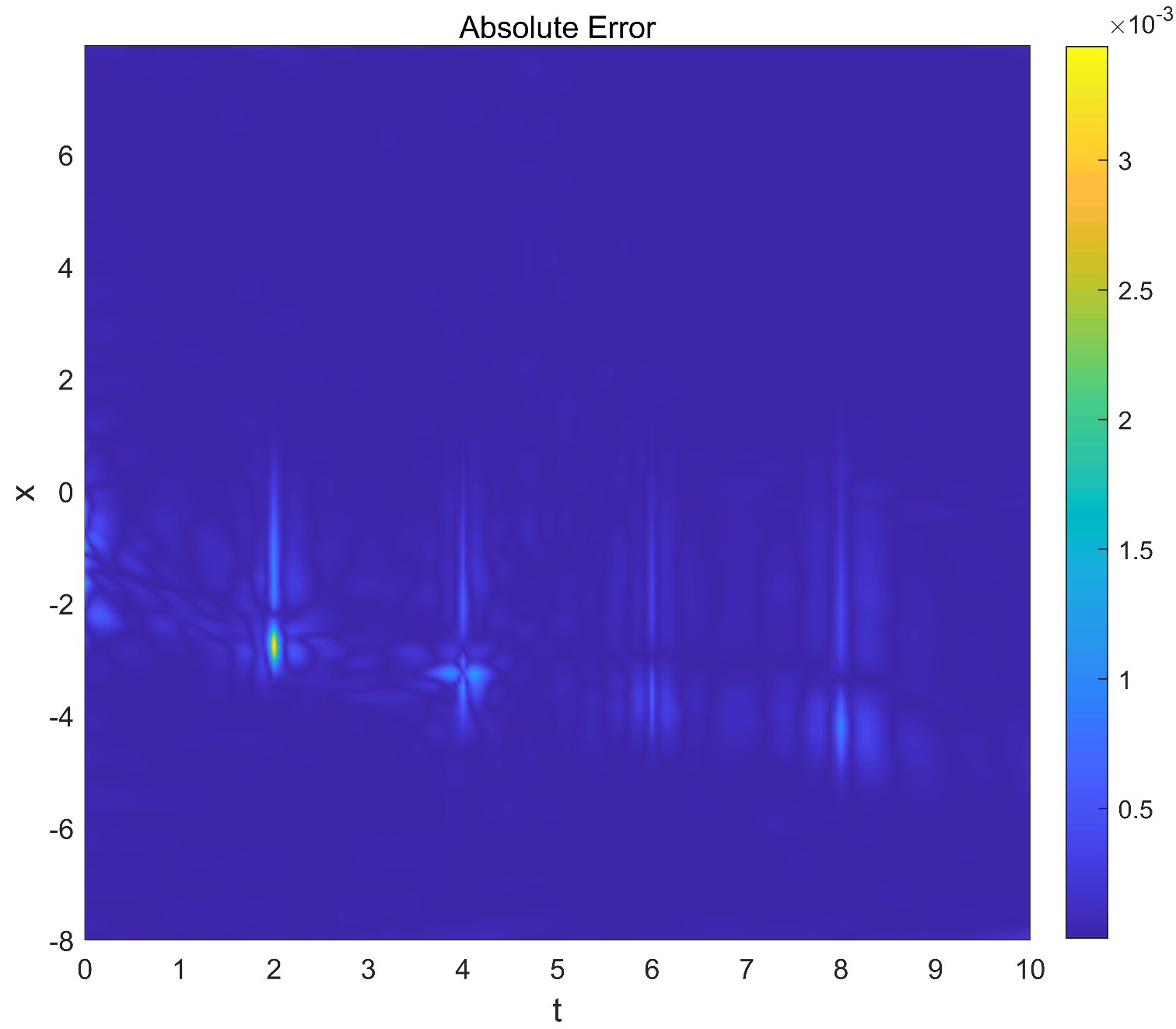}
			\end{minipage}
		}
		\subfigure{
			\begin{minipage}{0.32\textwidth}
				\centering
				\includegraphics[width=1\textwidth]{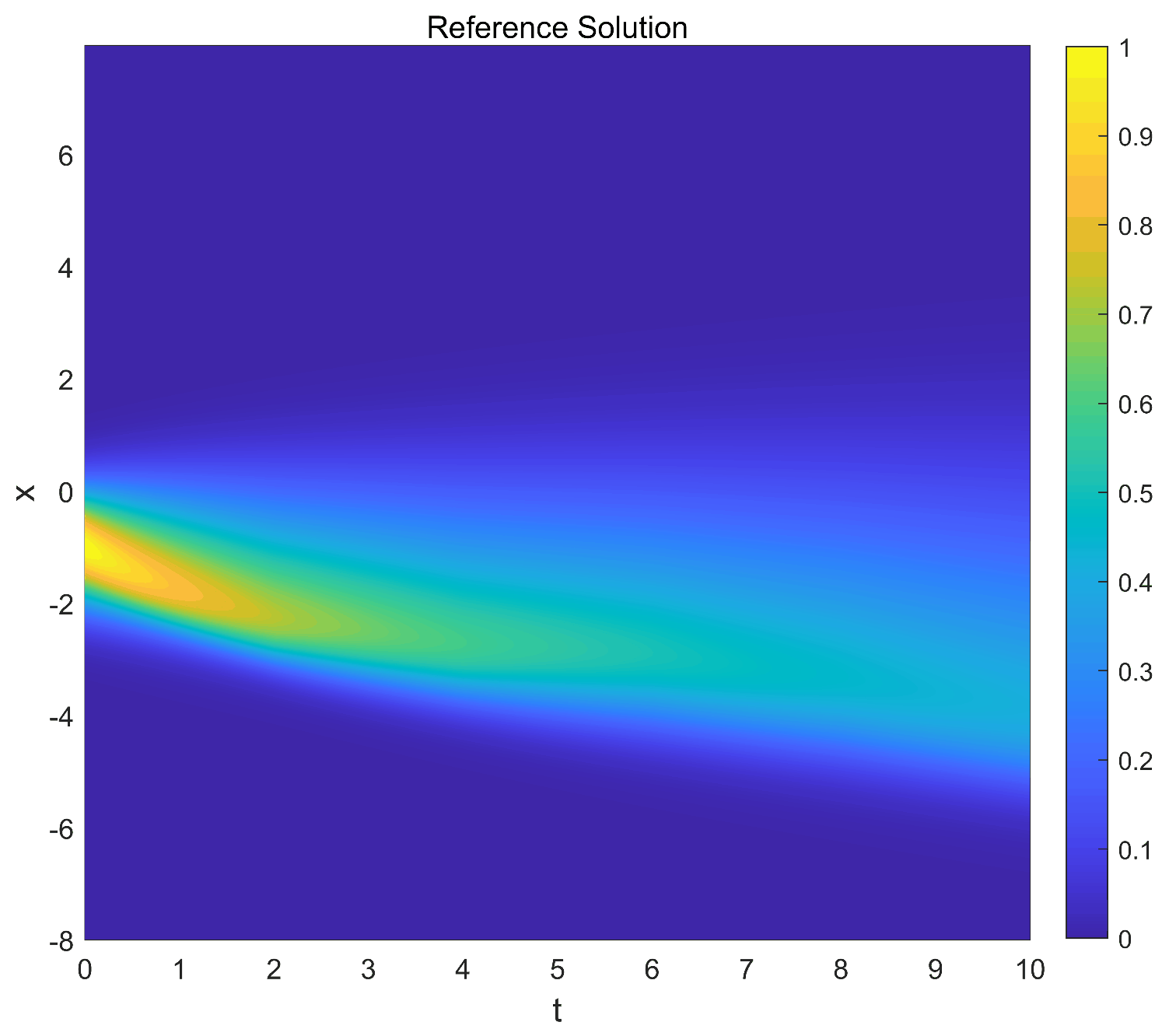}
			\end{minipage}
			\begin{minipage}{0.32\textwidth}
				\centering
				\includegraphics[width=1\textwidth]{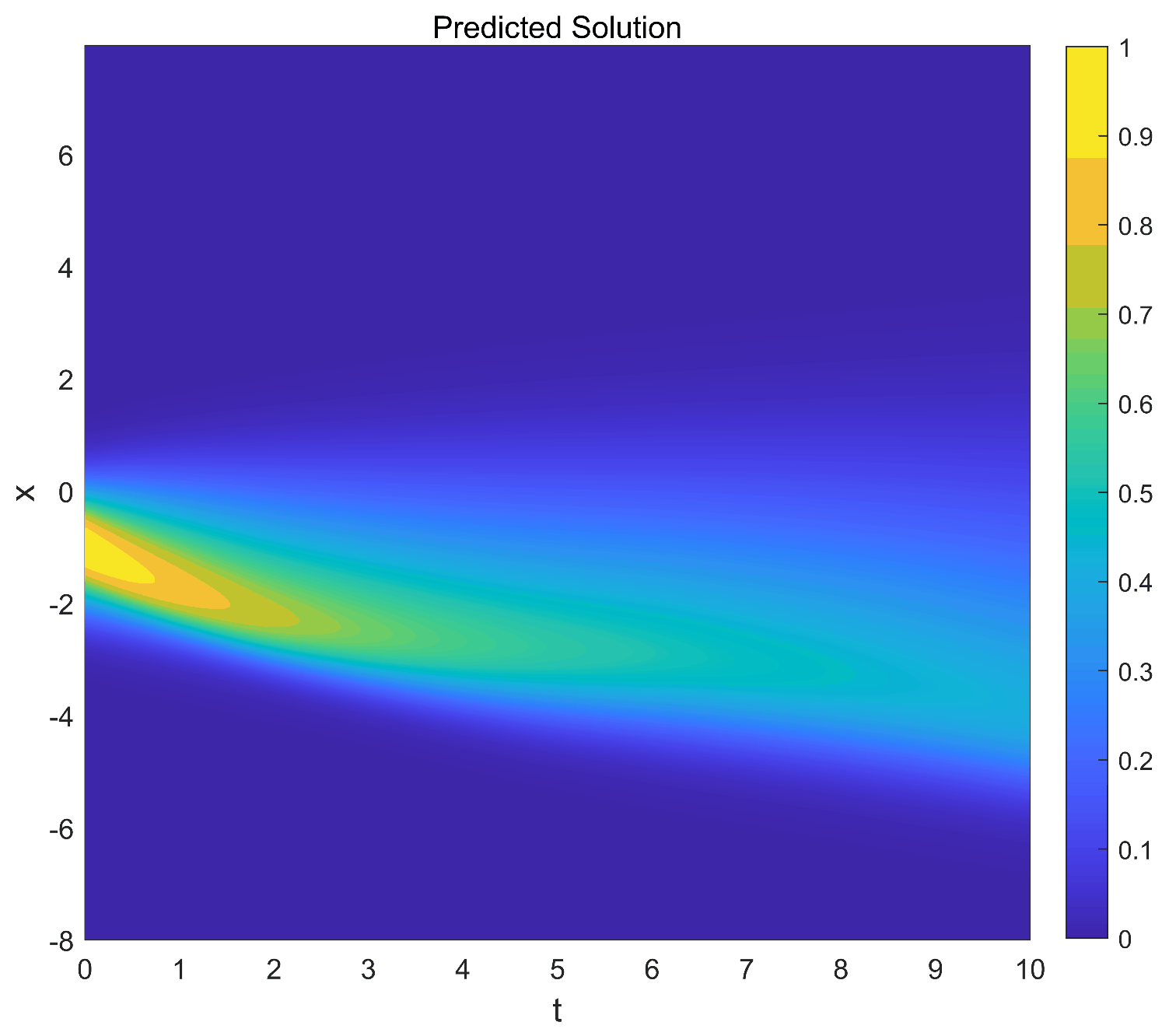}
			\end{minipage}
			\begin{minipage}{0.32\textwidth}
				\centering
				\includegraphics[width=1\textwidth]{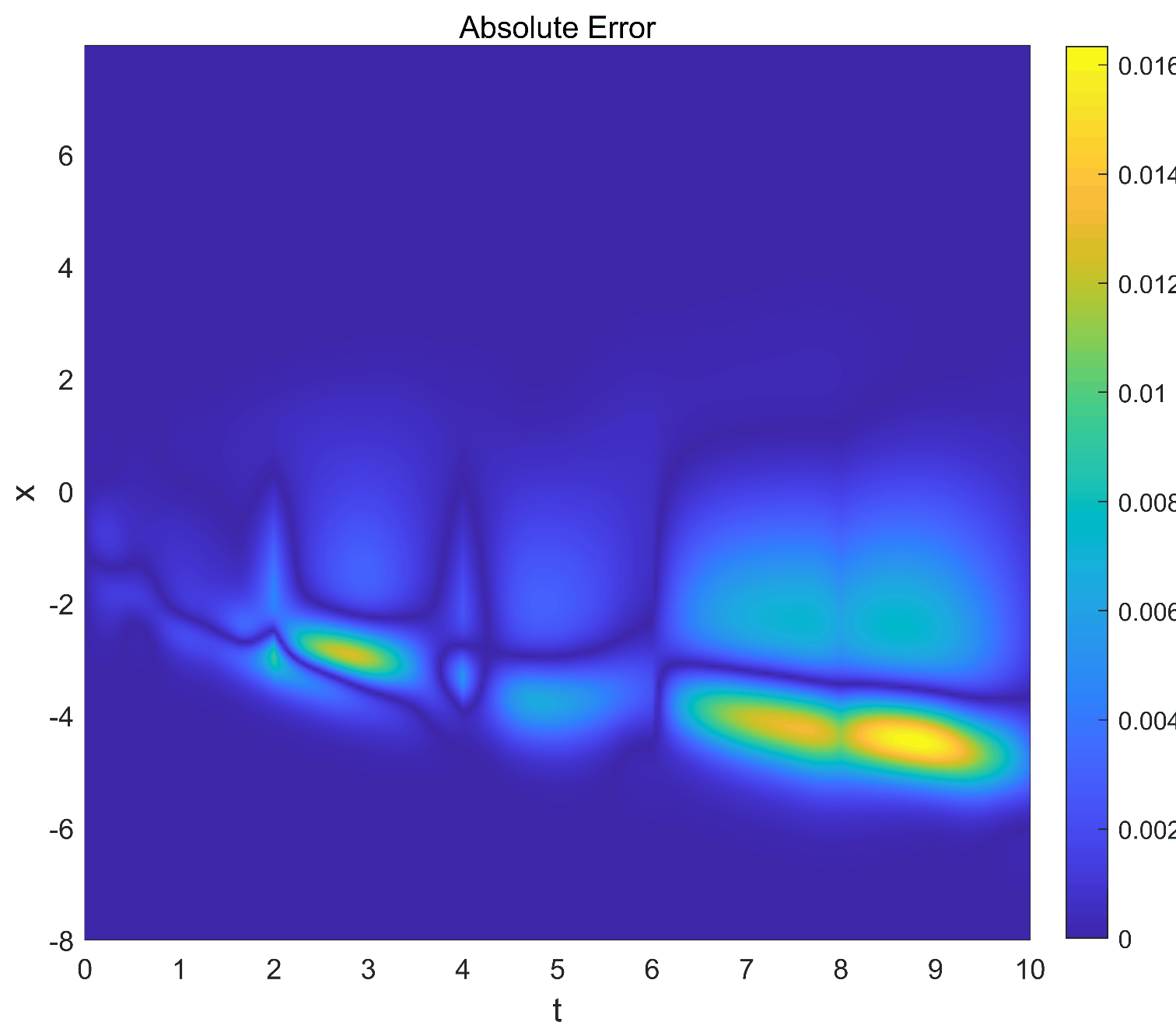}
			\end{minipage}
		}
		\caption{\label{Error for case4}Figures in the first line represent errors of modified cSPINNs and the second line is the error of bc-PINNs. The result of cSPINNs is more accurate than bc-PINNs.}
	\end{figure}  
	
	\newpage
	
	\section*{Appendix C: Absolute error between reference solution and predicted solution of 2D Space-varying Wave Equation}\label{Comparison}
	
	The following figure \ref{space_varying} is the error of reference solution, predicted solution, and absolute error. 
	\begin{figure}[H]
		\centering
		\subfigure{
			\begin{minipage}{0.31\textwidth}
				\centering
				\includegraphics[width=1\textwidth]{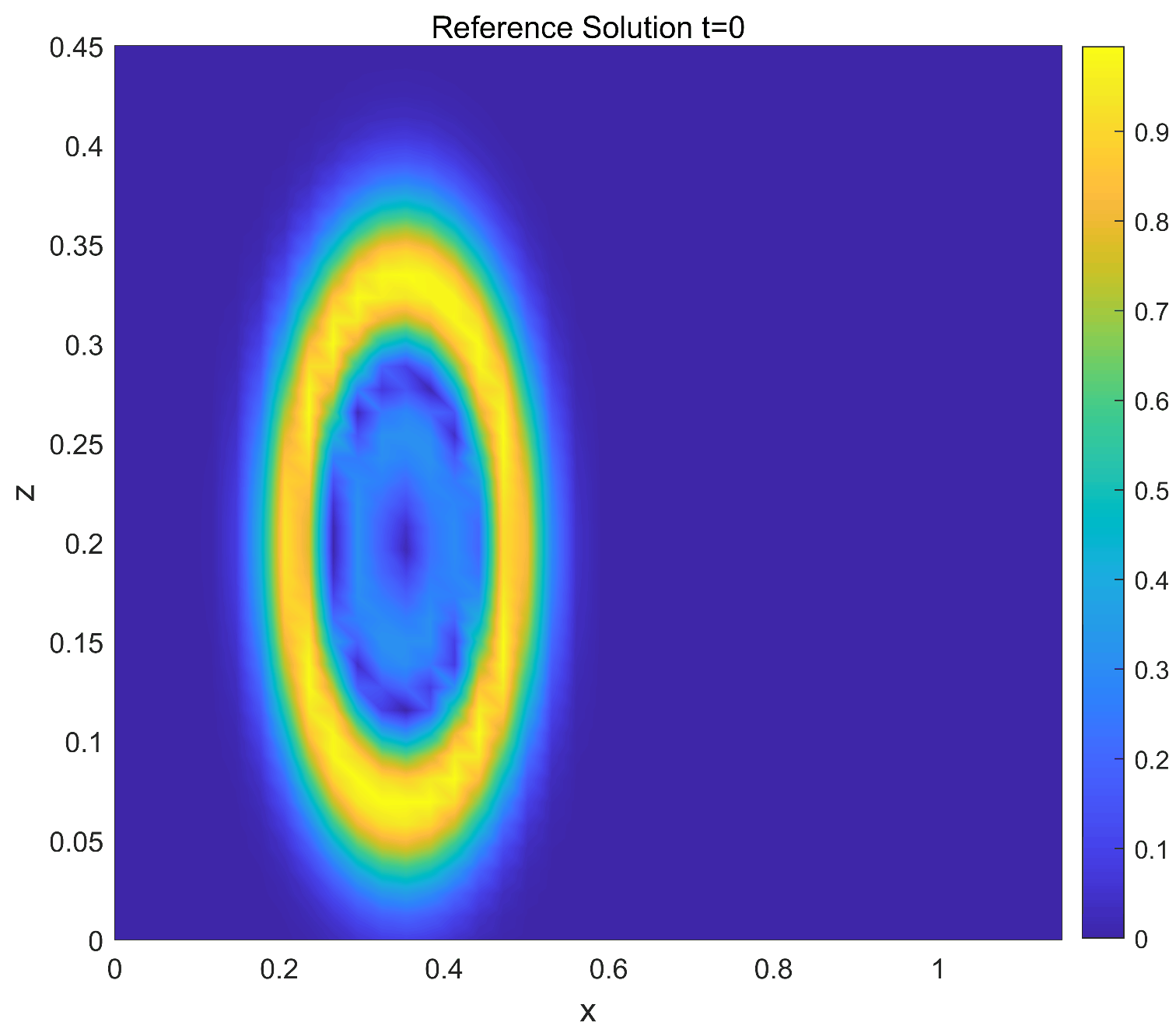}
			\end{minipage}
			\begin{minipage}{0.31\textwidth}
				\centering
				\includegraphics[width=1\textwidth]{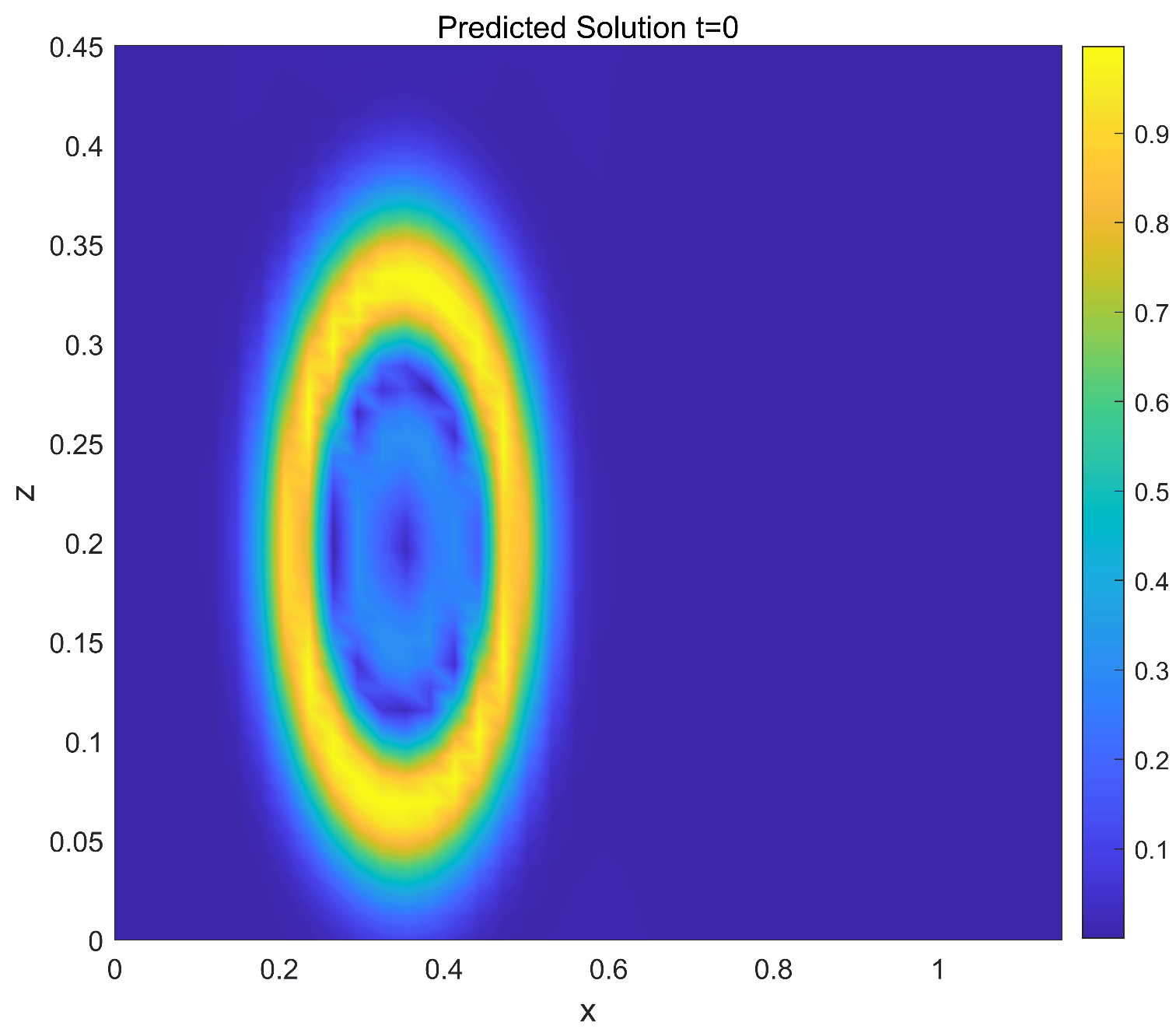}
			\end{minipage}
			\begin{minipage}{0.31\textwidth}
				\centering
				\includegraphics[width=1\textwidth]{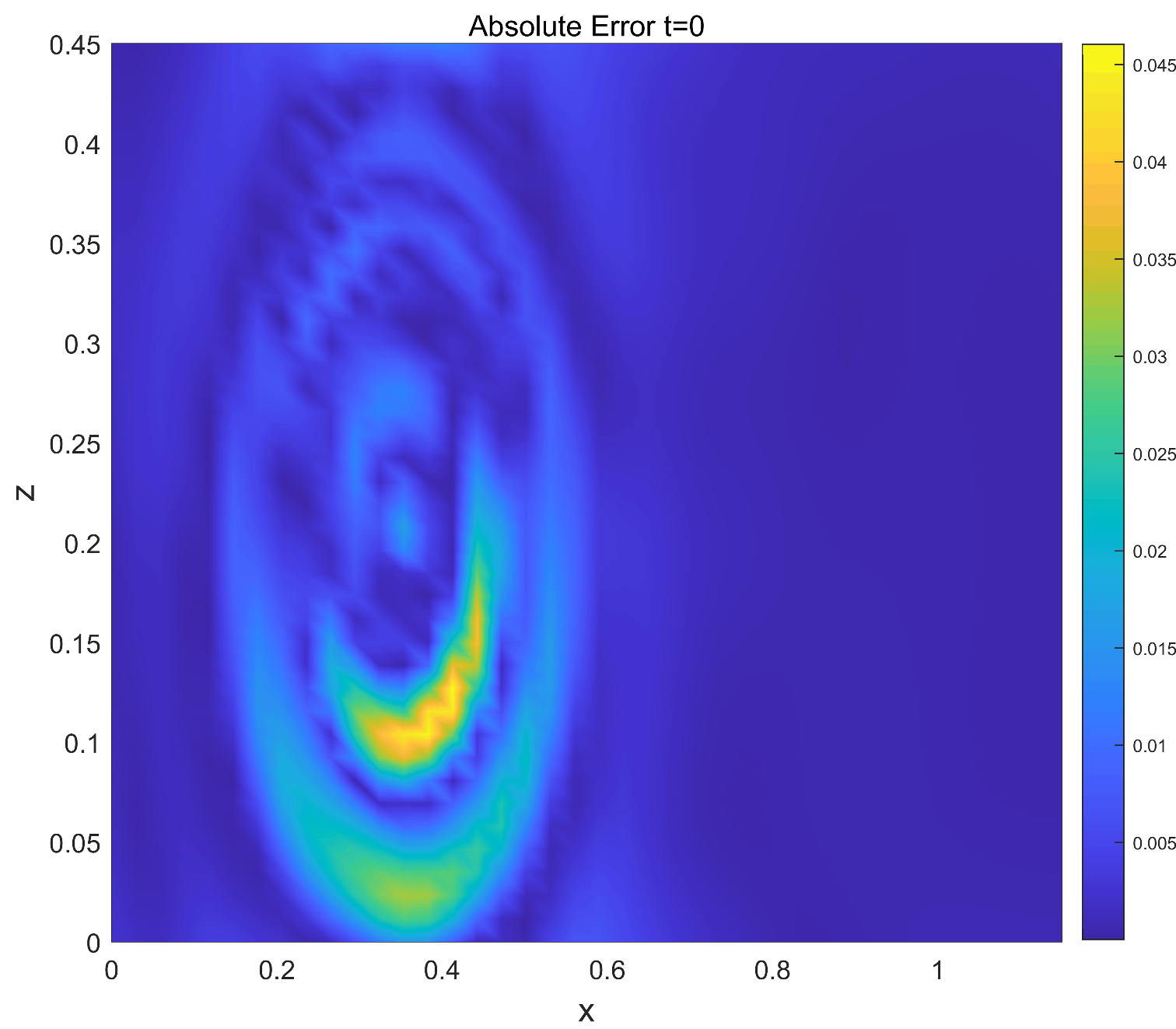}
			\end{minipage}
		}
		\subfigure{
			\begin{minipage}{0.31\textwidth}
				\centering
				\includegraphics[width=1\textwidth]{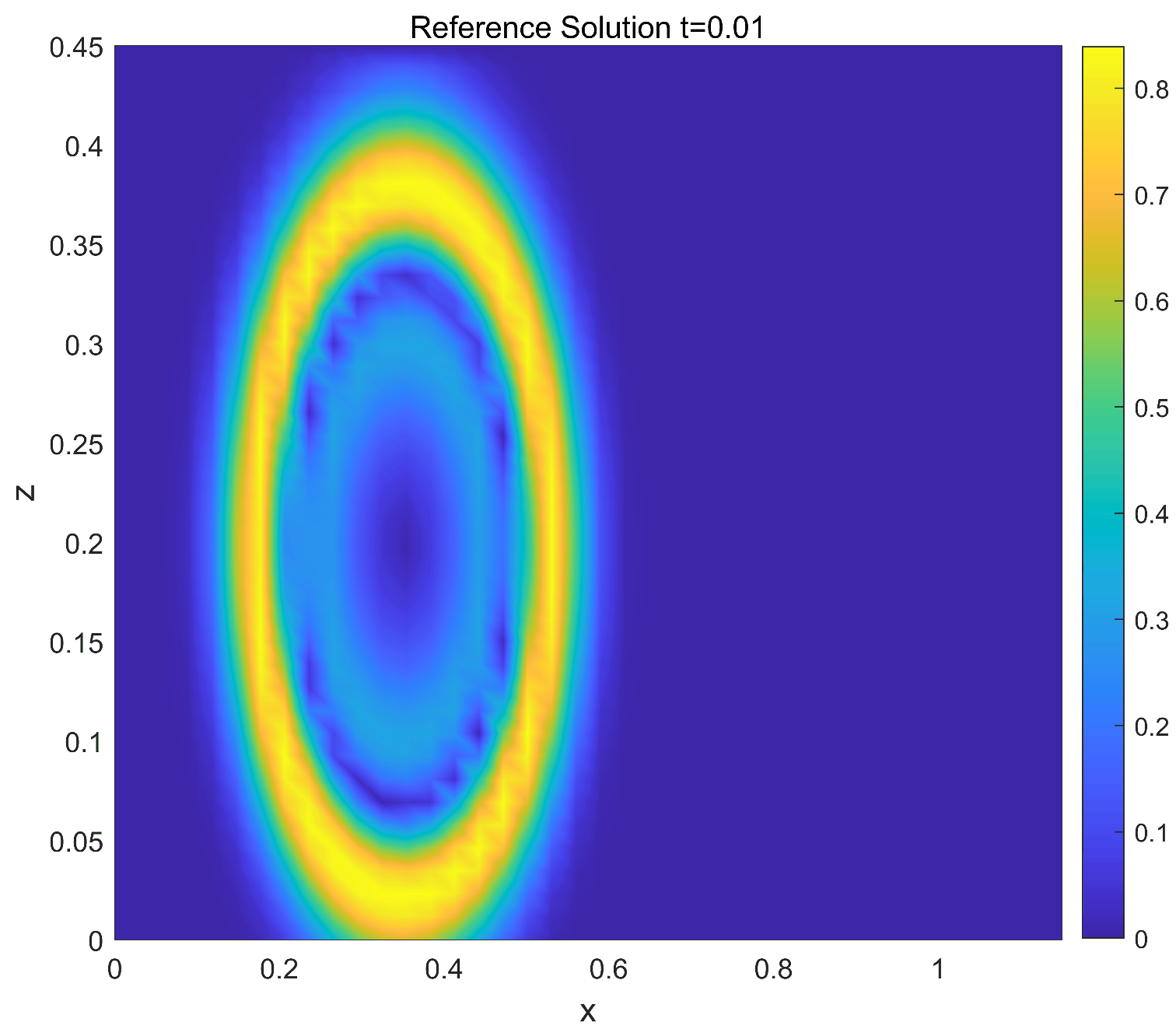}
			\end{minipage}
			\begin{minipage}{0.31\textwidth}
				\centering
				\includegraphics[width=1\textwidth]{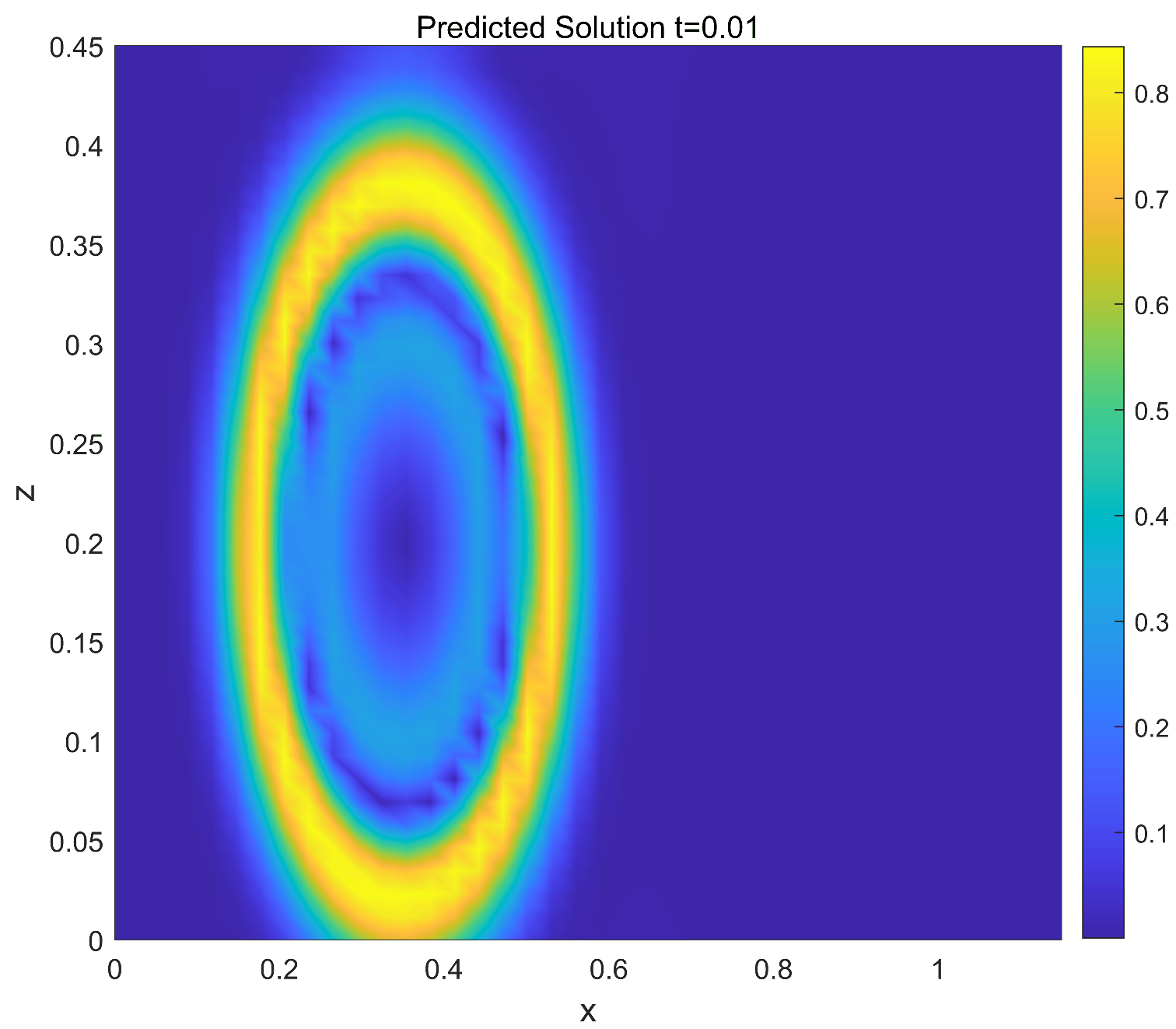}
			\end{minipage}
			\begin{minipage}{0.31\textwidth}
				\centering
				\includegraphics[width=1\textwidth]{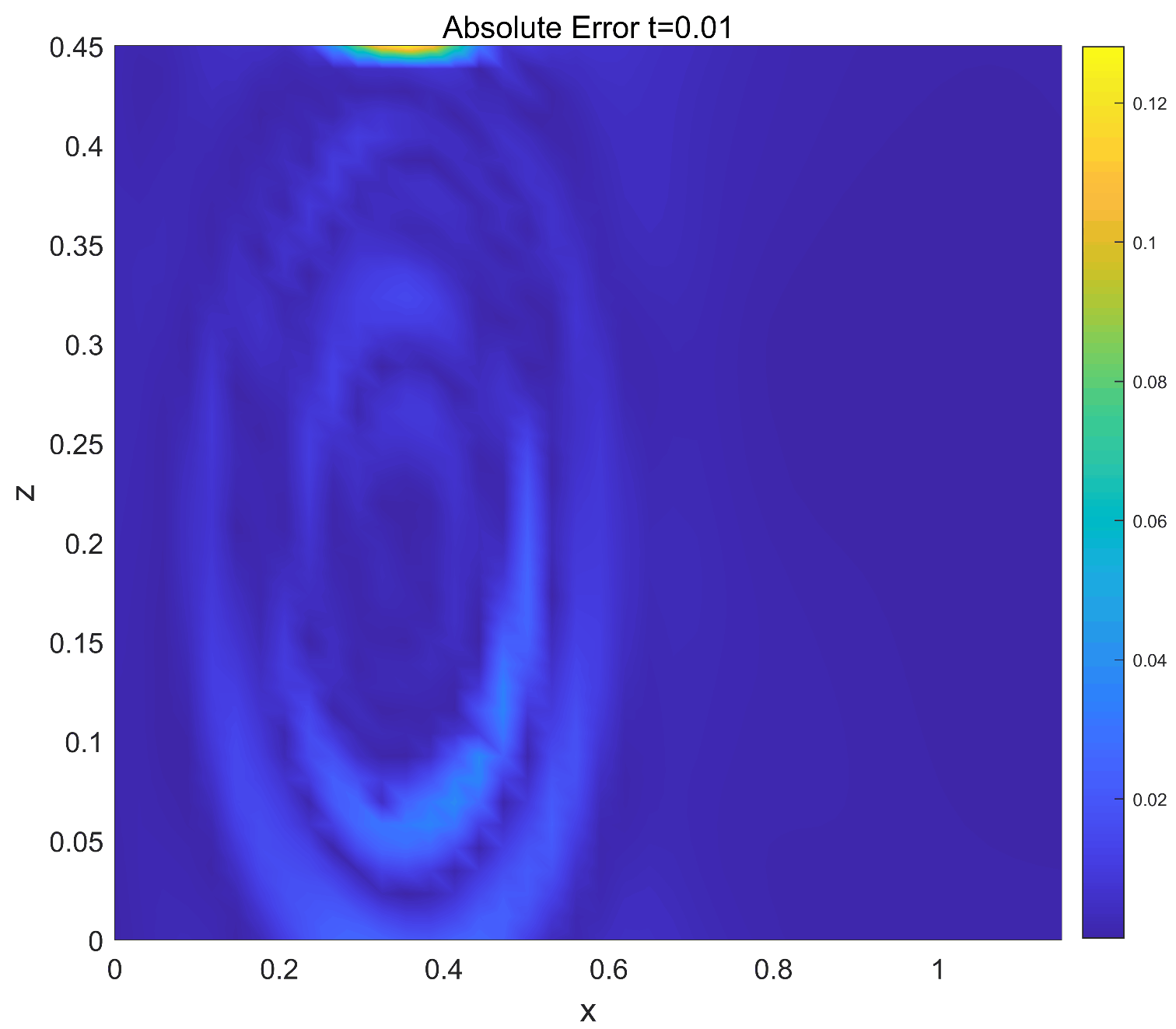}
			\end{minipage}
		}
		\subfigure{
			\begin{minipage}{0.31\textwidth}
				\centering
				\includegraphics[width=1\textwidth]{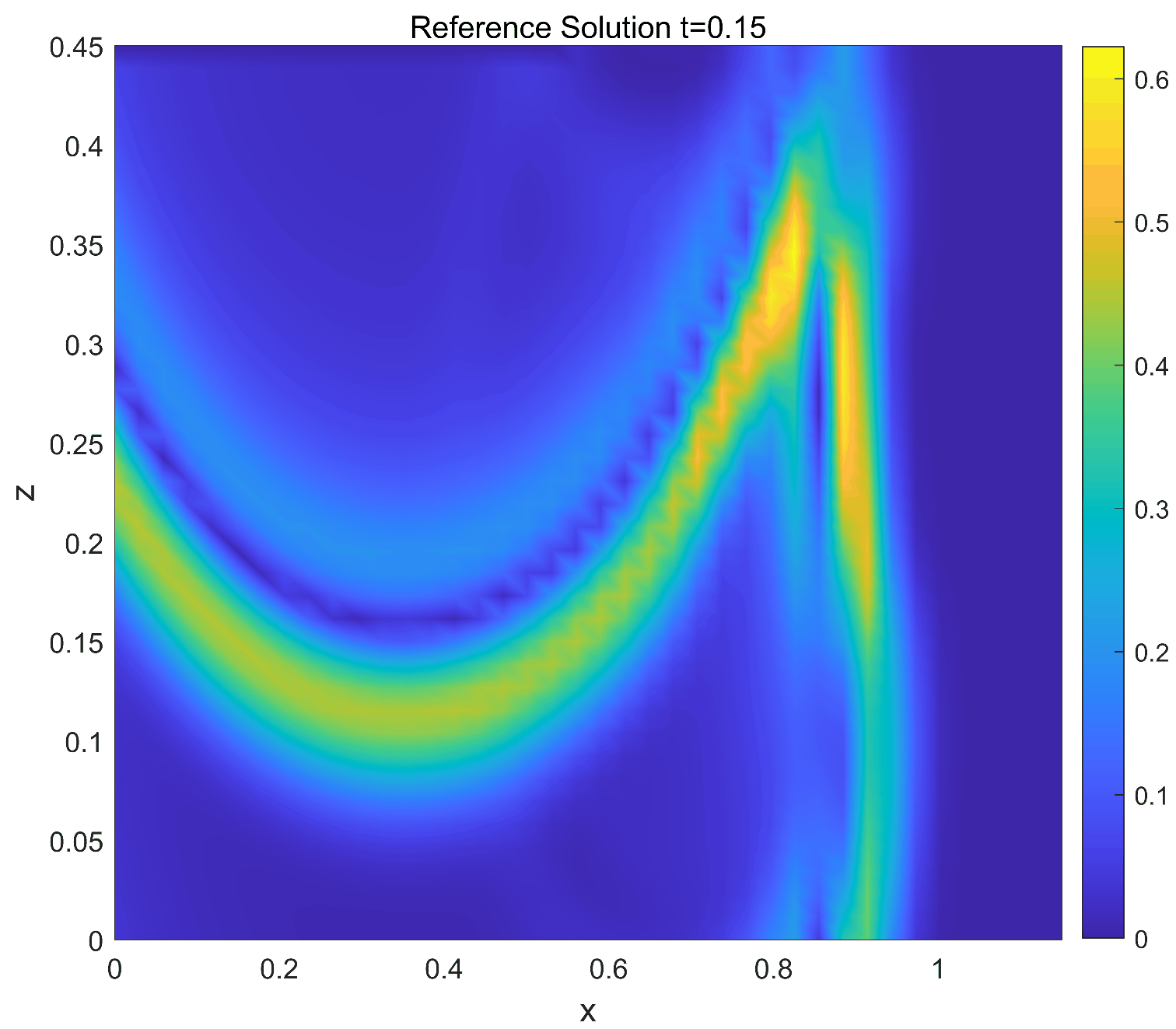}
			\end{minipage}
			\begin{minipage}{0.31\textwidth}
				\centering
				\includegraphics[width=1\textwidth]{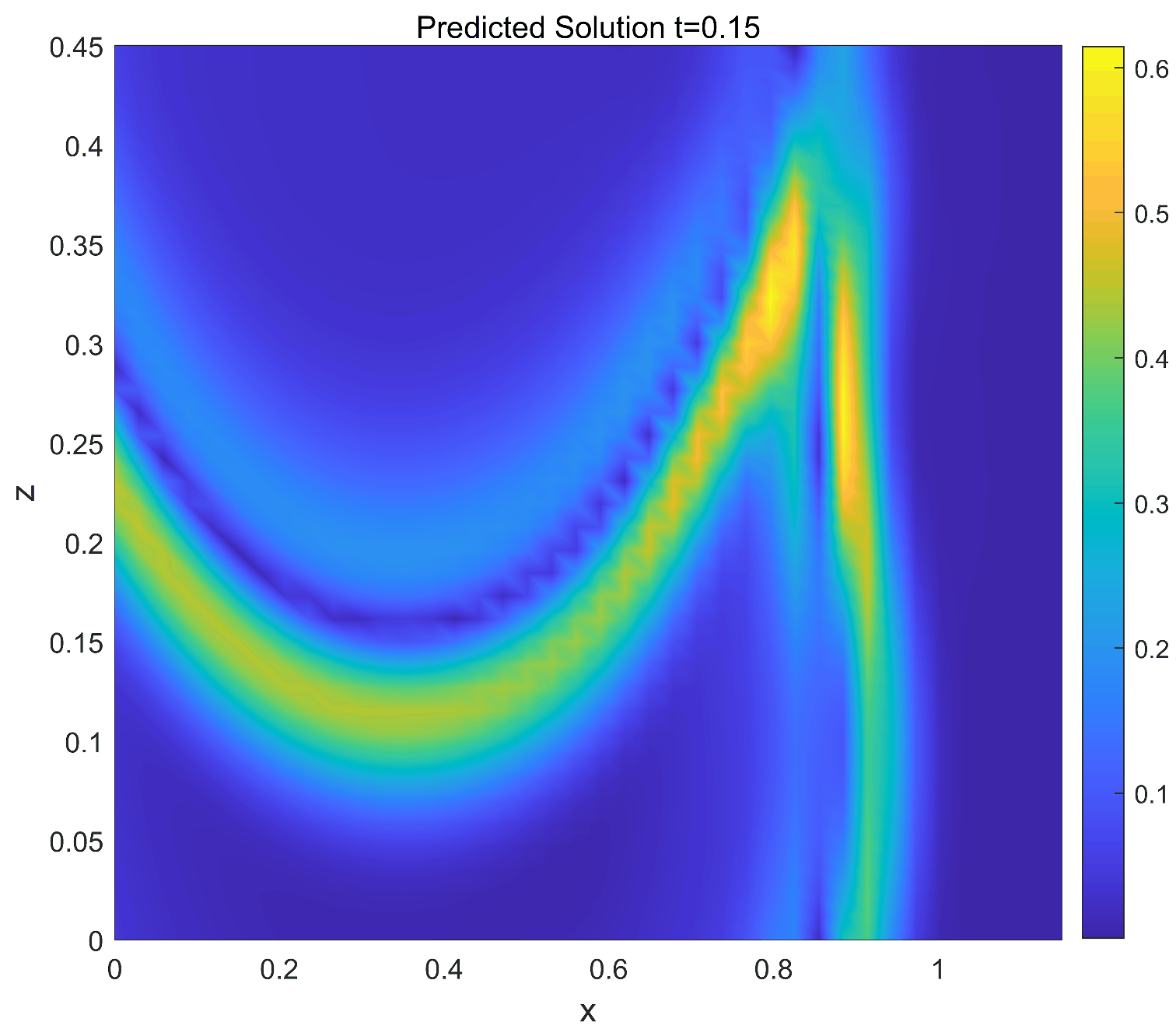}
			\end{minipage}
			\begin{minipage}{0.31\textwidth}
				\centering
				\includegraphics[width=1\textwidth]{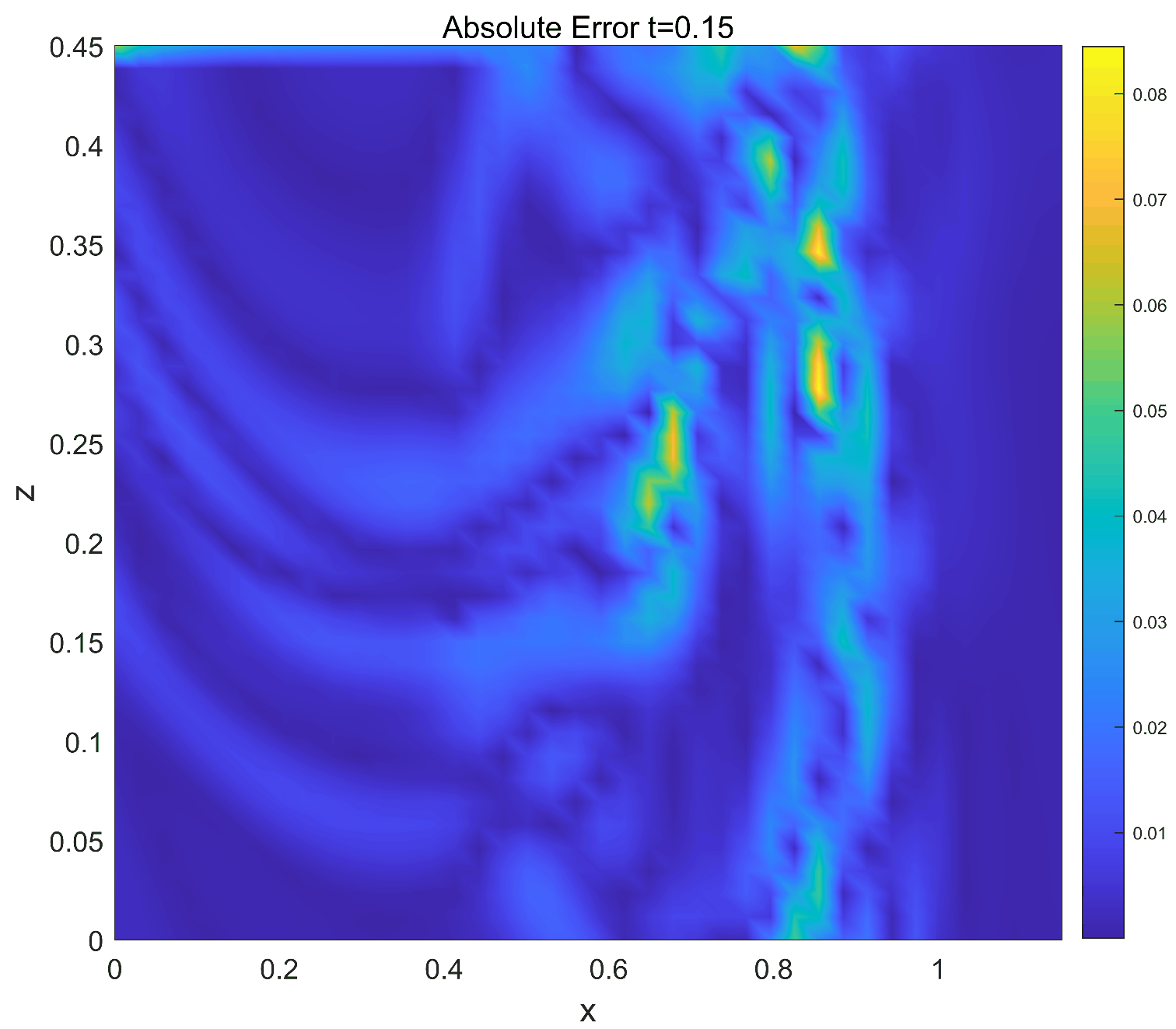}
			\end{minipage}
		}
		\caption{\label{space_varying} Comparison between ground truth and modeled wavefields and their absolute pointwise differences for the synthetic crosswell experiment with a discontinuous ellipsoidal anomaly from SpecFem2D at $t = 0, t = 0.01 s$ and $t=0.15s$ are used as the training data.}
	\end{figure}

	
	\newpage

\end{document}